\documentclass[10pt,journal,compsoc]{IEEEtran}
\usepackage[table,xcdraw]{xcolor}
\usepackage{caption}
\usepackage{hyperref}
\usepackage{float}
\usepackage{amsmath}
\usepackage{indentfirst}
\usepackage{amssymb}
\usepackage{amsmath,amsfonts}
\usepackage{algorithmic}
\usepackage{algorithm}
\usepackage{array}
\usepackage{color}
\usepackage{subfig}
\usepackage{textcomp}
\usepackage{stfloats}
\usepackage{url}
\usepackage{verbatim}
\usepackage{graphicx}
\usepackage{cite}
\usepackage{hyperref}
\usepackage{xr-hyper}
\usepackage[figuresright]{rotating}
\usepackage{pdfpages}
\usepackage{multirow}
\usepackage{colortbl}
\usepackage[normalem]{ulem}
\useunder{\uline}{\ul}{}
\usepackage{booktabs} 
\graphicspath{{figs/}}
\makeatletter
\newcommand*{\addFileDependency}[1]{
  \typeout{(#1)}
  \@addtofilelist{#1}
  \IfFileExists{#1}{}{\typeout{No file #1.}}
}
\makeatother

\begin{document}

\title{A Survey of Camouflaged Object Detection \\ and Beyond}

\author{Fengyang~Xiao,~\IEEEmembership{}
        Sujie~Hu,~\IEEEmembership{}
        Yuqi~Shen,~\IEEEmembership{}
        Chengyu~Fang,~\IEEEmembership{}
        Jinfa~Huang,~\IEEEmembership{} \\
        Chunming~He, ~\IEEEmembership{}
        Longxiang~Tang,~\IEEEmembership{}
        Ziyun~Yang,~\IEEEmembership{}
        Xiu~Li~\IEEEmembership{}
\thanks{
\textit{(Corresponding author: Chunming He, Xiu Li)
(Fengyang Xiao and Sujie Hu contributed equally.) 
}}
\thanks{Fengyang Xiao, Sujie Hu, Yuqi Shen, Chengyu Fang, Chunming He, Longxiang Tang and Xiu Li are with Tsinghua Shenzhen International Graduate School, Tsinghua University, Shenzhen 518055, China (e-mail: xiaofy5@mail2.sysu.edu.cn; husj24@mails.tsinghua.edu.cn; ericsyq\_buaa@163.com; chengyufang.thu@gmail.com; chunminghe19990224@gmail.com; lloong.x@gmail.com; li.xiu@sz.tsinghua.edu.cn).} 
\thanks{Jinfa Huang are with School of Electrical and Computer Engineering, Peking University, Shenzhen, Guangdong 518055, China(e-mail: vhjf305@gmail.com).}
\thanks{Chunming He and ziyun yang are with Department of Biomedical Engineering, Duke University, Durham, NC 27708, USA(e-mail: chunminghe19990224@gmail.com; ziyun.yang@duke.edu).}}

\markboth{Submitted to CAAI AIR}%
{Xiao \MakeLowercase{\textit{et al.}}: A Survey of Camouflaged Object Detection and Beyond}

\IEEEtitleabstractindextext{%
\begin{abstract}
Camouflaged Object Detection (COD) refers to the task of identifying and segmenting objects that blend seamlessly into their surroundings, posing a significant challenge for computer vision systems. In recent years, COD has garnered widespread attention due to its potential applications in surveillance, wildlife conservation, autonomous systems, and more. While several surveys on COD exist, they often have limitations in terms of the number and scope of papers covered, particularly regarding the rapid advancements made in the field since mid-2023. To address this void, we present the most comprehensive review of COD to date, encompassing both theoretical frameworks and practical contributions to the field. This paper explores various COD methods across four domains, including both image-level and video-level solutions, from the perspectives of traditional and deep learning approaches. We thoroughly investigate the correlations between COD and other camouflaged scenario methods, thereby laying the theoretical foundation for subsequent analyses. Furthermore, we delve into novel tasks such as referring-based COD and collaborative COD, which have not been fully addressed in previous works. Beyond object-level detection, we also summarize extended methods for instance-level tasks, including camouflaged instance segmentation, counting, and ranking. Additionally, we provide an overview of commonly used benchmarks and evaluation metrics in COD tasks, conducting a comprehensive evaluation of deep learning-based techniques in both image and video domains, considering both qualitative and quantitative performance.
Finally, we discuss the limitations of current COD models and propose 9 promising directions for future research, focusing on addressing inherent challenges and exploring novel, meaningful technologies. This comprehensive examination aims to deepen the understanding of COD models and related methods in camouflaged scenarios. For those interested, a curated list of COD-related techniques, datasets, and additional resources can be found at \url{https://github.com/ChunmingHe/awesome-concealed-object-segmentation}.
\end{abstract}

}

\maketitle

\IEEEdisplaynontitleabstractindextext

\IEEEpeerreviewmaketitle

\setlength{\abovedisplayskip}{2pt}
\setlength{\belowdisplayskip}{2pt}

\ifCLASSOPTIONcompsoc
\IEEEraisesectionheading{\section{Introduction}\label{sec:introduction}}

\IEEEPARstart{O}{bject} detection, a fundamental task in computer vision, involves identifying and locating objects within images or videos. It comprises various fine-grained subfields: generic object detection (GOD)\cite{pu2023fine,pu2023adaptive,pu2023rank,wang2023swinmm}, salient object detection (SOD)\cite{zhuge2022salient,zhao2019egnet,wu2019cascaded}, and camouflaged object detection (COD)\cite{fan2020camouflaged,yang2021uncertainty,He2023Camouflaged}. GOD aims to detect general objects, while SOD identifies prominent objects that stand out from the background. In contrast, COD targets those objects that blend into their surroundings, making it an extremely challenging task. Fig. \ref{fig:combined3examples} illustrates the relationship between the target dog and its background across GOD, SOD, and COD tasks, sourced from the classical datasets\cite{cocolin2014microsoft,DUTSwang2017learning,fan2022concealed} of them.

COD has recently garnered increasing attention and rapid development for its advantages in facilitating the development of visual perception for nuance discrimination and promoting various valuable real-life applications, rang-\vspace{-5mm}
    \begin{center}
    \setlength{\abovecaptionskip}{0.1cm}     
    \includegraphics[width=.49\textwidth]{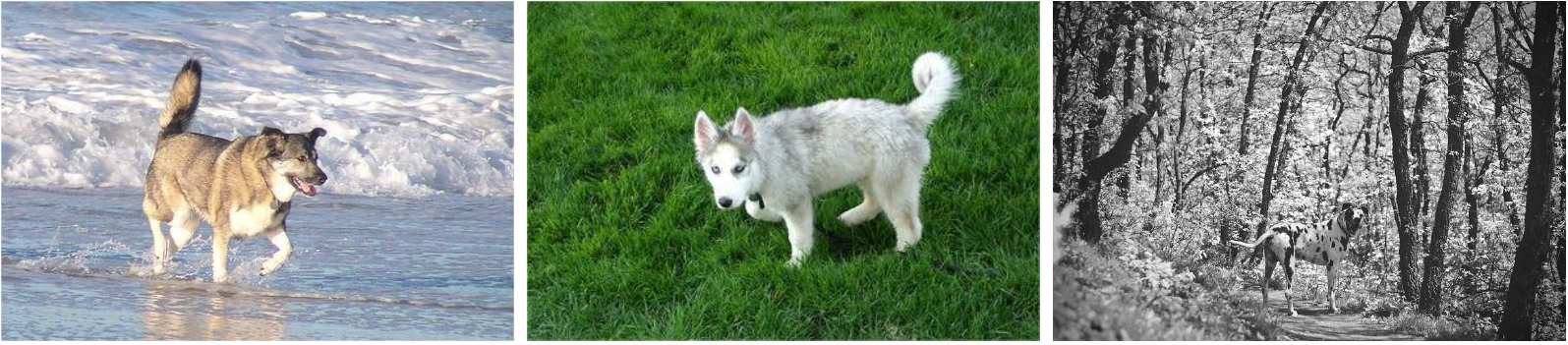}
    \captionof{figure}{Differences in input samples for GOD, SOD, and COD tasks. The three dogs, from left to right, are from the GOD dataset \textit{COCO}~\cite{cocolin2014microsoft}, the SOD dataset \textit{DUTS}~\cite{DUTSwang2017learning}, and the COD dataset \textit{COD10K}~\cite{fan2022concealed}, respectively.}
    \vspace{0.8mm}
    \label{fig:combined3examples}
    \end{center}ing from concealed defect detection\cite{kumar2008computer} in industry and pest monitoring\cite{Wang_2024_CVPR,rustia2020application} in agriculture to lesion segmentation in medical diagnosis\cite{fan2020pranet} and art, such as recreational art\cite{chu2010camouflage} and photo-realistic blending\cite{suo2021neuralhumanfvv}.
        

However, unlike GOD and SOD, COD involves detecting objects that are purposefully designed to be inconspicuous, like the Dalmatian hidden in the forest on the far right of Fig.\ref{fig:combined3examples} which is difficult to detect due to its camouflage with the surroundings, thus requiring more sophisticated detection strategies. COD can be further classified into image and video tasks\cite{he2023strategic,2018A}. Normal COD, \textit{i.e.}, image-level COD, to detect camouflaged objects in static images, whereas video-level COD, dubbed VCOD, deals with detecting these objects within video sequences. The latter one introduces additional complexity due to temporal continuity and dynamic changes, necessitating models capable of extracting both spatial and temporal features effectively.
        
Traditional methods for COD and VCOD, including tex-\vspace{-5mm}
    \begin{center}
    \setlength{\abovecaptionskip}{0.1cm}     
    \includegraphics[width=.49\textwidth]{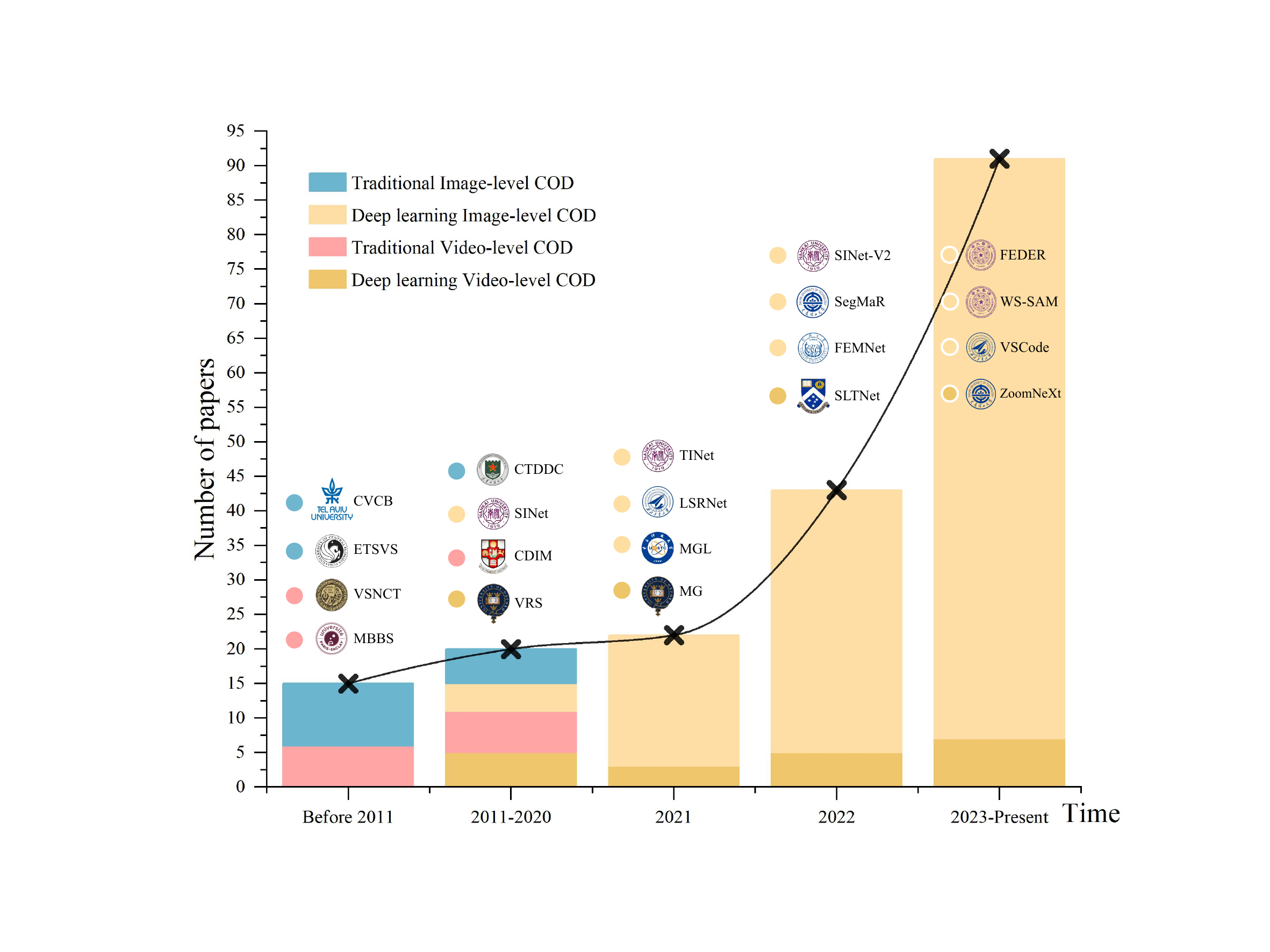}
    \captionof{figure}{The bar chart illustrates the continuous growth of COD methods across four scenarios. Representative works from various periods are categorized and marked on the line graph,
    with different colors corresponding to each scenario.}
    \vspace{0.8mm}
    \label{fig:time-papers}
\end{center}ture\cite{galun2003texture}, intensity\cite{tankus1998detection}, color\cite{2020Camouflaged}, motion\cite{2001Into}, optical flow\cite{2010Optical}, and multi-modal analysis\cite{2004Motion}, have demonstrated their strengths in specific scenarios but also exhibit notable shortcomings. These approaches, relying on manually designed operators, suffer from limited feature extraction capacity, thus struggling with complex backgrounds and varying object appearances, constraining the accuracy and robustness.

In contrast, deep learning-based COD method, \textit{e.g.}, convolutional neural network (CNN), transformer, and diffusion model, offer significant advantages by automatically learning rich feature representations~\cite{He2023Camouflaged,fan2020camouflaged}. In addition, these methods utilize various strategies to address such challenging task, \textit{e.g.}, aggregating multi-scale features~\cite{sun2021c2fnet,he2023weaklysupervised,chen2022cross,Huang2023Feature,Hu_Wang_Qin_Dai_Ren_Luo_Tai_Shao_2023}, bio-inspired mechanism-simulation~\cite{fan2022concealed,yunqiu_cod21,chen2023dual,Mei_2021_CVPR,Jia_2022_CVPR,ZoomNet-CVPR2022,he2023strategic}, fusing multi-source information~\cite{9878807,He2023Camouflaged,chen2024hierarchical,Wang_2024_CVPR,luo2024vscode}, learning multi-task~\cite{ltnghia-CVIU2019,yang2021uncertainty,zhou2022feature,zhang:hal-04142929,9962828}, jointing SOD~\cite{aixuan_cod_sod21,10231131,hao2024simpleeffectivenetworkbased,luo2024vscode}, and setting novel-task~\cite{cheng2023largemodelbasedreferring,zhang2023referring,OVCOS_ECCV2024,zhang2023collaborative,zhang2023unsupervisedcamouflagedobjectsegmentation}. Despite their advantages, these methods also face intractable challenges, including high computational demands\cite{ji2022fast} and the requirement for large annotated, clean, and paired datasets~\cite{he2023weaklysupervised}.

Several surveys have been conducted on COD, with three seminal works \cite{9598866,fan2023csu,LIANG2024127050} providing valuable overviews of the field. However, these surveys have limitations due to the narrow scope and limited number of papers they cover. For example, most of the methods discussed in these surveys are from before the first half of 2023, resulting in insufficient historical depth and domain breadth. As illustrated in Fig.~\ref{fig:time-papers}, the COD field has seen rapid development in 2023. To address these gaps, we propose a more comprehensive survey that not only covers traditional and deep learning COD methods across both image and video domains but also benchmarks deep learning models in these areas. Furthermore, to the best of our knowledge, this survey is the first to deeply explore novel tasks such as referring-based COD \cite{zhang2023referring} and collaborative COD \cite{zhang2023collaborative}. We also provide a broader review of commonly used COD datasets and comprehensively cover recent advancements, challenges, and future trends.

The motivation for this paper stems from the critical importance of COD and the inadequacies of existing surveys. Our survey aims to provide a more thorough and detailed examination of COD, address gaps in the current literature, and highlight recent developments. We systematically categorize and analyze existing cutting-edge techniques, identify critical challenges, and suggest future research directions to advance the field.
        
        Our contributions are summarized as follows:
        \begin{itemize}
            \item We provide a comprehensive review of existing COD methods and related tasks in camouflaged scenario understanding (CSU), along with commonly used datasets and evaluation metrics. To the best of our knowledge, this work represents the most extensive investigation to date, encompassing approximately \textbf{180} CSU-relevant cutting-edge studies.
            \item We methodically benchmark \textbf{40} representative image-level models and \textbf{8} representative video-level models based deep features on \textbf{6} characteristic datasets and \textbf{6} typical evaluation metrics, providing the quantitative and qualitative analysis of them.
            \item We systematically identify the limitations of existing COD methods and propose potential directions for future research. By shedding light on these challenges and opportunities, our work serves to guide and inspire further research efforts to advance the state-of-the-art in COD technology.
            \item We create a \href{https://github.com/ChunmingHe/awesome-concealed-object-segmentation}{repository} that houses a carefully curated collection of COD methods, datasets, and relevant resources, which will be consistently updated to ensure the latest information is accessible.
        \end{itemize}
    
        We hope that this survey on COD will not only enhance understanding of the field but also stimulate greater interest within the computer vision community, fostering further research initiatives in related areas.

        \begin{table*}[!htb]
\centering
\caption{\textcolor{red}{14} representative traditional methods of image-level COD. For more details, please refer to Section \ref{section:image-tradition}}
\label{tab:image-tradition}
\resizebox{\textwidth}{!}{%
\begin{tabular}{crcccc}
\toprule[1.5pt]
\textit{\textbf{Index}} &
  \textit{\textbf{Method}} &
  \textit{\textbf{Pub.}} &
  \textit{\textbf{Year}} &
  \textit{\textbf{Feature}} &
  \textit{\textbf{Core component}} \\ 
  \midrule[1pt]
1 &
  CBDCE\cite{tankus1998detection} &
  $ Vis. Surveill. $ &
  1998 &
  Intensity &
  Distinguishing the convex and concave, D-Arg operator \\
2 &
  CVCB\cite{tankus2001convexity} &
  $ CVIU $ &
  2001 &
  Intensity &
  Strengthened D-Arg operator \\
3 &
  TSMA\cite{galun2003texture} &
  $ ICCV $ &
  2003 &
  Texture &
  Bottom-up integration of structural features, Filtered response \\
4 & 
  ETSVS\cite{neider2006searching} &
  $ Vis. Res. $ &
  2006 &
  Texture &
  Four volunteer experiments \\
5 &
  CDIA\cite{bhajantri2006camouflage} &
  $ ICIT $ &
  2006 &
  Texture &
  Co-occurrence matrix, Watershed technology, Cluster analysis \\
6 &
  DTEIA\cite{sengottuvelan2008performance} &
  $ IConETech $ &
  2008 &
  Texture &
  Gray-level co-occurrence matrix, Dendrogram plot 
  \\
7 &
  TTSCT\cite{Walter2009Training} &
  $ AP\&P $ &
  2009 &
  Color &
  Volunteer experiment based on the color difference \\
8 &
  CYEM\cite{song2010new} &
  $ ICMCT $ &
  2010 &
  Texture &
  Weight structural similarity, evaluation method\\
9 &
  ROHS\cite{2010Robust} &
  $ ICGCS $ &
  2010 &
  \begin{tabular}[t]{@{}c@{}}Color \& \\Intensity\end{tabular} &
  Background subtraction, Color statistics, Edge information \\ 
10 &
  IRCT\cite{2011AN} &
  $ IJEST $ &
  2011 &
  \begin{tabular}[t]{@{}c@{}}Color \& \\Texture\end{tabular} &
  HSV 
  color, gray level co-occurrence matrix \\ 
11 &
  CTDDC\cite{pan2011study} &
  $ Mod. Appl. Sci. $ &
  2011 &
  Intensity &
  Gray-level image, 3D convexity \\ 
12 &
  CTESM\cite{feng2015camouflage} &
  $ Multimed Syst. $ &
  2015 &
  Texture &
  Saliency map, Evaluation method \\ 
13 &
  TGWV\cite{2017Foreground} &
  $ ICIP $ &
  2017 &
  \begin{tabular}[t]{@{}c@{}}Texture \& \\Intensity\end{tabular} &
  Texture-guided 
  voting, Stationary wavelet transform \\ 
14 &
  CODMVA\cite{2020Camouflaged} &
  \begin{tabular}[t]{@{}c@{}}$ Int. J. $\\ $ Speech Technol. $ \end{tabular} &
  2020 &
  \begin{tabular}[t]{@{}c@{}}Texture \& \\Intensity\end{tabular} &
  Characterizing entity texture, Statistical modeling \\ 
  \bottomrule[1.5pt]
\end{tabular}%
}
\end{table*}
        
        \noindent \textbf{Note}. In developing our search strategy, we conducted a thorough investigation across a variety of databases, including DBLP, Google Scholar, and ArXiv Sanity Preserver. Our focus is particularly directed toward reputable sources, such as TPAMI and IJCV, as well as prominent conferences like CVPR, ICCV, and ECCV. We prioritized studies that provided official codes to enhance reproducibility, as well as those with higher citations and Github stars, indicative of significant recognition and adoption within the academic community. Following this initial screening, our literature selection process involved a rigorous evaluation of each paper's novelty, contribution, and significance, and an assessment of its status as seminal work in the field. While we acknowledge the possibility of omitting some noteworthy papers, our aim is to present a comprehensive overview of the most influential and impactful research, promoting research advancement and suggesting potential future trends and directions.

	\section{Image-level COD models}

        Image-level COD refers to the process of identifying and distinguishing objects that are designed to blend into their surroundings 8 within static images, garnering significant attention currently. 
        In this section, we categorize these methods into two primary approaches based on the features they utilize: traditional COD methods and deep learning COD methods. Traditional approaches typically rely on handcrafted features, whereas deep learning methods leverage neural networks to automatically learn and extract discriminative features from data. Given the rapid advancements in technology, we will focus primarily on deep learning COD methods, which have recently become the predominant approach.
        
        \subsection{Traditional COD methods} 
        
        \label{section:image-tradition}
        COD is initially rooted in traditional image-level methods, leveraging hand-crafted low-level features tailored to capture nuances in textures, intensities, and colors. These approaches constituted the foundation of early efforts in this domain.  Tab.~\ref{tab:image-tradition} summarizes the relevant models and characteristics of these methods.
        
        \noindent\textbf{Texture feature-based approaches.}
        Texture features capture the surface properties and distinctive patterns of images, usually manifested through grayscale distributions across pixels and their spatial neighborhoods. These features are utilized to distinguish camouflaged objects from their backgrounds based on texture differences.
        
        Galun \textit{et al.}\cite{galun2003texture} introduce a bottom-up approach, TSMA, for texture segmentation. This method leverages the adaptive identification and characterization of texture elements--such as size, aspect ratio, orientation, and brightness--combined with filter response statistics for enhanced differentiation and noise reduction. 
        Unlike TSMA, which extracts global texture features, Bhajantri \textit{et al.}\cite{bhajantri2006camouflage} employed a co-occurrence matrix for block-based local texture analysis. Their method uses Watershed segmentation for each defective block, processed through a dendrogram to distinguish the target, and concludes with cluster analysis to further confirm detection results.
        
        Building upon earlier works, Sengottuvelan \textit{et al.}\cite{sengottuvelan2008performance} propose an unsupervised technique for decamouflaging that converts images to grayscale and then splits them into blocks for gray-level co-occurrence matrix analysis, capturing pixel-neighbor relationships. They use a dendrogram plot to identify camouflaged objects without requiring prior background knowledge. 
        Traditional evaluation methods often rely on subjective assessments, which can be cumbersome and ambiguous processes. To address it, Song \textit{et al.}\cite{song2010new} leverage a weighted structural similarity index and intrinsic image feature analysis to assess camouflage textures. To be more objective, Feng \textit{et al.}\cite{feng2015camouflage} utilize human visual-based saliency maps to quantitatively assess texture differences. 

        \begin{figure*}[ht]
        {\includegraphics[width=\textwidth]{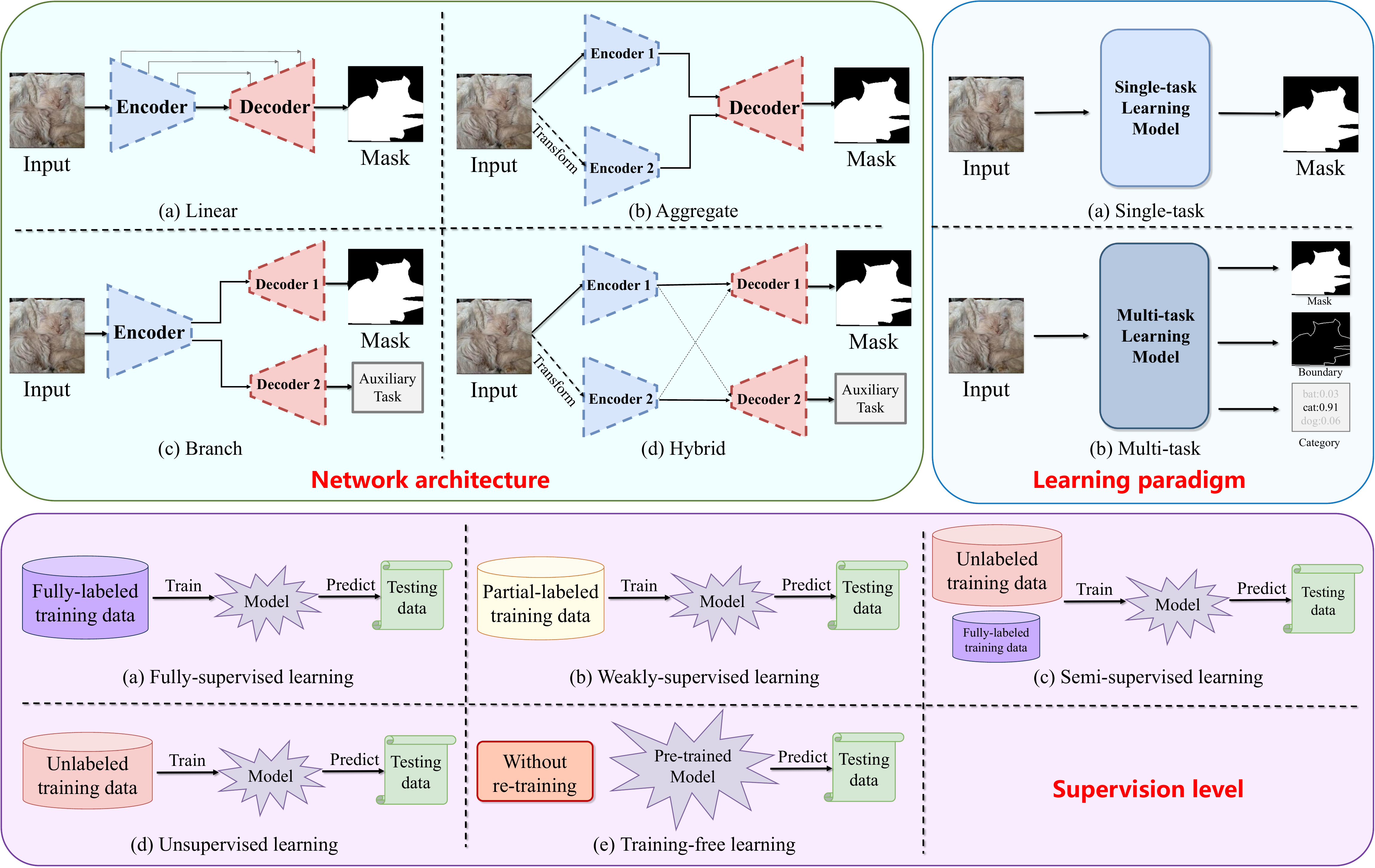}}
        \caption{Network architecture, learning paradigm, and supervision level for deep learning COD methods. 
        } 
        \vspace{-1mm}
        \label{fig:three-types-deep-cod}
        \end{figure*}
        
        \noindent\textbf{Intensity feature-based approaches.}
        These methods progressively advance from basic intensity-based techniques to more sophisticated exploitation of 3D convexity.
        CBDCE~\cite{tankus1998detection} detected regions of interest by processing intensity images directly, distinguishing between 3D convex and concave regions. This approach demonstrates robustness against variations in illumination, orientation, and scale. 
        To enhance the detection of 3D objects camouflaged in complex scenes, CVCB\cite{tankus2001convexity} enhanced the D-Arg operator in CBDCE. It maximizes the response to curved 3D objects against flat backgrounds, effectively mitigating visual camouflage in both natural and artificial environments.
        CTDDC\cite{pan2011study} employed another 3D convexity-based operator that exploits image gray levels and median filtering to eliminate background noise, enabling effective and robust detection.
        
        \noindent\textbf{Color feature-based approaches.} In certain scenarios, color contrast and distribution can provide significant distinctiveness for separating camouflaged objects from their surroundings.
        Siricharoen \textit{et al.}\cite{2010Robust} optimized a statistical background subtraction and shadow detection algorithm by integrating color, edge, and intensity features for outdoor human segmentation, where strong shadows and low contrast are common. They employ vector median filtering to remove outlier pixels and combine color statistics with edge to generate initial coarse results, which are then refined using intensity features for accuracy.
        Kavitha \textit{et al.}\cite{2011AN} propose an image retrieval technique for COD, which involves segmenting images into blocks and extracting Hue, Saturation, Value (HSV) color, and gray-level co-occurrence matrix texture features from each block. They use the principle of matching images based on similarity and Euclidean distance to improve detection accuracy.

        Handcrafted low-level features are specifically designed to be highly discriminative, making them effective in detecting and segmenting objects. By accentuating differences in texture and color, these features help isolate concealed objects. However, camouflage seeks to minimize these distinctions, reducing visibility and blending objects into the background. Hence, traditional methods fail in COD, succeeding only in simple scenes with uniform backgrounds. These methods tend to underperform with low-resolution images or in cases where the foreground and background share significant visual similarities.

        \subsection{Deep learning COD methods}
        \label{section:image-deep}

        While traditional methods rely on hand-crafted low-level features to capture key attributes of an image, deep learning methods extract complex and deep features directly from the data through automatic learning representations, demonstrating superior performance across various computer vision tasks. According to the existing surveys~\cite{fan2023csu,LIANG2024127050}, deep learning methods for COD can be broadly categorized based on three fundamental criteria: network architecture, learning paradigm, and supervision level. A detailed description of the three criteria is shown in Fig. \ref{fig:three-types-deep-cod}. What's more, Tab.~\ref{tab:image-deep-1922} and Tab.~\ref{tab:image-deep-2324} outline the key characteristics of a total of \textbf{104} representative methods for image-level COD, published in 2019-2022 and 2023\&2024, respectively. 
        
        Network architecture delineates how the input and output configurations are structured within the models. The linear~\cite{fan2020camouflaged,liu2021integrating,Mei_2021_CVPR,sun2021c2fnet,he2019image,9430677,9606888,DBLP:journals/corr/abs-2101-04704,dong2021accuratecamouflagedobjectdetection,zhang2022learning,yin2022camoformermaskedseparableattention,sun2022dqnetcrossmodelqueryingcamouflaged,9955520,ZHANG2022103450,9859854,chen2022camouflaged,9815160,xu2022multi,chen2023recursive,Jia_2022_CVPR,10.1145/3545609,fan2022concealed,10379651,li2023crosslevelattentionoverlappedwindows,zhang2023unsupervisedcamouflagedobjectsegmentation,dong2023needadditionalpriorscamouflage,xu2023dm,chen2023diffusionmodelcamouflagedobject,10180211,OPNet,10219858,10095226,10219627,LIU2023109514,he2023weaklysupervised,chen2023camodiffusioncamouflagedobjectdetection,DBFN,10103836,10065514,10045692,Huang2023Feature,rs15051188,S-COD,Hu_Wang_Qin_Dai_Ren_Luo_Tai_Shao_2023,xiao2023concealed,10417767,chen2024greencodgreencamouflagedobject,chen2024adaptiveguidancelearningcamouflaged,guo2024cofinetunveilingcamouflagedobjects,10483928} employs bottom-up/top-down network, where the data flows in a single feed-forward pass. Aggregative architecture~\cite{Jinnan-IEEEAccess2021,9878807,ZoomNet-CVPR2022,tang2023generalizationhallucinationlargevisionlanguage,10.1145/3581783.3611874,he2023reti-LLIE3,cheng2023largemodelbasedreferring,zhang2024cross,zhang2023referring,pang2024zoomnextunifiedcollaborativepyramid,cong2023frequencyperceptionnetworkcamouflaged,song2023camouflagedobjectdetectionfeature,he2023degradation,wu2023object,10377562,Zheng_2023_WACV,Wang_2024_CVPR,hu2023relax} combines features from multiple input streams, while branched architecture~\cite{ltnghia-CVIU2019,kajiura2021improving,yunqiu_cod21,9577564,zhang2021perturbed,xiang2022exploringdepthcontributioncamouflaged,Zhu_Zhang_Zhang_Liu_2021,yang2021uncertainty,zhou2022feature,9923635,10.1145/3503161.3548178,sun2022bgnet,ZHUGE2022108644,Liu_2022_WACV,Zhu_Li_Xie_Yan_Liang_Chen_Wei_Qin_2022,ji2022fast,10.1145/3581783.3611773,10262011,10234216,ijcai2023p124,ju2022ivf,ZHANG2023103719,sun2023edgeawaremirrornetworkcamouflaged,10183371,zhang:hal-04142929,chen2023bioinspiredthreestagemodelcamouflaged,He2023Camouflaged,9962828,he2023hqg,ji2023gradient,Lv2022TowardDU,liu2023camouflaged,10448139,deng2022pcgan,nguyen2024art,hao2024simpleeffectivenetworkbased} is characterized by multiple pathways for multiple outputs, associated with the multi-task learning paradigm. The hybrid~\cite{aixuan_cod_sod21,OVCOS_ECCV2024,tang2023source,zhang2023collaborative,10231131,10224812,tang2023consistency,li2023jointsalientobjectdetection,tang2024mind,huang2024evolver,xing2023pretrainadaptdetectmultitask,luo2024vscode,he2023strategic} integrates elements and strategies from the aforementioned ones to leverage their strengths. 
        
        The learning paradigm pertains to the approach adopted by the models to learn and adapt. This includes single-task and multi-task learning, specifically the former only involving COD, while the latter often importing auxiliary tasks, \textit{e.g.}, localization/ranking~\cite{yunqiu_cod21,Lv2022TowardDU}, reconstruction~\cite{yunqiu_cod21,Lv2022TowardDU} and predicting associated cues, such as boundary~\cite{9577564,zhou2022feature,9923635,sun2022bgnet,ZHUGE2022108644,Zhu_Li_Xie_Yan_Liang_Chen_Wei_Qin_2022,ji2022fast,10262011,10231131,10224812,ijcai2023p124,ZHANG2023103719,sun2023edgeawaremirrornetworkcamouflaged,10183371,chen2023bioinspiredthreestagemodelcamouflaged,He2023Camouflaged,9962828,liu2023camouflaged,10448139,chen2024adaptiveguidancelearningcamouflaged,he2023strategic}, texture~\cite{Zhu_Zhang_Zhang_Liu_2021,9923635,ji2023gradient,chen2024adaptiveguidancelearningcamouflaged} and uncertainty~\cite{kajiura2021improving,aixuan_cod_sod21,xiang2022exploringdepthcontributioncamouflaged,yang2021uncertainty,10.1145/3503161.3548178,Liu_2022_WACV,10183371,li2023jointsalientobjectdetection,zhang:hal-04142929}, leading to improved accuracy and performance. 

\begin{table*}[!htb]
\centering
\caption{\textcolor{red}{36} representative deep learning methods of image-level COD published from 2019 to 2022. \textcolor{red}{N.A.}: network architecture according to input-output configuration, including \textcolor{blue}{L} (Linear), \textcolor{blue}{A} (Aggregative), \textcolor{blue}{B} (Branched), and \textcolor{blue}{H} (Hybrid). \textcolor{red}{L.P.}: learning paradigm, including \textcolor{blue}{S} (Single-task) and \textcolor{blue}{M} (Multi-task). \textcolor{red}{S.L.}: supervision level, including \textcolor{blue}{F} (Full-supervision), \textcolor{blue}{W} (Weak-supervision), \textcolor{blue}{S} (Semi-supervision), \textcolor{blue}{U} (Unsupervision) and \textcolor{blue}{TF} (Training-free).
Clicking on some method names may redirect you to their open-source codes or projects.
For more details, please refer to Section \ref{section:image-deep}.}
\label{tab:image-deep-1922}
\resizebox{\textwidth}{!}{%
\begin{tabular}{crcccccc}
\toprule[1.5pt]
\textit{\textbf{Index}} &
  \textit{\textbf{Method}} &
  \textit{\textbf{Pub./Year}} &
  \textit{\textbf{Backbone}} &
  \textit{\textbf{N.A.}} &
  \textit{\textbf{L.P.}} &
  \textit{\textbf{S.L.}} &
  \textit{\textbf{Core component}} 
  \\
\midrule[1pt]

1 &
  ~\href{https://github.com/ltnghia/ANet}{ANet}~\cite{ltnghia-CVIU2019} &
  $ CVIU_{19} $ &
  DHS, DSS, SRM, WSS &
  B &
  M &
  F &
  Classification \& segmentation streams
  \\[0.02cm]
2 &
  ~\href{https://github.com/DengPingFan/SINet}{SINet}~\cite{fan2020camouflaged} &
  $ CVPR_{20} $ &
  ResNet50 &
  L &
  S &
  F &
  Search \& identification
  
   \\[0.02cm]
3 &
  TINet~\cite{Zhu_Zhang_Zhang_Liu_2021} &
  $ AAAI_{21} $ &
  ResNet50 &
  B &
  M &
  F &
  \begin{tabular}[t]{@{}c@{}}Feature interaction, texture \& holistic perception\end{tabular}  \\[0.02cm]
4 &
  ~\href{https://github.com/nobukatsu-kajiura/UR-COD}{UR-COD}~\cite{kajiura2021improving} &
  $ ACM MM_{21} $ &
  ResNet50, ResNet34 &
  B &
  M &
  F &
  \begin{tabular}[t]{@{}c@{}}Pseudo-edge \& pseudo-map generators,\\ uncertainty-aware refinement\end{tabular} 
   \\[0.02cm]
5 &
  ~\href{https://github.com/JingZhang617/Joint\_COD\_SOD}{UJSCOD}~\cite{aixuan_cod_sod21} &
  $ CVPR_{21} $ &
  ResNet50 &
  H &
  M &
  F &
  \begin{tabular}[t]{@{}c@{}}Uncertainty-aware adversarial learning,\\ joint SOD \& COD, confidence estimation\end{tabular} 
   \\[0.02cm]
6 &
  ~\href{https://github.com/JingZhang617/COD-Rank-Localize-and-Segment}{LSR}~\cite{yunqiu_cod21} &
  $ CVPR_{21} $ &
  ResNet50 &
  B &
  M &
  F &
  Simultaneously localization, segmentation \& ranking 
  \\[0.02cm]
7 &
  ~\href{https://mhaiyang.github.io/CVPR2021_PFNet/index.html}{PFNet}~\cite{Mei_2021_CVPR} &
  $ CVPR_{21} $ &
  ResNet50 &
  L &
  S &
  F &
  Distraction mining strategy, positioning \& focus 
   \\[0.02cm]
8 &
  ~\href{https://github.com/fanyang587/MGL}{MGL}~\cite{9577564} &
  $ CVPR_{21} $ &
  ResNet50 &
  B &
  M &
  F &
  Mutual Graph Learning 
   \\[0.02cm]
9 &
  ~\href{https://github.com/fanyang587/UGTR}{UGTR}~\cite{yang2021uncertainty} &
  $ ICCV_{21} $ &
  ResNet50 &
  B &
  M &
  F &
  \begin{tabular}[t]{@{}c@{}}Uncertainty quantification network with Bayesian learning,\\ prototyping \& uncertainty-guided transformers\end{tabular} 
   \\[0.02cm]
10 &
  ~\href{https://sites.google.com/view/ltnghia/research/camo}{MirrorNet}~\cite{Jinnan-IEEEAccess2021} &
  $ IEEE Access_{21} $ &
  \begin{tabular}[t]{@{}c@{}}ResNet50, ResNet101
  \end{tabular} &
  H &
  M &
  F &
  Main \& mirror streams embedded image flipping 
  \\[0.02cm]
11 &
  ~\href{https://github.com/thograce/C2FNet}{C$^{2}$F-Net}~\cite{sun2021c2fnet} &
  $ IJCAI_{21} $ &
  Res2Net50 &
  L &
  S &
  F &
  \begin{tabular}[t]{@{}c@{}}Attention-induced cross-level fusion \&\\ dual-branch global context modules\end{tabular} 
   \\[0.02cm]
12 &
  TANet~\cite{9606888} &
  $ TCSVT_{21} $ &
  ResNeXt50 &
  L &
  S &
  F &
  \begin{tabular}[t]{@{}c@{}}Residual refinement, texture-aware refinement,\\ boundary-consistency loss\end{tabular} 
  \\[0.02cm]
13 &
  ~\href{https://github.com/MS-KangWang/COD-D2Net}{D$^{2}$C-Net}~\cite{9430677} &
  $ TIE_{21} $ &
  Res2Net50 &
  L &
  S &
  F &
  \begin{tabular}[t]{@{}c@{}}Dual-branch features extraction \&\\ gradually refined cross fusion\end{tabular} 
  \\[0.02cm]
14 &
  ~\href{https://github.com/liuyi1989/POCINet}{POCINet}~\cite{liu2021integrating} &
  $ TIFS_{21} $ &
  VGG16 &
  L &
  S &
  F &
  \begin{tabular}[t]{@{}c@{}}Search \& identification,\\ part-object relationship \& contrast integrated\end{tabular} 
   \\[0.02cm]
15 &
  DCE~\cite{xiang2022exploringdepthcontributioncamouflaged} &
  $ arXiv_{21} $ &
  ResNet50 &
  B &
  M &
  F &
  \begin{tabular}[t]{@{}c@{}}Depth contribution exploration,\\multi-modal confidence-aware loss\end{tabular} 
  \\[0.02cm]
16 &
  ~\href{https://github.com/xuebinqin/BASNet}{BAS}~\cite{DBLP:journals/corr/abs-2101-04704} &
  $ arXiv_{21} $ &
  ResNet34 &
  L &
  S &
  F &
  Residual refinement, hybrid loss 
   \\[0.02cm]
17 &
  ~\href{https://github.com/BigHeartDB/MCIFNet}{MCIF-Net}~\cite{dong2021accuratecamouflagedobjectdetection} &
  $ arXiv_{21} $ &
  ResNet50 &
  L &
  S &
  F &
  \begin{tabular}[t]{@{}c@{}}Dual-branch mixture convolution \&\\ multi-level interactive fusion\end{tabular} 
   \\[0.02cm]
18 &
  ~\href{https://github.com/zhuhongwei1999/BSA-Net}{BSA-Net}~\cite{Zhu_Li_Xie_Yan_Liang_Chen_Wei_Qin_2022} &
  $ AAAI_{22} $ &
  Res2Net &
  B &
  M &
  F &
  Boundary guider, reverse and normal attention streams 
   \\[0.02cm]
19 &
  ~\href{https://github.com/sxu1997/PreyNet}{PreyNet}~\cite{10.1145/3503161.3548178} &
  $ ACM MM_{22} $ &
  ResNet50 &
  B &
  M &
  F &
  \begin{tabular}[t]{@{}c@{}}Initial detection and predator learning,\\ bidirectional bridging interaction, uncertainty estimation\end{tabular} 
   \\[0.02cm]
20 &
  NCHIT~\cite{ZHANG2022103450} &
  $ CVIU_{22} $ &
  ResNet50 &
  L &
  S &
  F &
  Neighbor connection mode, hierarchical information transfer 
  \\[0.02cm]
21 &
  FEMNet~\cite{9878807} &
  $ CVPR_{22} $ &
  Res2Net, ResNet50 &
  A &
  S &
  F &
  \begin{tabular}[t]{@{}c@{}}Frequency enhancement \& high-order relation modules,\\ frequency perception loss\end{tabular} 
  \\[0.02cm]
22 &
  ~\href{https://github.com/dlut-dimt/SegMaR}{SegMaR}~\cite{Jia_2022_CVPR} &
  $ CVPR_{22} $ &
  ResNet50 &
  L &
  S &
  F &
  Iteratively segmenting \& magnifying, discriminative mask 
   \\[0.02cm]
23 &
  ~\href{https://github.com/lartpang/ZoomNet}{ZoomNet}~\cite{ZoomNet-CVPR2022} &
  $ CVPR_{22} $ &
  ResNet50 &
  A &
  S &
  F &
  \begin{tabular}[t]{@{}c@{}}Scale integration \& hierarchical mixed-scale units,\\ uncertainty-aware loss\end{tabular} 
   \\[0.02cm]
24 &
  ~\href{https://github.com/michael861227/PINet}{PINet}~\cite{9859854} &
  $ ICME_{22} $ &
  ResNet50 &
  L &
  S &
  F &
  \begin{tabular}[t]{@{}c@{}}Cascaded decamouflage,\\ classification-based label reweighting\end{tabular} 
   \\[0.02cm]
25 &
TCU-Net~\cite{9955520} &
  $ IEEE Access_{22} $ &
  SwinT &
  L &
  S &
  F &
  \begin{tabular}[t]{@{}c@{}}Transformer and CNN-based U-Net, weighted hybrid loss,\\ multi-dilated residual blocks, attention inception decoder\end{tabular} 
  \\[0.02cm]
26 &
~\href{https://github.com/thograce/BGNet}{BGNet}~\cite{sun2022bgnet} &
  $ IJCAI_{22} $ &
  Res2Net50 &
  B &
  M &
  F &
  \begin{tabular}[t]{@{}c@{}}Edge-aware, edge-guidance feature \&\\ context aggregation modules\end{tabular} 
   \\[0.02cm]
27 &
CubeNet~\cite{ZHUGE2022108644} &
  $ PR_{22} $ &
  ResNet50 &
  B &
  M &
  F &
  Square fusion \& sub edge decoders, X-connection 
  \\[0.02cm]
28 &
~\href{https://github.com/GewelsJI/ERRNet}{ERRNet}~\cite{ji2022fast} &
  $ PR_{22} $ &
  ResNet50 &
  B &
  M &
  F &
  \begin{tabular}[t]{@{}c@{}}Selective edge aggregation,\\ reversible re-calibration, NGES Priors\end{tabular} 
   \\[0.02cm]
29 &
~\href{https://github.com/Ben57882/C2FNet-TSCVT}{C$^{2}$F-Net-V2}~\cite{chen2022camouflaged} &
  $ TCSVT_{22} $ &
  Res2Net50 &
  L &
  S &
  F &
  \begin{tabular}[t]{@{}c@{}}Attention-induced cross-level fusion \&\\ dual-branch global context modules,~\cite{sun2021c2fnet} extention\end{tabular} 
  \\[0.02cm]
30 &
~\href{https://github.com/taozh2017/FAPNet}{FAP-Net}~\cite{zhou2022feature} &
  $ TIP_{22} $ &
  Res2Net50 &
  B &
  M &
  F &
  \begin{tabular}[t]{@{}c@{}}Boundary guidance, multi-scale feature aggregation,\\ cross-level fusion and propagation\end{tabular}
   \\[0.02cm]
31 &
FindNet~\cite{9923635} &
  $ TIP_{22} $ &
  Res2Net &
  B &
  M &
  F &
  Boundary \& texture enhancement modules 
  \\[0.02cm]
32 &
DTC-Net~\cite{9815160} &
  $ TMM_{22} $ &
  ResNet50 &
  L &
  S &
  F &
  Local bilinear, spatial coherence organization 
  \\[0.02cm]
33 &
FBNet~\cite{10.1145/3545609} &
  $ TOMM_{22} $ &
  ResNet50 &
  L &
  S &
  F &
  \begin{tabular}[t]{@{}c@{}}Frequency-aware context aggregation and\\  adaptive frequency attention
  \end{tabular} 
  \\[0.02cm]
34 &
~\href{https://github.com/GewelsJI/SINet-V2}{SINet-V2} ~\cite{fan2022concealed} &
  $ TPAMI_{22} $ &
  Res2Net50 &
  L &
  S &
  F &
  \begin{tabular}[t]{@{}c@{}}Search \& identification, neighbor connection decoder,\\group-reversal attention,~\cite{fan2020camouflaged} extention\end{tabular} 
   \\[0.02cm]
35 &
~\href{https://github.com/Carlisle-Liu/OCENet}{OCENet}~\cite{Liu_2022_WACV} &
  $ WACV_{22} $ &
  ResNet50 &
  B &
  M &
  F &
  \begin{tabular}[t]{@{}c@{}}Dynamic confidence supervision,\\ explicitly aleatoric uncertainty estimation\end{tabular} 
   \\[0.02cm]
36 &
~\href{https://github.com/CVPR23/DQnet}{DQnet}~\cite{sun2022dqnetcrossmodelqueryingcamouflaged} &
  $ arXiv_{22} $ &
  ResNet50, ViTB &
  L &
  S &
  F &
  Cross-model detail querying, relation-based querying 
   \\[0.02cm]
  \bottomrule[1.5pt]
\end{tabular}%
}
\end{table*}

        
\begin{table*}[!htb]
\centering
\caption{\textcolor{red}{68} representative deep learning methods of image-level COD published in 2023\&2024. The same caption as Tab.~\ref{tab:image-deep-1922}}
\label{tab:image-deep-2324}
\resizebox{\textwidth}{!}{%
\begin{tabular}{crcccccc}
\toprule[1.5pt]

  \textit{\textbf{Index}} &
  \textit{\textbf{Method}} &
  \textit{\textbf{Pub./Year}} &
  \textit{\textbf{Backbone}} &
  \textit{\textbf{N.A.}} &
  \textit{\textbf{L.P.}} &
  \textit{\textbf{S.L.}} &
  \textit{\textbf{Core component}} 
  \\ 
  
\midrule[1pt]

1 &
~\href{https://github.com/HUuxiaobin/HitNet}{HitNet}~\cite{Hu_Wang_Qin_Dai_Ren_Luo_Tai_Shao_2023} &
  $ AAAI_{23} $ &
  PVTv2 &
  L &
  S &
  F &
  High-resolution iterative feedback, iterative feedback loss 
  \\[0.01cm] 
2 &
~\href{https://github.com/dddraxxx/Weakly-Supervised-Camouflaged-Object-Detection-with-Scribble-Annotations}{CRNet}~\cite{S-COD} &
  $ AAAI_{23} $ & 
  ResNet50 &
  L &
  S &
  W &
  consistency loss \& feature-guided loss 
  \\[0.01cm] 
3 &
~\href{https://github.com/iCVTEAM/FRINet}{FRINet}~\cite{10.1145/3581783.3611773} &
  $ ACM MM_{23} $ & 
  ViT &
  B &
  M &
  F &
  Frequency representation integration \& reasoning 
  \\[0.01cm] 
4 &
~\href{https://github.com/qingwei-wang/DaCOD}{DaCOD}~\cite{10.1145/3581783.3611874} &
  $ ACM MM_{23} $ & 
  SwinL, ResNet50 &
  A &
  S &
  F &
  Multi-modal collaborative learning, cross-modal asymmetric fusion 
  \\[0.01cm] 
5 &
~\href{https://github.com/rmcong/FPNet\_ACMMM23}{FPNet} \cite{cong2023frequencyperceptionnetworkcamouflaged} &
  $ ACM MM_{23} $ & 
  PVT &
  A &
  S &
  F &
  Frequency-guided positioning, detail-preserving fine localization 
  \\[0.01cm] 
6 &
~\href{https://github.com/Zongwei97/XMSNet}{XMSNet}\cite{wu2023object} &
  $ ACM MM_{23} $ &  
  PVT &
  A &
  S &
  F &
  \begin{tabular}[t]{@{}c@{}}All-round attentive fusion,\\coarse-to-fine decoder, cross-layer self-supervision\end{tabular}
   \\[0.01cm] 
7 &
HCM~\cite{xiao2023concealed} &
  $ CAAI_{23} $ &
  ResNet50 &
  L &
  S &
  F &
  Hierarchical coherence modeling, reversible re-calibration decoder 
  \\[0.01cm] 
8 &
~\href{https://github.com/zhangqiao970914/ASBI}{ASBI}~\cite{ZHANG2023103719} &
  $ CVIU_{23} $ & 
  \begin{tabular}[t]{@{}c@{}}ResNet50, Res2Net50,\\ PVTv2, EfficientNet\end{tabular} &
  B &
  M &
  F &
  Attention-induced semantic \& boundary interaction 
  \\[0.01cm] 
9 &
~\href{https://github.com/ZhouHuang23/FSPNet}{FSPNet}~\cite{Huang2023Feature} &
  $ CVPR_{23} $ & 
  ViT &
  L &
  S &
  F &
  Nonlocal token enhancement, a feature shrinkage decoder 
  \\[0.01cm] 
10 &
~\href{https://github.com/ChunmingHe/FEDER}{FEDER}~\cite{He2023Camouflaged} &
  $ CVPR_{23} $ & 
  ResNet50, Res2Net50 &
  B &
  M &
  F &
  Deep wavelet-like decomposition, ODE-inspired edge reconstruction 
  \\[0.01cm]
11 &
Semi-SINet~\cite{liu2023camouflaged} &
  $ Def. Technol._{23} $ &
  ResNet50 &
  B &
  M &
  S &
  Search \& identification, edge attention, semi-supervised 
  \\[0.01cm] 
12 &
~\href{https://github.com/ZNan-Chen/diffCOD}{diffCOD}\cite{chen2023diffusionmodelcamouflagedobject} &
  $ ECAI_{23} $ & 
  ViT &
  L &
  S &
  F &
  Denoising diffusion, injection attention module 
  \\[0.01cm] 
13 &
~\href{https://github.com/Haozhe-Xing/TinyCOD}{TinyCOD}~\cite{10095226} &
  $ ICASSP_{23} $ & 
  TinyNet-a &
  L &
  S &
  F &
  Adjacent scale features fusion, edge area focus 
  \\[0.01cm] 
14 &
~\href{https://github.com/Zongwei97/PopNet}{PopNet}~\cite{10377562} &
  $ ICCV_{23} $ & 
  --- &
  A &
  S &
  F &
  \begin{tabular}[t]{@{}c@{}}Source-free depth,  object popping,\\segmentating contact surface, object separation\end{tabular}
   \\[0.01cm] 
15 &
~\href{https://github.com/Jun-Pu/UCOS-DA}{UCOS-DA}\cite{zhang2023unsupervisedcamouflagedobjectsegmentation} &
  $ ICCVW_{23} $ & 
  DINO &
  L &
  S &
  U &
  \begin{tabular}[t]{@{}c@{}}Source-free unsupervised domain adaptation task,\\ foreground-background contrastive self-adversarial\end{tabular} 
  \\[0.01cm] 
16 &
~\href{https://github.com/sdy1999/EAMNet}{EAMNet}~\cite{sun2023edgeawaremirrornetworkcamouflaged} &
  $ ICME_{23} $ & 
  Res2Net &
  B &
  M &
  F &
  \begin{tabular}[t]{@{}c@{}}Segmentation-induced edge aggregation, edge-induced\\ integrity aggregation, guided-residual channel attention\end{tabular} 
  \\[0.01cm] 
17 &
~\href{https://github.com/syxvision/FDNet}{FDNet}~\cite{song2023camouflagedobjectdetectionfeature} &
  $ ICME_{23} $ & 
  PVT &
  A &
  S &
  F &
  Feature grafting \& distractor aware 
  \\[0.01cm] 
18 &
~\href{https://github.com/ZhangQing0329/CFANet}{CFANet}~\cite{10219858} &
  $ ICME_{23} $ & 
  Res2Net50 &
  L &
  S &
  F &
  Cross-layer feature fusion, uniqueness enhancement strategy 
  \\[0.01cm] 
19 &
~\href{https://github.com/xinyang920/OAFormer}{OAFormer}~\cite{10219627} &
  $ ICME_{23} $ &  
  PVTv2 &
  L &
  S &
  F &
  Hierarchical location guidance, neighborhood searching 
  \\[0.01cm] 
20 &
PENet~\cite{ijcai2023p124} &
  $ IJCAI_{23} $ & 
  Res2Net50 &
  B &
  M &
  F &
  Locatint, refining \& restoring 
  \\[0.01cm] 
21 &
OPNet~\cite{OPNet} &
  $ IJCV_{23} $ & 
  Conformer-B &
  L &
  S &
  F &
  Pyramid positioning, dual focus 
  \\[0.01cm] 
22 &
~\href{https://github.com/GewelsJI/DGNet}{DGNet}~\cite{ji2023gradient} &
  $ MIR_{23} $ & 
  EfficientNet &
  B &
  M &
  F &
  Deep gradient network, gradient-induced transition 
  \\[0.01cm] 
23 &
~\href{https://github.com/vishal3477/Proactive-Object-Detection}{PrObeD}\cite{asnani2023probedproactiveobjectdetection} &
  $ NeurIPS_{23} $ & 
  --- &
  --- &
  --- &
  --- &
  Wrapper based on proactive schemes 
  \\[0.01cm] 
24 &
~\href{https://github.com/ChunmingHe/WS-SAM}{WS-SAM}~\cite{he2023weaklysupervised} &
  $ NeurIPS_{23} $ & 
  ResNet50 &
  L &
  S &
  W &
  Pseudo labeling with SAM, multi-scale feature grouping 
  \\[0.01cm] 
25 &
Bi-RRNet~\cite{LIU2023109514} &
  $ PR_{23} $ & 
  VAN-samll &
  L &
  S &
  F &
  Multi-scale scene perception, region-consistency enhancement 
  \\[0.01cm] 
26 &
CPNet~\cite{rs15051188} &
  $ RS_{23} $ & 
  ResNet50, ResNet101, SwinT &
  L &
  S &
  F &
  Ternary cascade perception 
  \\[0.01cm] 
27 &
DBFN~\cite{DBFN} &
  $ SCIS_{23} $ & 
  Res2Net50 &
  L &
  S &
  F &
  Double-branch fusion, parallel attention selection mechanism 
  \\[0.01cm] 
28 &
~\href{https://github.com/hu-xh/PRNet}{PRNet}~\cite{10379651} &
  $ TCSVT_{23} $ &
  SMT-T &
  L &
  S &
  F &
  Cascaded attention perceptron, guided refinement decoder 
  \\[0.01cm] 
29 &
DCNet\cite{10262011} &
  $ TCSVT_{23} $ & 
  PVTv2 &
  B &
  M &
  F &
  Area-boundary decoder, search \& refinement 
  \\[0.01cm] 
30 &
CMNet\cite{10231131} &
  $ TCSVT_{23} $ &
  Res2Net50 &
  H &
  M &
  F &
  De-camouflaging manner, modeling task-conflicting\&consistent attribute 
  \\[0.01cm] 
31 &
~\href{https://github.com/Haozhe-Xing/SARNet}{SARNet}~\cite{10065514} &
  $ TCSVT_{23} $ & 
  PVT &
  L &
  S &
  F &
  Searching, amplifying \& recognizing 
  \\[0.01cm] 
32 &
~\href{https://github.com/yuliu316316/MSCAF-COD}{MSCAF-Net}~\cite{10045692} &
  $ TCSVT_{23} $ & 
  PVTv2 &
  L &
  S &
  F &
  Enhanced receptive field, cross-scale feature fusion, dense interactive 
  \\[0.01cm] 
33 &
~\href{https://github.com/JingZhang617/COD-Rank-Localize-and-Segment}{LSR-V2} ~\cite{Lv2022TowardDU} &
  $ TCSVT_{23} $ & 
  ResNet50 &
  B &
  M &
  F &
  Triple-task learning,~\cite{yunqiu_cod21} extention 
  \\[0.01cm] 
34 &
~\href{https://github.com/jiangxinhao2020/JCNet}{JCNet}~\cite{10224812} &
  $ TIM_{23} $ &  
  SwinT &
  H &
  M &
  F &
  Contrastive learning with salient object 
  \\[0.01cm]  
35 &
ZSCOD~\cite{10234216} &
  $ TIP_{23} $ & 
  ResNet50 &
  B &
  M &
  F &
  Dynamic graph searching, camouflaged visual reasoning generator
  \\[0.01cm] 
36 &
~\href{https://github.com/Jun-Pu/PUENet}{PUENet}~\cite{zhang:hal-04142929} &
  $ TIP_{23} $ &  
  ResNet50, Res2Net50, ViT &
  B &
  M &
  F &
  \begin{tabular}[t]{@{}c@{}}Bayesian conditional variational auto-encoder,\\predictive uncertainty approximation\end{tabular}
   \\[0.01cm] 
37 &
~\href{ttps://github.com/SongZeHNU/FSNet}{FSNet}~\cite{10103836} &
  $ TIP_{23} $ & 
  SwinT &
  L &
  S &
  F &
  Two-stage focus \& scanning, dynamic difficulty aware loss 
  \\[0.01cm] 
38 &
~\href{https://github.com/fanyang587/MGL}{MGL-V2}~\cite{9962828} &
  $ TIP_{23} $ & 
  ResNet50 &
  B &
  M &
  F &
  \begin{tabular}[t]{@{}c@{}}Mutual graph learning,\\multi-source attention contextual recovery,~\cite{9577564} extention\end{tabular}
   \\[0.01cm] 
39 &
~\href{https://github.com/lyu-yx/UEDG}{UEDG}~\cite{10183371} &
  $ TMM_{23} $ & 
  PVTv2 &
  B &
  M &
  F &
  Uncertainty reasoning \& edge inference 
  \\[0.01cm] 
40 &
~\href{https://github.com/zc199823/BBNet--CoCOD}{BBNet}~\cite{zhang2023collaborative} &
  $ TNNLS_{23} $ &  
  Res2Net &
  H &
  S &
  F &
  \begin{tabular}[t]{@{}c@{}}Inter-image collaborative feature exploration,\\ intra-image object feature search, local-global refinement\end{tabular} 
  \\[0.01cm] 
41 &
~\href{https://github.com/XinyuYanTJU/MRR-Net}{MRR-Net}~\cite{10180211} &
  $ TNNLS_{23} $ & 
  ResNet50, Res2Net50 &
  L &
  S &
  F &
  Matching, recognition \& refinement 
  \\[0.01cm] 
42 &
~\href{https://github.com/dwardzheng/MFFN\_COD}{MFFN}~\cite{Zheng_2023_WACV} &
  $ WACV_{23} $ & 
  ResNet50 &
  A &
  S &
  F &
  Co-attention of multi-view, channel fusion unit 
  \\[0.01cm] 
43 &
OWinCANet~\cite{li2023crosslevelattentionoverlappedwindows} &
  $ arXiv_{23} $ & 
  PVTv2 &
  L &
  S &
  F &
  Overlapped window cross-level attention 
  \\[0.01cm] 
44 &
~\href{https://github.com/clelouch/BTSNet}{BTSNet}~\cite{chen2023bioinspiredthreestagemodelcamouflaged} &
  $ arXiv_{23} $ & 
  ResNet50 &
  B &
  M &
  F &
  Three-stage coarse-to-fine, boundary enhancement, mask-guided fusion 
  \\[0.01cm] 
45 &
MLKG~\cite{cheng2023largemodelbasedreferring} &
  $ arXiv_{23} $ & 
  SAM, CLIP &
  A &
  S &
  F &
  Multi-level knowledge descriptions from Multimodal LLM 
  \\[0.01cm] 
46 &
~\href{https://github.com/zhangxuying1004/RefCOD}{R2CNet}~\cite{zhang2023referring} &
  $ arXiv_{23} $ & 
  ResNet50 &
  A &
  S &
  F &
  Reference \& segmentation branches 
  \\[0.01cm] 
47 &
SAM-Adapter~\cite{chen2023samfailssegmentanything} &
  $ arXiv_{23} $ & 
  SAM &
  --- &
  --- &
  --- &
  Incorporating domain-specific information or visual prompts 
  \\[0.01cm] 
48 &
PFRNet\cite{dong2023needadditionalpriorscamouflage} &
  $ arXiv_{23} $ & 
  Res2Net50 &
  L &
  S &
  F &
  \begin{tabular}[t]{@{}c@{}}Adaptive feature aggregation, feature refinement,\\context-aware feature decoding\end{tabular}
  \\[0.01cm] 
49 &
CamoFourier~\cite{le2023unveilingcamouflagelearnablefourierbased} &
  $ arXiv_{23} $ & 
  PatchGAN &
  --- &
  --- &
  F &
  Learnable Fourier-based augmentation, adaptive hybrid swapping 
  \\[0.01cm] 
50 &
~\href{https://npucvr.github.io/UJSCOD/}{UJSCOD-V2}~\cite{li2023jointsalientobjectdetection} &
  $ arXiv_{23} $ & 
  ResNet50 &
  H &
  M &
  F &
  Joint-task contrastive \& uncertainty-aware learning,~\cite{aixuan_cod_sod21} extention 
  \\[0.01cm] 
51 &
PAD~\cite{xing2023pretrainadaptdetectmultitask} &
  $ arXiv_{23} $ & 
  ViT & 
  H &
  M &
  F &
  Parameter efficient adapter, pre-train-adapt-detect 
  \\[0.01cm] 
52 &
~\href{https://github.com/Rapisurazurite/CamoDiffusion}{CamoDiffusion}~\cite{chen2023camodiffusioncamouflagedobjectdetection} &
  $ AAAI_{24} $ & 
  PVT &
  L &
  S &
  F &
  Adaptive transformer conditional network,  denoising network 
  \\[0.01cm] 
53 &
~\href{https://github.com/jyLin8100/GenSAM}{GenSAM}~\cite{hu2023relax} &
  $ AAAI_{24} $ &
  CLIP, BLIP2 &
  A &
  S &
  TF &
  Cross-modal chains of thought prompting, progressive mask generation 
  \\[0.01cm]
54 &
~\href{https://github.com/luckybird1994/MMCPF}{CoVP}~\cite{tang2023generalizationhallucinationlargevisionlanguage} &
  $ ACM MM_{24} $ & 
  Shikra, SAMHQ &
  A &
  S &
  TF &
  Chain of Visual Perception 
  \\[0.01cm] 
55 &
~\href{https://github.com/Kki2Eve/RISNet}{RISNet}~\cite{Wang_2024_CVPR} &
  $ CVPR_{24} $ &
  PVT &
  A &
  S &
  F &
  Depth-guided feature decoder, iterative feature refinement 
  \\[0.01cm] 
56 &
~\href{https://github.com/Sssssuperior/VSCode}{VSCode}~\cite{luo2024vscode} &
  $ CVPR_{24} $ &
  Swin &
  H &
  M &
  F &
  \begin{tabular}[t]{@{}c@{}}2D domain-specific and task-specific prompt learning,\\prompt discrimination loss\end{tabular}
   \\[0.01cm]
57 &
OVCoser~\cite{OVCOS_ECCV2024} &
  $ ECCV_{24} $ & 
  CLIP &
  H &
  M &
  F &
  Semantic guidance, structure enhancement, iterative refinement 
  \\[0.01cm] 
58 &
EANet~\cite{10448139} &
  $ ICASSP_{24} $ &
  EfficientNetB4 &
  B &
  M &
  F &
  Edge-attention guidance, progressive recognition 
  \\[0.01cm] 
59 &
~\href{https://github.com/ChunmingHe/Camouflageator}{Camouflageator}~\cite{he2023strategic} & 
$ ICLR_{24} $ &
ResUNet, ResNet50 &
H &
S &
F &
Adversarial training, internal coherence and edge guidance 
\\[0.01cm] 
60 &
FS-CDIS~\cite{nguyen2024art} &
  $ IEEE Access_{24} $ &
  ResNet101 &
  B &
  M &
  F &
  Instance memory storage, instance triplet loss 
  \\[0.01cm] 
61 &
DINet~\cite{10417767} &
  $ TMM_{24} $ &
   Res2Net50 &
  L &
  S &
  F &
  Body \& detail blocks, global context unit 
  \\[0.01cm] 
62 &
~\href{https://github.com/HVision-NKU/CamoFormer}{CamoFormer}~\cite{yin2022camoformermaskedseparableattention} &
  $ {TPAMI}_{24} $ &
  \begin{tabular}[t]{@{}c@{}}PVTv2, ResNet50,\\ Swin, ConvNeXt-B\end{tabular} &
  L &
  S &
  F &
  Masked separable attention 
  \\[0.01cm]
63 &
~\href{https://github.com/abbasmbz3797/CamoFocus}{CamoFocus}~\cite{10483928} &
  $ WACV_{24} $ &
  PVTv2 &
  L &
  S &
  F &
  Feature split and modulation, context refinement 
  \\[0.01cm] 
64 &
~\href{https://hongshuochen.com/GreenCOD/}{GreenCOD}~\cite{chen2024greencodgreencamouflagedobject} &
  $ arXiv_{24} $ &
  EfficientNetB4 &
  L &
  S &
  F &
  Gradient boosting without back propagation 
  \\[0.01cm] 
65 &
~\href{https://github.com/ZNan-Chen/AGLNet}{AGLNet}~\cite{chen2024adaptiveguidancelearningcamouflaged} &
  $ arXiv_{24} $ &
  EfficientNetB4 &
  L &
  M &
  F &
  Hierarchical feature combination, recalibration decoder 
  \\[0.01cm] 
66 &
~\href{https://github.com/linuxsino/SENet}{SENet}~\cite{hao2024simpleeffectivenetworkbased} &
  $ arXiv_{24} $ &
  ViT &
  B &
  M &
  F &
  Local information capture module, dynamic weighted loss 
  \\[0.01cm] 
67 &
CoFiNet~\cite{guo2024cofinetunveilingcamouflagedobjects} &
  $ arXiv_{24} $ &
  SwinV2 &
  L &
  S &
  F &
  \begin{tabular}[t]{@{}c@{}}Multi-scale feature integration,\\multi-activation selective kernel, dual-mask\end{tabular}
  \\[0.01cm] 
68 &
SCLoss~\cite{yang2024spatialcoherencelosssalient} &
  $ arXiv_{24} $ &
  --- &
  --- &
  --- &
  --- &
  Spatial coherence loss 
  \\[0.01cm] 
  \bottomrule[1.5pt]
\end{tabular}%
}
\end{table*}
        
        The supervision level describes the degree and nature of supervision provided during the training phase. This can be classified into four categories: fully supervised, with complete ground truth data; weakly-supervised, which utilizes limited or imprecise annotations~\cite{he2023weaklysupervised,S-COD}; semi-supervised, which combines labeled and unlabeled data, generating pseudo labels for the latter~\cite{liu2023camouflaged}; and unsupervised, where no explicit labels are provided~\cite{zhang2023unsupervisedcamouflagedobjectsegmentation}. Notably, the "train-free" mode~\cite{tang2023generalizationhallucinationlargevisionlanguage,he2024diffusion,hu2023relax} refers to models that do not require a traditional training process.

        Beyond these categories, existing works also employ several strategies, such as aggregating multi-scale features, simulating bio-inspired mechanisms, fusing multi-source information, learning multiple tasks, jointing SOD, and establishing novel tasks, all aimed at enhancing COD performance. These strategies underscore the diverse and innovative approaches that researchers adopt to address the challenges in COD, revealing the underlying principles and techniques driving advancements. By focusing on these strategies, we can better understand the strengths and limitations of different methods, identify emerging trends, and provide a clearer roadmap for future research. However, there is currently a lack of detailed categorization and analysis of the various strategies employed. Hence, we aim to provide an elaborate introduction to this field. Fig.~\ref{fig:deep-COD-strategy-rate} presents a comprehensive breakdown of the various methods discussed in this paper, along with their respective contributions to the overall analysis as detailed in the following subsections, including \textbf{6} types as well as the sub-types of multi-task and multi-source strategies.
        
        Besides, recent COD methods commonly utilize encoder-decoder architectures with the above strategies. The encoder is responsible for capturing rich semantic and structural information from input images, while the decoder reconstructs the image with the desired changes with the discriminative features extracted by the encoder. 
        As shown in Tabs.~\ref{tab:image-deep-1922} and~\ref{tab:image-deep-2324}, they often extract discriminative features via convolution-based encoders, such as VGG~\cite{simonyan2014very}, ResNet~\cite{he2016deep}, Res2Net~\cite{gao2019res2net}, and EfficientNet~\cite{tan2019efficientnet}, or transformer-based encoders, such as ViT~\cite{dosovitskiy2020vit}, CLIP~\cite{radford2021learning}, BLIP2~\cite{li2023blip}, PVT~\cite{wang2021pyramid}, and Swin~\cite{liu2021swin}. 

        \vspace{-5mm}
        \begin{center}
            \setlength{\abovecaptionskip}{0.1cm}
            \includegraphics[width=.49\textwidth]{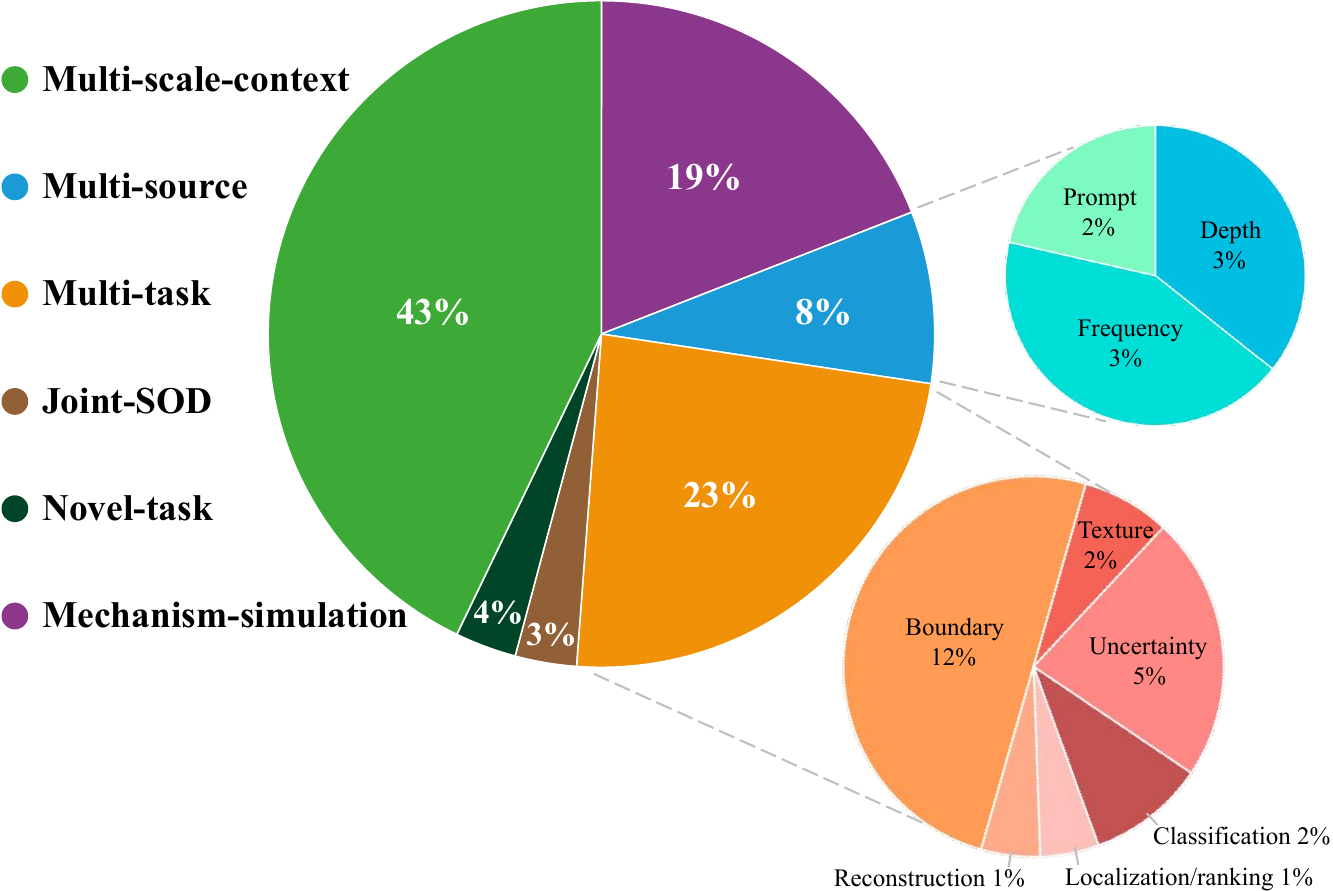}
            \captionof{figure}{The ratio of deep learning COD papers analyzed in this survey to each strategy}
            \label{fig:deep-COD-strategy-rate}
        \end{center}

        \subsubsection{Multi-scale-context based strategy}
        
        This strategy captures the diverse appearances and varying scales of camouflaged objects with rich context information, and then aggregate cross-level features~\cite{ji2022fast,10.1145/3581783.3611773,10219858} while gradually refining features~\cite{dong2023needadditionalpriorscamouflage,fang2024real,10379651,10483928}, specifically in a hierarchical~\cite{ZHANG2022103450,ZoomNet-CVPR2022,10219627,chen2024adaptiveguidancelearningcamouflaged}, residual~\cite{9606888,DBLP:journals/corr/abs-2101-04704}, dual-branch~\cite{chen2022camouflaged,dong2021accuratecamouflagedobjectdetection,OPNet,guo2024cofinetunveilingcamouflagedobjects,9430677,DBFN}, X-connection~\cite{ZHUGE2022108644} or iterative~\cite{Wang_2024_CVPR,Jia_2022_CVPR,Hu_Wang_Qin_Dai_Ren_Luo_Tai_Shao_2023} manner,  to enhance the representation. Some are further guided by edge~\cite{sun2023edgeawaremirrornetworkcamouflaged,zhou2022feature,sun2022bgnet,10448139}, frequency~\cite{10.1145/3545609,cong2023frequencyperceptionnetworkcamouflaged}, querying~\cite{sun2022dqnetcrossmodelqueryingcamouflaged} or coarse prediction maps~\cite{chen2023bioinspiredthreestagemodelcamouflaged} in fusion process.

        ERRNet~\cite{ji2022fast} and HCM~\cite{xiao2023concealed} propose a reversible re-calibration mechanism that leverages prior prediction maps, specifically targeting low-confidence regions to detect previously missed parts. This approach refines detection by focusing on regions that are initially overlooked. To improve efficiency, TinyCOD~\cite{10095226} introduces an adjacent scale feature fusion strategy using the lightweight TinyNet~\cite{10095226} as the backbone.
        Inspired by the Transformer architecture~\cite{vaswani2017attention}, CamoFormer~\cite{yin2022camoformermaskedseparableattention} adopts masked separable attention, where multi-head self-attention is divided into three components. This approach allows for the simultaneous refinement of features at different levels in a top-down manner.
        While vision transformers excel in global context modeling, they often struggle with locality modeling and feature fusion. To address these issues, FSPNet~\cite{Huang2023Feature} introduces a non-local token enhancement mechanism for improved feature interaction and a feature shrinkage decoder. Building on the Swin transformer~\cite{liu2021swin} and its shifted window strategy, OWinCANet~\cite{li2023crosslevelattentionoverlappedwindows} employs overlapped window cross-level attention. This method enhances low-level features with high-level guidance by sliding aligned window pairs across feature maps, ensuring a balance between local and global for superior performance.

        Pixel-wise annotation of camouflaged objects is time-consuming and labor-intensive. Weakly supervised methods, using limited or scribble annotations, aim to reduce labeling effort and address boundary ambiguities. CRNet~\cite{S-COD} designs a local-context contrasted module to enhance image contrast and a logical semantic relation module to analyze semantic relations, combined with feature-guided and consistency losses to impose stability on the predictions. He \textit{et al.}~\cite{he2023weaklysupervised} utilize the visual foundation model SAM~\cite{kirillov2023segment} with sparse annotations as prompts to achieve initial coarse segmentation. Enhanced with multi-scale feature grouping, they generate reliable pseudo-labels for training off-the-shelf methods, addressing intrinsic similarity issues in coherent segmentation.

        \subsubsection{Mechanism-simulation based strategy}
        
        This bio-inspired strategy simulates the behavior of predators in nature or the visual detection mechanisms of humans. This is a multi-stage, coarse-to-fine strategy to incrementally enhance the accuracy of outcomes. 
        SINet~\cite{fan2020camouflaged} simulates the search and identification process of predators, utilizing densely connected features and receptive fields, further enhanced by SINet-V2~\cite{fan2022concealed} with group-reversal attention. Semi-SINet~\cite{liu2023camouflaged} addresses training data scarcity with semi-supervised learning, generating pseudo labels for unlabeled data and introducing edge attention to improve SINet. Such localization-segmentation two-stage process has also inspired many approaches~\cite{liu2021integrating, Mei_2021_CVPR,10262011,10103836,cong2023frequencyperceptionnetworkcamouflaged,OPNet,10219627}. Besides, extra stages, \textit{e.g.}, restoring~\cite{ijcai2023p124}, matching~\cite{10180211} and amplifying~\cite{10065514}, have also been introduced for more precise detection 
        and enhanced adaptability.

        MirrorNet~\cite{Jinnan-IEEEAccess2021} employs a mirror stream with embedded image flipping as a bio-inspired strategy to disrupt camouflage. ZoomNet~\cite{ZoomNet-CVPR2022} emulates human visual patterns when observing ambiguous images, specifically through zooming in and out, and leverages scale integration and hierarchical units to capture mixed-scale semantics. In contrast to multi-scale input strategies, MFFN~\cite{Zheng_2023_WACV} acquires complementary information through multi-view inputs from various angles, distances, and perspectives. This approach is designed to address the visually indistinguishable characteristics of camouflaged objects that arise under complex conditions, such as tiny object size, fuzzy objects, and blurred boundaries.

        Moreover, incorporating auxiliary tasks could contribute to a better understanding of camouflage. PreyNet~\cite{10.1145/3503161.3548178} mimics the sensory and cognitive mechanism of predation combined with the auxiliary task, uncertainty estimation, through a bidirectional interaction module for feature aggregation, as well as a policy-and-calibration paradigm for feature calibration.
        Camouflageator~\cite{he2023strategic}, inspired by the prey-predator dynamics, presents an adversarial training framework with an auxiliary generator creating challenging camouflaged objects on the prey side. As for the predator side, Camouflageator employs a camouflaged feature coherence module and utilizes edge-guided calibration to enhance complete segmentation and boundary clarity.
    
        \subsubsection{Multi-source information fusion strategy}
        
        This strategy integrates diverse supplementary information sources to enhance COD, improving robustness and accuracy by leveraging complementary data from different domains, mainly frequency, depth, or text.

            \noindent\textbf{Frequency-domain integrated approaches.}
            FEMNet~\cite{9878807} is the first to extend COD into the spatial domain, incorporating frequency enhancement and high-order relation for effective digging and fusion of the frequency clues with RGB features.
            FEDER~\cite{He2023Camouflaged} further decomposes the features into different frequency bands via learnable wavelets, using frequency attention and guidance-based aggregation to differentiate foreground-background, complemented by an ordinary differential equation-inspired edge reconstruction for precise boundaries.

            \noindent\textbf{Depth-perceptual integrated approaches.}
            DCE\cite{xiang2022exploringdepthcontributioncamouflaged} introduce the first depth-guided COD network, which incorporates an auxiliary depth estimation branch and a multi-modal confidence-aware loss function via GAN\cite{goodfellow2014generative} for effective depth integration. 
            Building on DCE, DaCOD~\cite{10.1145/3581783.3611874} also utilizes existing monocular depth estimation methods to generate depth maps and proposes a novel cross-modal asymmetric fusion strategy to blend these modalities asymmetrically. 
            XMSNet~\cite{wu2023object} implements all-around attentive fusion to facilitate explicit cross-modal semantic mining and consistency constraints across decoding layers, thereby mitigating bias from inherent noise in depth estimation. 
            PopNet~\cite{10377562} employs source-free depth for object pop-out, identifying contact surfaces under weak supervision and leveraging 3D priors for segmentation without source data, which enables efficient depth-to-semantics transfer. 
            RISNet~\cite{Wang_2024_CVPR}, designed for extreme agricultural application scenarios, uses multi-scale receptive fields and depth feature fusion to enhance dense and small COD in multiple stages.
            
            \noindent\textbf{Prompt-learning integrated approaches.}
            These approaches utilize textual or visual prompts to adaptively guide and enhance COD models. CoVP~\cite{tang2023generalizationhallucinationlargevisionlanguage} introduces a chain of visual perception and language text prompts, which linguistically and visually enhance the camouflaged scene perception of large vision-language models (LVLM), thereby reducing hallucinations. In contrast, GenSAM~\cite{hu2023relax} employs cross-modal chains of thought prompting to generate visual prompts and progressively produce masks, iteratively refining the detection results. Both CoVP and GenSAM operate in a training-free manner, leveraging pre-trained models to avoid the need for extensive retraining, save computational resources, and enable rapid adaptation to COD. Rather than relying on text prompts, VSCode~\cite{luo2024vscode} introduces 2D domain-specific and task-specific prompt learning, effectively disentangling domain and task peculiarities. Through joint training, VSCode emerges as the first generalist model capable of addressing multimodal SOD and COD tasks with remarkable zero-shot generalization ability.

        \subsubsection{Multi-task learning strategy}
        
       The fundamental premise of multi-task learning is that different tasks share common information for data processing. By leveraging these additional cues, multi-task learning is extensively employed to extract complementary insights from positively correlated tasks—such as reconstruction, classification, and localization—to enhance the detection accuracy of camouflaged objects.
        
            \noindent\textbf{Boundary-supervision integrated approaches.}
            MGL~\cite{9577564} integrates camouflaged object-aware edge extraction (COEE) with graph-based mutual learning, incorporating typed functions to enhance feature representation by capturing both semantic and spatial information, which benefits COD and boundary detail accuracy. To address interpolation accuracy loss and computational redundancy in MGL, MGL-V2~\cite{9962828} incorporates multi-source attention contextual recovery, iteratively leveraging pixel feature information. Inspired by human COD processes, BSA-Net~\cite{Zhu_Li_Xie_Yan_Liang_Chen_Wei_Qin_2022} employs a two-stream separated attention mechanism—reverse and normal attention streams—followed by a boundary guider to enhance performance by focusing on object boundaries. Beyond boundary enhancement, the extension of BSA-Net, FindNet~\cite{9923635}, embeds texture information into feature representation, focusing on local patterns to perform effectively under complex COD conditions. To better integrate multiple features, ASBI~\cite{ZHANG2023103719} introduces an attention-induced semantic and boundary interaction network, utilizing attention-induced interaction to completely fuse multivariate and heterogeneous information.

            \noindent\textbf{Category-prediction integrated approaches.}
            ANet~\cite{ltnghia-CVIU2019} is the very classic method for COD, which includes a classification stream to predict whether they contain camouflaged objects and a segmentation stream to recognize them. 
            To leverage the strong generalization ability and rich semantic knowledge of large-scale pre-trained foundation models, PAD~\cite{xing2023pretrainadaptdetectmultitask} utilizes the pre-trained ViT~\cite{dosovitskiy2020vit} with lightweight parallel adapters by only tuning a small number of parameters for multi-task learning in a “pre-train, adapt, and detect” paradigm, achieving zero-shot task transferability, multi-task adaption, and cross-task generalization.
            As unseen classes are more general in real-world scenarios, Li \textit{et al.} propose ZSCOD~\cite{10234216} for zero-shot learning, which employs dynamic graph searching to adaptively capture edge details and a camouflaged visual reasoning generator for generating pseudo-features with the object-wise graph learning strategy to dynamically sample nodes to reduce background interference.
            Considering the limitation of data on COD, FS-CDIS~\cite{nguyen2024art} leverages few-shot learning, simultaneously implementing instance segmentation task, with proposed instance triplet loss and instance memory storage, enhancing distinguishable features between background and foreground areas.
            
            \noindent\textbf{Localization/ranking integrated approaches.}
            LSR~\cite{yunqiu_cod21} introduces the first ranking-based COD network, which infers the detectability of different camouflaged objects through instance segmentation and classification branches. LSR-V2~\cite{Lv2022TowardDU} builds on the pioneering work of LSR by further exploring the interdependencies among these tasks. It not only establishes a new baseline and benchmark for both individual and joint tasks within a triple-task learning framework but also revisits the role of the camouflaged object ranking (COR) task.

            \noindent\textbf{Reconstruction integrated approaches.}
            FRINet~\cite{10.1145/3581783.3611773} uses Laplacian pyramid-like decomposition and transformer-CNN hybrid encoders with a reasoning module to capture and integrate high and low-frequency components, guided by an auxiliary image reconstruction task. 
            Considering the strong performance of the transformer, Hao \textit{et al.}~\cite{hao2024simpleeffectivenetworkbased} uses ViT and regards both image reconstruction and binary segmentation tasks as training targets, with a local information capture module and dynamic weighted loss to enhance local modeling and handle complex cases.
            
            \noindent\textbf{Texture-detection integrated approaches.}
            Motivated by the need to leverage complementary texture and camouflaged object cues, Zhu \textit{et al.} utilizes texture labels and an interactive guidance framework. This framework consists of feature interaction guidance as well as texture and holistic perception decoders, aimed at refining segmentation with a particular focus on indefinite boundaries and texture differences. 
            To better exploit the discriminative patterns within the objects, DGNet~\cite{ji2023gradient} decouples COD into context and texture branches based on gradient generation. It employs a gradient-induced transition to softly group features from both branches, resulting in superior efficiency.
            
            \noindent\textbf{Uncertainty-estimation based approaches.}
            By approximating the uncertainty across different areas, these methods enhance the model's ability to focus on less identifiable regions, ultimately achieving high-confidence detection. UR-COD~\cite{kajiura2021improving} utilizes uncertainty-aware refinement to reduce the noise of pseudo-edge and pseudo-map labels. Yang \textit{et al.}~\cite{yang2021uncertainty} introduce Bayesian learning into transformer-based reasoning which revolutionizes the traditional deterministic mapping process employed in conventional COD by transitioning it into an uncertainty-guided context reasoning procedure. 
            OCENet~\cite{Liu_2022_WACV} employs aleatoric uncertainty estimation for confidence-aware COD, using dynamic supervision for accurate maps and assessing pixel-wise accuracy without ground truth.
            PUENet~\cite{zhang:hal-04142929} integrates model and data uncertainty, implementing predictive uncertainty estimation and predictive uncertainty approximation for efficient test-time alongside SAM refining hierarchical features.

        \subsubsection{Joint-SOD based strategy}
        SOD and COD seem opposing but share common ground in the necessity to discern objects from a background based on contrast and contextual features. This strategy combines SOD with COD, leveraging the contradictory information or shared characteristics to gain more thorough comprehension of COD. Interestingly, many works are built in a multi-task manner, \textit{e.g.}, boundary detection~\cite{10231131,10224812} and image reconstruction~\cite{hao2024simpleeffectivenetworkbased}.
        
        UJSCOD~\cite{aixuan_cod_sod21} uses uncertainty-aware adversarial learning with a similarity measure module for modeling contradicting attributes and a data interaction strategy by defining simple COD samples as hard SOD samples for SOD data augmentation and higher robustness. Additionally, UJSCOD-V2~\cite{li2023jointsalientobjectdetection} introduces contrastive learning to further investigate the cross-task correlations and random sampling-based foreground-cropping for COD data augmentation. 
        CMNet~\cite{10231131} proposes a new perspective, de-camouflaging, modeling task-conflicting and task-consistent attributes to destroy the camouflage.

        \subsubsection{Novel-task setting strategy}
        As COD continues to gain attention, its application scenarios have become increasingly diverse. Consequently, researchers have proposed various novel task settings to address COD challenges in different contexts, offering significant advantages, such as improved generalization to unseen data, better handling of complex scenes, and broadening the applicability of COD technologies.

        \noindent\textbf{Unsupervised camouflaged object segmentation (UCOS).}
        Unsupervised learning is necessary due to the challenges in gaining extensive human labels for open-world applications, where supervised models often exhibit poor generalization. Domain adaptation is crucial because it allows models to effectively transfer knowledge from a source domain to a target domain. This adaptation helps bridge the gap between different data distributions, enhancing the model's performance in real-world scenarios where data characteristics may vary significantly. Therefore, a novel task, termed UCOS-DA~\cite{zhang2023unsupervisedcamouflagedobjectsegmentation}, is formulated for unsupervised COD where both source and target labels are absent in the training phase. UCOS-DA leveraging foreground-background contrastive self-adversarial for pseudo-labels and domain-specific adaptation to address the UCOS task.

        \noindent\textbf{Collaborative camouflaged object detection (CoCOD).}
        Zhang \textit{et al.}~\cite{zhang2023collaborative} introduce this novel task to address the limitations of single-image analysis by jointly segmenting the same camouflaged object or objects belonging to the same class across multiple distinct images. This strategy leverages shared similarities and complementary cues inherent in the related images, thereby enhancing the accuracy of COD.
        To tackle CoCOD, BBNet~\cite{zhang2023collaborative} extracts and integrates camouflaged object features through inter-image collaborative feature exploration and intra-image object feature search. Additionally, BBNet enhances the representation of co-camouflaged features by employing the strategy of local-global feature refinement.

        \noindent\textbf{Referring camouflaged object detection (RefCOD).}
        Standard COD aims to detect all camouflaged objects in a given scene. However, in certain real-world scenarios, such as ecological species protection and discovery, explorers may be interested only in locating specific camouflaged objects. To address this need, target references are introduced into COD, where referring text or images containing salient targets guide the detection of specified camouflaged objects. R2CNet~\cite{zhang2023referring} introduces reference and segmentation branches, combined with referring mask generation, to create pixel-level priors and enrich referring features, thereby enhancing detection capabilities. In contrast, MLKG~\cite{cheng2023largemodelbasedreferring} uses text as a reference in a multi-level knowledge-guided multimodal approach. By leveraging multimodal large language models (MLLMs), it achieves deep alignment, improves performance, and enables zero-shot generalization on unimodal datasets.

        \noindent\textbf{Open-vocabulary camouflaged object segmentation (OVCOS).} 
        Open-vocabulary refers to models that recognize and segment objects from novel classes not seen during training, leveraging vision-language models(VLM) like CLIP~\cite{radford2021learning}. To fill in the gaps for COD in this field, Pang \textit{et al.}~\cite{OVCOS_ECCV2024} introduce the novel task OVCOS and built the corresponding baseline OVCoser, which integrates semantic guidance and visual structure cues in an iterative refinement manner via a transformer-based architecture attached to a frozen CLIP~\cite{radford2021learning}. It also incorporates diverse sources of information, such as class semantic cues, spatial depth structures, object edge details, and iterative top-down guidance from the output space. To enhance task-relevant semantic context, OVCoser employs carefully crafted prompt templates, achieving robust and generalized performance.
    
        \section{Video-level COD models}
        
        Unlike image-level Camouflaged Object Detection (COD) techniques, which focus on single static images, video-level COD requires greater emphasis on motion cues to identify and localize camouflaged objects within continuous video frames. Video COD (VCOD) typically leverages temporal information, such as motion and changes across frames, to reveal objects that are difficult to detect in individual frames. However, this task presents significant challenges, including complex background noise, lighting variations, occlusions, and diverse camouflage strategies. Additionally, the high-dimensional nature of video data necessitates algorithms that are not only spatially accurate but also temporally consistent and stable. In this section, we categorize existing methods into two main types: traditional approaches and deep learning-based approaches.
        \subsection{Traditional VCOD methods} 
        \label{section:video-tradition}
        
\begin{table*}[!htb]
\centering
\caption{\textcolor{red}{12} representative traditional methods of VCOD. For more details, please refer to Section \ref{section:video-tradition}}
\label{tab:video-tradition}
\resizebox{\textwidth}{!}{%
\begin{tabular}{crcccc}
\toprule[1.5pt]
\textit{\textbf{Index}} &
  \textit{\textbf{Method}} &
  \textit{\textbf{Pub.}} &
  \textit{\textbf{Year}} &
  \textit{\textbf{Feature}} &
  \textit{\textbf{Key Tech.}} \\ 
\midrule[1pt]  
1 &
  VSNCT\cite{2001Into} &
  $ P IEEE $ &
  2001 &
  Motion &
  Dynamic \& global thresholds \\ 
  
2 &
  MBBS\cite{2004Motion} &
  $ CVPR $ &
  2004 &
  \begin{tabular}[t]{@{}c@{}}Color \&\\Optical flow\end{tabular} &
  \begin{tabular}[t]{@{}c@{}}The Normalized RGB color for estimating thedensity,\\the Optical flow feature for distinguishing motion exchange\end{tabular} \\ 
3 &
  CCMS\cite{2007Improving} &
  $ IbPRIA $ &
  2007 &
  \begin{tabular}[t]{@{}c@{}} Color \&\\Motion\end{tabular} &
  Background subtraction, integration of color and motion \\ 
4 &
  TAMVF\cite{2008A} &
  $ ICALIP $ &
  2008 &
  Intensity &
  Bayesian classification, Gaussian mixture model \\ 
5 &
  PCPD\cite{2009An} &
  $ AVSS $ &
  2009 &
  Motion &
  \begin{tabular}[t]{@{}c@{}}Post-processing phase,\\integrating fragmented blocks to recover the original input\end{tabular} \\ 
6 &
  OTCID\cite{2010Optical} &
  $ Opt. Commun. $ &
  2010 &
  Optical flow &
  \begin{tabular}[t]{@{}c@{}}Spatially coherent beam illuminates for identification
  \end{tabular} \\ 
7 &
  DBBOF\cite{Jianqin2011Detection} &
  $ Procedia Eng. $ &
  2011 &
  Optical flow &
  Clustering motion pattern, Kalman filter \\ 
8 &
  FODCC\cite{2014Foreground} &
  $ INDICON $ &
  2013 &
  Texture &
  Multiple camera-based codebooks, disparity map \\ 
9 &
  CDIM\cite{2013Camouflage} &
  $ P. Royal Soc. B $ &
  2013 &
  Motion &
  Detection, identification \& capture \\ 
10 &
  USFS\cite{Sungho2015Unsupervised} &
  $ Sci. World J. $ &
  2015 &
  Optical flow &
  \begin{tabular}[t]{@{}c@{}}Statistical distance, entropy-based spatial grouping, K-means clustering\end{tabular} \\ 
11 &
  BASCMOD\cite{2018A} &
  $ TCSVT $ &
  2015 &
  Color &
  \begin{tabular}[t]{@{}c@{}}Discriminative \& camouflage modeling, Bayesian framework\end{tabular} \\ 
12 &
  FWFC\cite{2018A} &
  $ TIP $ &
  2018 &
  Intensity &
  \begin{tabular}[t]{@{}c@{}}Fusion in the wavelet domain, foreground \& background models\end{tabular} \\ 
  \bottomrule[1.5pt]
\end{tabular}%
}
\end{table*}
        
        Compared to image-level methods, the video-level ones include more techniques, like optical flow analysis and motion detection, making them practical in video surveillance and security. As illustrated in Tab.~\ref{tab:video-tradition}, this section will delve into \textbf{12} representative traditional VCOD models, and we also categorize and introduce these methods based on the types of features they rely on.
        
        \noindent\textbf{Texture feature-based approaches.}
        Malathi \textit{et al.}~\cite{2014Foreground} employ multi-camera codebooks for the detection of foreground objects. In their approach, texture pixels resembling the background are extracted and quantized into distinct codebooks. These codebooks are then integrated into a weighted framework to guide the abstraction of the foreground. To enhance the accuracy of predictions, disparity maps generated from the codebooks are used as supplementary information, particularly in cases where the color contrast between the foreground and background is weak. However, there remains the potential for shadows to be misclassified as targets.
        
        \noindent\textbf{Intensity feature-based approaches.}
        When applied to video, these methods track dynamic changes between frames by leveraging temporal information, thereby improving the detection and localization of moving camouflaged objects. To tackle the challenge of foreground-background segmentation in video surveillance, particularly when complicated by camouflage, Guo \textit{et al.}\cite{2008A} combine Bayesian classification with Gaussian mixture models and apply temporal averaging across multiple frames in video sequences to reduce the bias of background models. However, detecting subtle differences remains difficult when the foreground and background are highly similar. To address this issue, Li \textit{et al.}\cite{2018A} extend COD to the wavelet domain, where subtle differences are emphasized in specific wavelet bands. They apply wavelet transforms to video sequences and estimate the probability of a wavelet coefficient belonging to the foreground by constructing foreground and background models within each individual wavelet band. This approach detects camouflaged moving foregrounds through wavelet-domain multi-scale fusion.

        \begin{table*}[!htb]
\centering
\caption{\textcolor{red}{16} representative deep learning methods of VCOD. 
\textcolor{red}{N.A.}: network architecture, including two-stage framework, specifically, \textcolor{blue}{EM}(Explicit Motion-based methods) and \textcolor{blue}{IM}(Implicit Motion-based methods), and End-to-End framework (\textcolor{blue}{EE}). 
\textcolor{red}{O.F.}: whether pre-generating optical flow map.
\textcolor{red}{S.L.}: supervision level, including \textcolor{blue}{F}(Fully-supervision), \textcolor{blue}{S}(Self-supervision) and \textcolor{blue}{U}(Un-supervision). 
\textcolor{red}{S.D.}: whether generating synthetic dataset.
Clicking on some method names may redirect you to their open-source codes or projects.
For more details, please refer to Section \ref{section:video-deep}}
\label{tab:video-deep}
\resizebox{\textwidth}{!}{%
\begin{tabular}{crccccccccc}
\toprule[1.5pt]
\textit{\textbf{Index}} &
  \textit{\textbf{Method}} &
  \textit{\textbf{Pub./Year}} &
  \textit{\textbf{Backbone}} &
  \textit{\textbf{N.A.}} &
  \textit{\textbf{O.F.}} &
  \textit{\textbf{S.L.}} &
  \textit{\textbf{S.D.}} &
  \textit{\textbf{Core component}} &
  \textit{\textbf{Link}} \\
\midrule[1pt]
1 &
  FMC~\cite{xie2019object} &
  $CVPR_{19}$ &
  --- &
  EM &
  \checkmark &
  F &
  --- &
  Pixel-trajectory RNN,  clustering 
  \\ [0.5cm]
2 &
  ~\href{https://github.com/hlamdouar/MoCA/}{VRS}~\cite{Lamdouar20} &
  $ACCV_{20}$ &
  ResNet18 &
  EM &
  \checkmark &
  F &
  --- &
  Differentiable registration \& motion segmentation 
  \\
3 &
  ~\href{https://charigyang.github.io/motiongroup/}{MG}~\cite{yang2021self} &
  $ICCV_{21}$ &
  --- &
  EM &
  \checkmark &
  S &
  --- &
  \begin{tabular}[t]{@{}c@{}}Motion grouping,  \\ self-supervised temporal consistency loss\end{tabular} 
  \\
4 &
  ~\href{https://www.robots.ox.ac.uk/~vgg/research/simo/}{SIMO}~\cite{lamdouar2021segmenting} &
  $BMVC_{21}$ &
  --- &
  EM &
  \checkmark &
  F &
  \checkmark &
  Dual-head architecture 
  \\
5 &
  ~\href{https://github.com/XuelianCheng/SLT-Net}{SLT-Net}~\cite{cheng2022implicit} &
  $CVPR_{22}$ &
  PVT &
  IM &
  --- &
  F &
  --- &
  Short-term detection \& long-term refinement 
  \\
6 &
  ~\href{https://github.com/Etienne-Meunier-Inria/EM-Flow-Segmentation}{OFS}~\cite{meunier2022driven} &
  $TPAMI_{22}$ &
  FakeModel &
  EM &
  \checkmark &
  U &
  --- &
  \begin{tabular}[t]{@{}c@{}}Expectation-Maximization framework,\\motion augmentation\end{tabular} 
  \\
7 &
  ~\href{https://yorkucvil.github.io/Static-Dynamic-Interpretability/}{QSDI}~\cite{kowal2022deeper} &
  $CVPR_{22}$ &
  \begin{tabular}[t]{@{}c@{}} ResNet,\\MobileNetV2 \end{tabular} &
  EM &
  \checkmark &
  F &
  --- &
  Quantifying the static and dynamic biases 
  \\
8 &
  ~\href{https://github.com/Jyxarthur/OCLR_model}{OCLR}~\cite{xie2022segmenting}&
  $NeurIPS_{22}$ &
  ViT-S/8 &
  EM &
  \checkmark &
  F &
  \checkmark &
  Object-centric layered representation 
  \\
9 &
  RCF~\cite{bideau2024right} &
  $IJCV_{22}$ &
  --- &
  EM &
  \checkmark &
  F &
  --- &
  \begin{tabular}[t]{@{}c@{}} Rotation-compensated flow,\\camera motion estimation \end{tabular} 
  \\
10 &
  ~\href{https://github.com/hlamdouar/CAMEVAL}{CAMEVAL}~\cite{lamdouar2023making}&
  $ICCV_{23}$ &
  ResNet50 &
  EM &
  \checkmark &
  F &
  \checkmark &
  Camouflage scores 
  \\
11 &
  ~\href{https://github.com/lartpang/ZoomNeXt}{ZoomNeXt}~\cite{pang2024zoomnextunifiedcollaborativepyramid} &
  $TPAMI_{24}$ &
  \begin{tabular}[t]{@{}c@{}} ResNet, PVTv2,\\EfficientNet\end{tabular}&
  EE &
  --- &
  F &
  --- &
  \begin{tabular}[t]{@{}c@{}} Scale merging subnetwork,\\hierarchical difference propagation decoder \end{tabular} 
  \\
12 &
  TMNet~\cite{yu2024tokenmotion} &
  $ICASSP_{24}$ &
  SegFormer &
  EE &
  --- &
  F &
  --- &
  Learnable token selection 
  \\
13 &
  IMEX~\cite{hui2024implicit} &
  $TMM_{24}$ &
  ResNet50 &
  EE &
  --- &
  F &
  --- &
  Implicit-explicit motion learning 
  \\
14 &
  \href{https://github.com/WenjunHui1/TSP-SAM}{TSP-SAM}~\cite{hui2024endow} &
  $CVPR_{24}$ &
  SAM &
  EE &
  --- &
  F &
  --- &
  Temporal-spatial injection, prompt learning 
  \\
15 &
  \href{https://github.com/SpiderNitt/SAM-PM}{SAM-PM}~\cite{meeran2024sam} &
  $CVPRW_{24}$ &
  SAM-L &
  EE &
  --- &
  F &
  --- &
  Temporal fusion mask, memory prior affinity 
  \\
16 &
  EMIP~\cite{zhang2024explicit} &
  $arXiv_{24}$ &
  PVTv2 &
  EE &
  --- &
  F &
  --- &
  Explicit motion modeling, interactive prompting 
  \\
\bottomrule[1.5pt]
\end{tabular}%
}
\end{table*}

        \noindent\textbf{Color feature-based approaches.}
        Zhang \textit{et al.}\cite{2017A} employs computer-assisted detection of camouflaged targets, which first globally models the background of the input image, with the modeling of the foreground divided into global and local models. The global model captures the overall color and texture information of the foreground, while the local model focuses on details or changes that may exist in the foreground. Based on the models of the background and foreground, a factor measuring the degree of camouflage is introduced, determining whether a pixel is camouflaged by comparing the color differences between the background and foreground at that pixel. By comparing color contrast measurement results, true camouflaged regions are identified. Finally, the camouflage and identification models are fused under a Bayesian framework to perform complete target detection. 
        
        \noindent\textbf{Motion feature-based approaches.}
        These methods leverage the relative movement between the target object and the background between consecutive frames to identify camouflaged objects. 
        Boult \textit{et al.}\cite{2001Into} propose a method for distinguishing background and foreground objects in visual surveillance systems using motion features. They introduce a conditional incremental model to update multiple background models in real-time and employed quasi-connected components to fill gaps caused by slight object movements, adapting to scene changes. 
        However, background subtraction under camouflage conditions often results in fragmented, discontinuous object pieces, complicating subsequent classification or tracking processes.
        To address this, Conte \textit{et al.}\cite{2009An} developed an algorithm for detecting camouflaged personnel by aggregating fragmented detection blocks to reconstruct the complete shape of the target object during post-processing. This algorithm relies on the consistent segmentation of parts of the human body, such as the head, torso, and legs, across video sequences. By defining specific parameters to model the bounding boxes of human figures, the algorithm merges two or more boxes according to a set of rules. Considering perspective effects, a semi-automatic calibration phase dynamically adjusts parameters to ensure the bounding boxes accurately reflect the actual size of the individuals. 

        \noindent\textbf{Optical flow feature-based approaches.}
        Optical flow represents the motion of objects in a sequence as a vector field, providing crucial information on object speed and direction. To address challenges in classifying and identifying objects from low spatial resolution images—particularly in security-related applications—Beiderman \textit{et al.}~\cite{2010Optical} introduce a technique that uses spatially coherent light beams to illuminate scenes and capture secondary speckle patterns from reflections. By tracking the temporal variations of these speckle patterns, the method extracts temporal feature signatures of objects. Comparing these signatures allows the algorithm to differentiate camouflaged objects from their surroundings, leveraging the unique physical properties of object surfaces for effective detection even under heavy camouflage. Building on this approach, Yin \textit{et al.}~\cite{Jianqin2011Detection} propose a different method for detecting camouflaged objects in dynamic backgrounds using optical flow techniques. By simulating the motion patterns of objects and backgrounds and employing clustering analysis of optical flow features, this method accurately identifies objects. The algorithm further optimizes detection results through Kalman filtering, enhancing performance and accuracy while adapting to dynamic environments. Acknowledging the high data dependency of supervised methods and the difficulty of obtaining high-quality data in practical applications, Kim~\cite{Sungho2015Unsupervised} proposes an unsupervised method using a visible-near-infrared hyperspectral camera. This method selects spectral and spatial features online through statistical distance to generate candidate feature bands, followed by entropy-based analysis to remove ineffective features. This approach achieves precise detection of camouflaged objects while reducing computational load.

        \subsection{Deep learning VCOD methods}
        \label{section:video-deep}
        
        Compared to traditional VCOD techniques, deep learning methods verify a significant advantage by automatically learning complex feature representations from large datasets. These methods have a unique ability to capture intricate and subtle patterns, which in turn enhances the understanding of object dynamics within video sequences. However, video data presents greater complexity compared to image-based approaches. This added complexity stems from factors such as higher data dimensions, temporal continuity across frames, and the constantly evolving nature of dynamic changes in video content. These factors require models to possess not only strong spatial feature extraction capabilities but also a robust mechanism for accurately capturing subtle temporal variations. Despite promising advancements, existing work has largely focused on leveraging motion cues between different frames, and these approaches remain in their early stages of development, with considerable room for growth before they reach maturity.

 In recent years, researchers have increasingly adopted a two-step framework, where optical flow maps or pseudo masks are pre-generated to serve as motion cues for video object detection. However, due to the challenges associated with cumulative errors and weak generalization~\cite{hui2024endow}, there is a growing trend towards employing end-to-end universal models to enhance reliability. The key characteristics of a total of \textbf{16} representative methods for VCOD are detailed in Tab.~\ref{tab:video-deep}. We categorize these methods based on various criteria, \textit{e.g.},  network architecture, the use of optical flow maps, supervision level, and synthetic dataset generation. Besides, some methods provide links to their open-source projects.

        \begin{figure*}[ht]
        {\includegraphics[width=\textwidth]{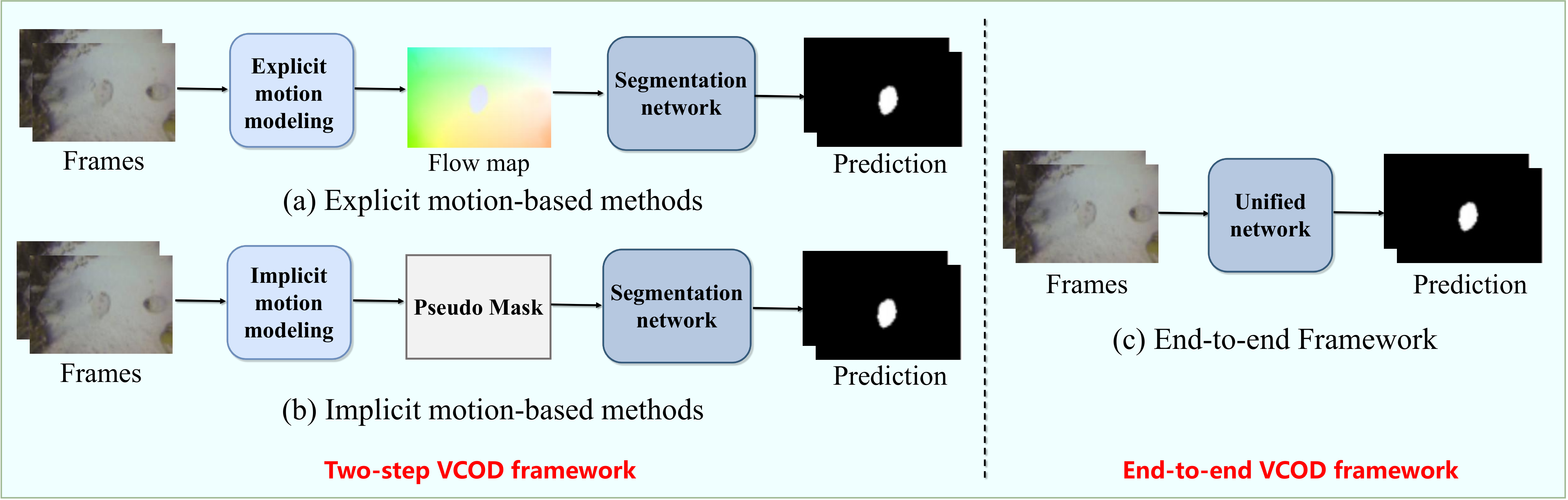}}
        \caption{Two types of framework for deep learning VCOD methods. See Section \ref{section:video-deep} for more details} 
        \vspace{-4mm}
        \label{fig:two-types-deep-vcod}
        \end{figure*}

        \subsubsection{Two-step VCOD framework}
        Since motion cues are crucial for distinguishing moving camouflaged objects from their backgrounds, researchers often incorporate a stage to pre-generate optical flow maps or pseudo masks for extracting motion information. Optical flow maps directly capture motion fields, \textit{i.e.}, the optical flow, between consecutive frames, providing compensation or registration to mitigate the camouflage effect. In contrast, pseudo masks implicitly learn temporal correspondences and motion patterns within the network, reducing reliance on external optical flow estimation and enabling the network to capture motion and maintain temporal consistency.
        
        \noindent\textbf{Explicit motion-based methods.}
        Lamdouar \textit{et al.}~\cite{Lamdouar20} adopt optical flow and a different image as inputs, and propose a differentiable registration module for background alignment and a motion segmentation module with memory for moving object discovery.
        Facing the challenge of massive human annotation, Yang et al.~\cite{yang2021self} introduce a self-supervised method without any manual supervision, which groups motion with similar optical flow according to perceptual grouping principles.
        Meunier \textit{et al.}~\cite{meunier2022driven} utilize unsupervised CNN-based motion segmentation from optical flow, leveraging the Expectation-Maximization framework for loss design and training, as well as designing data augmentation on the optical flow field, enabling real-time segmentation without annotations or iterative motion model estimation.
        Xie \textit{et al.}~\cite{xie2022segmenting} introduce object-centric layered representation and generate synthetic data for multi-object segmentation and tracking. However, reliance on external optical flow estimation can introduce errors that accumulate over time, potentially compromising the final mask prediction.

        \noindent\textbf{Implicit motion-based methods.}
        SLT-Net~\cite{cheng2022implicit} leverages short and long-term spatiotemporal relationships, specifically, utilizing the short-term motion capture between consecutive frames to produce pseudo masks and long-term temporal consistency to refine the former predictions, mitigating the flow estimation error.
        However, such implicit modeling may suffer from limited VCOD data.

        \subsubsection{End-to-end VCOD framework}
        This framework emerges as a solution to address the limitations of two-step approaches, particularly the cumulative errors stemming from intermediate stages and the weak generalization ability caused by limited training data. By integrating feature extraction, motion modeling, and segmentation into a unified network, this framework aims to minimize error propagation and maximize the utilization of scarce data, thereby enhancing robustness and performance.
        To extend the previous version~\cite{ZoomNet-CVPR2022} which zooms in and out on images with a shared triplet feature encoder, ZoomNeXt~\cite{pang2024zoomnextunifiedcollaborativepyramid} implements image-video unified framework, and furtherly integrates multi-head scale integration and rich granularity perception, enhancing the structural representation and discrimination.
        IMEX~\cite{hui2024implicit} unifies implicit and explicit motion learning in a cohesive framework, achieving inter-frame alignment and consistency preserving of camouflaged objects respectively.
        TSP-SAM~\cite{hui2024endow} and EMIP~\cite{zhang2024explicit} are both prompt learning-based methods for VCOD. The difference is that TSP-SAM utilizes frozen SAM embedded with temporal-spatial injection, motion-driven self-prompt learning and long-range consistency to learn reliable visual prompts, while EMIP adopts two-stream architecture, incorporating segmentation-to-motion and motion-to-segmentation prompts. Notably, although EMIP handles motion cues explicitly, it is not a two-step method, as it simultaneously conducts optical flow estimation and VCOD by interactive prompting.
        These methods demonstrate the evolving trend towards holistic, unified frameworks that not only address the shortcomings of traditional two-step approaches but also push the boundaries of performance and efficiency in VCOD.


        \begin{figure*}[ht]
        {\includegraphics[width=\textwidth]{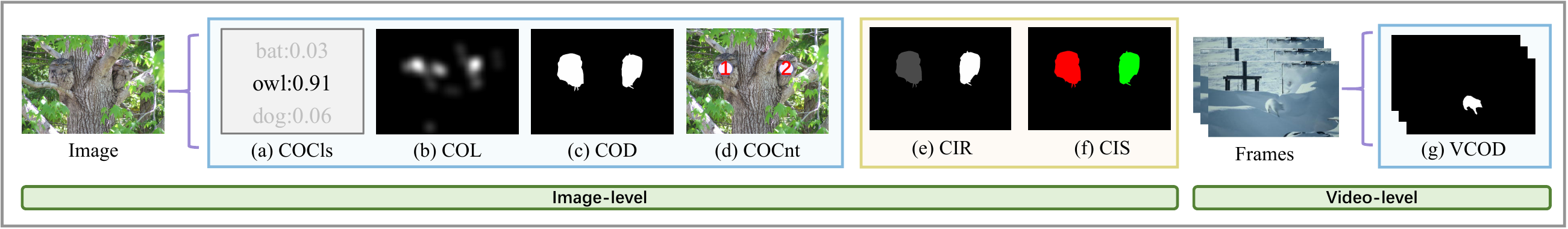}}
        \caption{A depiction of COD, VCOD, and other camouflaged scenario tasks. Five of these in the blue box are object-level tasks: (a) Camouflaged objects classification (COCLs), (b) Camouflaged objects localization (COL), (c) Camouflaged object detection (COD), (d) Camouflaged instance count (COCnt), and (g) Video-level camouflaged object detection (VCOD).
        The remaining two in the yellow box are instance-level tasks: (e) Camouflaged instance rank (CIR) and (f) Camouflaged instance segmentation (CIS). For further elaboration on these tasks, refer to Section \ref{section:other-csu}.} 
        \vspace{-4mm}
        \label{fig:CSU-tasks}
        \end{figure*}

        \section{Other camouflaged scenario tasks}
        \label{section:other-csu}

        Beyond the fundamental tasks of COD and VCOD, the domain of concealed scene understanding has evolved to encompass a wider array of high-level semantic tasks. These tasks aim to provide a deeper comprehension of camouflaged objects, extending from their classification to the generation of new camouflaged images. This section delves into the following advanced tasks, each addressing unique aspects of concealed scene understanding. The detail descriptions are illustrated in Fig. \ref{fig:CSU-tasks}.
        
        \noindent\textbf{Camouflaged objects classification (COCls)}
        focuses on distinguishing between different types of camouflaged objects. This task involves categorizing camouflaged objects into predefined or never seen classes, \textit{i.e.}, zero-shot ~\cite{10234216} and open-vocabulary~\cite{OVCOS_ECCV2024} learning, to handle the subtle differences and similarities among various camouflaged entities. Many COD methods integrate COCls~\cite{xing2023pretrainadaptdetectmultitask,nguyen2024art}, training with datasets that are labeled with specific categories~\cite{fan2022concealed}, for better performance. Accurate classification of camouflaged objects can aid in better understanding and various ecological studies.
        
        \noindent\textbf{Camouflaged objects localization (COL)}
        aims to identify the most detectable regions of camouflaged objects, and further localize the discriminative regions that make the camouflaged object stand out. Lv \textit{et al.}~\cite{yunqiu_cod21, Lv2022TowardDU} is the first to propose this task and leverage an eye tracker to record human gaze patterns, pinpointing salient discriminative regions. Simultaneously, they relabel existing datasets with fixation annotation, providing a valuable resource for training and evaluating models. This task enhances the ability to pinpoint critical areas within a camouflaged scene, which is crucial for applications in wildlife monitoring, and search and rescue operations. 

        \noindent\textbf{Camouflaged instance count (COCnt)}
        is focused on quantifying the number of camouflaged objects within a given scene, even in complex environments where objects may overlap or partially obscure each other. 
        Sun \textit{et al.}~\cite{sun2023indiscernible} introduce a correlated task,  indiscernible object counting (IOC), to count objects that blend seamlessly with their surroundings. To tackle this challenge, they propose a unified framework, IOCFormer, integrating density-based~\cite{wang2020distribution} and regression-based~\cite{liang2022end} counting methods. Due to the scarcity of suitable datasets, they also created IOCfish5K, full of high-resolution images for underwater IOC, with dense annotations.
        This emerging and promising task helps in assessing population densities and ensuring comprehensive area surveillance.
        
        \noindent\textbf{Camouflaged instance rank (CIR)}
        addresses the challenge of ranking multiple camouflaged instances based on specific criteria such as visibility, detectability, or relevance. With the introduction of the COL task, Lv \textit{et al.} propose the CIR task, which is based on the difficulty level of camouflage. To this end, the proposed \textit{CAM-FR}~\cite{yunqiu_cod21} and \textit{CAM-LDR}~\cite{Lv2022TowardDU} datasets also include ranking labels. Furthermore, they devise a triple-task learning framework that simultaneously localizes, segments and ranks camouflaged objects, efficiently utilizing the inner correlation among COD, COL, and CIR.

        \noindent\textbf{Camouflaged instance segmentation (CIS)}
        is a more granular task that involves segmenting individual camouflaged objects, \textit{i.e.}, instances, from their background and from each other. 
        CIS is proposed by Le et al.~\cite{le2021camouflaged}, who also introduce CIS dataset, \textit{i.e.}, \textit{CAMO++}, by extending the previous \textit{CAMO} dataset. They also conduct camouflage fusion learning, which fuses existing instance segmentation models, \textit{e.g.},  Cascade Mask RCNN~\cite{cai2018cascade},  by learning to predict the best model per image.
        To better break the deceptive camouflage, Luo et al.~\cite{luo2023camouflaged} propose a framework with a pixel-level camouflage decoupling module and an instance-level camouflage suppression module.
        Pei et al.~\cite{pei2022osformer} introduce the first one-stage transformer in CIS by fusing local features and long-range context dependencies.
        Recently, Vu et al.~\cite{vu2023leveraging} leverage text-to-image diffusion~\cite{rombach2022high} and CLIP~\cite{radford2021learning} for CIS, while utilizing open-vocabulary capabilities to learn multi-scale textual-visual features.
        This level of detail allows for more precise identification and interaction with each object, which is essential for applications that require accurate object differentiation, such as advanced robotic vision, detailed ecological studies, and targeted medical imaging.
        

        
        \section{Experiments}

        
        \subsection{Datasets}
        \label{section:dataset}

        Datasets play a pivotal role in the development and evaluation of COD algorithms. In this subsection, we provide an overview of prominent datasets relevant to COD tasks, categorized into image-level and video-level datasets. Tab.~\ref{tab:dataset} summarizes essential information along with their respective links for access. Additionally, Fig.~\ref{fig:cod-dataset} and Fig.~\ref{fig:vcod-dataset} showcases exemplar images from these datasets, offering a visual insight into the challenges posed by COD.
        
        \begin{table*}[!htb]
        \centering
        \caption{The basic information of both image-level and video-level COD datasets. 
        \textcolor{blue}{Level}: data type of dataset, \textit{i.e.}, Image(I) and Video(V). 
        \textcolor{blue}{Train/Test}: number of samples for training/testing (\textit{e.g.}, images for image dataset or frames for video dataset). \textcolor{blue}{N. Cam.}: whether collecting non-camouflaged samples. \textcolor{blue}{Type}: object categories of datasets. 
        \textcolor{blue}{Cls.}: whether providing classification labels for COCls. \textcolor{blue}{Fix.}: whether providing fixation annotation for COL. \textcolor{blue}{B. Box}: whether providing bounding box labels. \textcolor{blue}{Obj.}: whether providing object-level segmentation masks. \textcolor{blue}{Ins.}: whether providing instance-level segmentation masks for COL. \textcolor{blue}{Ran.}: whether providing ranking labels for CIR. \textcolor{blue}{Scr.}: whether providing weakly-supervised labels in scribbled form. \textcolor{blue}{Gro.}: whether providing corresponding category annotation within group images for CoCOD. \textcolor{blue}{Uns.}: whether providing unseen classes for OVCOS. \textcolor{blue}{Ref.}: whether providing referring images for RefCOD.  
        Clicking on the dataset will redirect you to its download link.}
        \label{tab:dataset}
        \resizebox{\textwidth}{!}{%
        \begin{tabular}{crcc|ccc|cc|cccccccccc}
        \toprule[1.5pt]
        & & & &
        \multicolumn{3}{c|}{\textbf{\textit{Statistics}}} &
        \multicolumn{2}{c|}{\textbf{\textit{Objects}}} &
          \multicolumn{10}{c}{\textbf{\textit{Labels}}} 
        \\ 
        \cline{5-19} 
        \multirow{-2}{*}{\textit{\textbf{Index}}} &
        \multirow{-2}{*}{\textbf{\textit{Dataset}}} &
        \multirow{-2}{*}{\textbf{\textit{Pub./Year}}} &
        \multirow{-2}{*}{\textbf{\textit{Level}}} &
          Total &
          Train &
          Test &
          N.Cam. &
          Type &
          Cls. &
          Fix. &
          B.Box &
          Obj. &
          Ins. &
          Ran. &
          Scr. &
          Gro. &
          Uns. &
          Ref. 
          \\
        
        \midrule[1pt]
        1 &
          ~\href{http://vis-www.cs.umass.edu/motionSegmentation/}{\textit{CAD2016}}\cite{bideauECCV16} &
          $ECCV_{16}$ &
          V &
          \multicolumn{1}{c}{836} &
          \multicolumn{1}{c}{0} &
          836 &
          \multicolumn{1}{c}{---} &
          animals &
          \multicolumn{1}{c}{\checkmark} &
          \multicolumn{1}{c}{---} &
          \multicolumn{1}{c}{---} &
          \multicolumn{1}{c}{\checkmark} &
          \multicolumn{1}{c}{---} &
          \multicolumn{1}{c}{---} &
          \multicolumn{1}{c}{---} &
          \multicolumn{1}{c}{---} &
          \multicolumn{1}{c}{---} &
          --- 
          \\ 
        2 &
          ~\href{https://www.polsl.pl/rau6/chameleon-database-animal-camouflage-analysis/}{\textit{CHAMELEON}}\cite{skurowski2018animal} &
          $Webpage_{17}$ &
          I &
          \multicolumn{1}{c}{76} &
          \multicolumn{1}{c}{0} &
          76 &
          \multicolumn{1}{c}{---} &
          animals &
          \multicolumn{1}{c}{---} &
          \multicolumn{1}{c}{---} &
          \multicolumn{1}{c}{---} &
          \multicolumn{1}{c}{\checkmark} &
          \multicolumn{1}{c}{---} &
          \multicolumn{1}{c}{---} &
          \multicolumn{1}{c}{---} &
          \multicolumn{1}{c}{---} &
          \multicolumn{1}{c}{---} &
          --- 
          
          \\ 
        3 &
          \href{https://github.com/xfflyer/Camouflaged-people-detection}{\textit{CPD1K}}\cite{CPD1K} &
          $SPL_{18}$ &
          I &
          \multicolumn{1}{c}{1000} &
          \multicolumn{1}{c}{0} &
          1000 &
          \multicolumn{1}{c}{---} &
          people &
          \multicolumn{1}{c}{---} &
          \multicolumn{1}{c}{---} &
          \multicolumn{1}{c}{---} &
          \multicolumn{1}{c}{\checkmark} &
          \multicolumn{1}{c}{---} &
          \multicolumn{1}{c}{---} &
          \multicolumn{1}{c}{---} &
          \multicolumn{1}{c}{---} &
          \multicolumn{1}{c}{---} &
          --- 
          
          \\ 
        4 &
          ~\href{https://sites.google.com/view/ltnghia/research/camo}{\textit{CAMO-COCO}}\cite{ltnghia-CVIU2019} &
          $CVIU_{19}$ &
          I &
          \multicolumn{1}{c}{2500} &
          \multicolumn{1}{c}{2000} &
          500 &
          \multicolumn{1}{c}{\checkmark} &
          \begin{tabular}[t]{@{}c@{}}animals\\ \&humans\end{tabular} &
          \multicolumn{1}{c}{---} &
          \multicolumn{1}{c}{---} &
          \multicolumn{1}{c}{---} &
          \multicolumn{1}{c}{\checkmark} &
          \multicolumn{1}{c}{---} &
          \multicolumn{1}{c}{---} &
          \multicolumn{1}{c}{---} &
          \multicolumn{1}{c}{---} &
          \multicolumn{1}{c}{---} &
          --- 
          
          \\ 
        5 &
          ~\href{https://www.robots.ox.ac.uk/~vgg/data/MoCA/}{\textit{MoCA}}\cite{Lamdouar20} &
          $ACCV_{20}$ &
          V &
          \multicolumn{1}{c}{37250} &
          \multicolumn{1}{c}{0} &
          37250 &
          \multicolumn{1}{c}{---} &
          animals &
          \multicolumn{1}{c}{\checkmark} &
          \multicolumn{1}{c}{---} &
          \multicolumn{1}{c}{\checkmark} &
          \multicolumn{1}{c}{---} &
          \multicolumn{1}{c}{---} &
          \multicolumn{1}{c}{---} &
          \multicolumn{1}{c}{---} &
          \multicolumn{1}{c}{---} &
          \multicolumn{1}{c}{---} &
          --- 
          
          \\ 
        6 &
          ~\href{https://github.com/JingZhang617/COD-Rank-Localize-and-Segment}{\textit{NC4K}}\cite{yunqiu_cod21} &
          $CVPR_{21}$ &
          I &
          \multicolumn{1}{c}{4121} &
          \multicolumn{1}{c}{0} &
          4121 &
          \multicolumn{1}{c}{---} &
          animals &
          \multicolumn{1}{c}{---} &
          \multicolumn{1}{c}{---} &
          \multicolumn{1}{c}{---} &
          \multicolumn{1}{c}{\checkmark} &
          \multicolumn{1}{c}{\checkmark} &
          \multicolumn{1}{c}{---} &
          \multicolumn{1}{c}{---} &
          \multicolumn{1}{c}{---} &
          \multicolumn{1}{c}{---} &
          --- 
          
          \\ 
        7 &
          ~\href{https://github.com/JingZhang617/COD-Rank-Localize-and-Segment}{\textit{CAM-FR}}\cite{yunqiu_cod21} &
          $CVPR_{21}$ &
          I &
          \multicolumn{1}{c}{2280} &
          \multicolumn{1}{c}{2000} &
          280 &
          \multicolumn{1}{c}{---} &
          \begin{tabular}[t]{@{}c@{}}animals\\ \&humans\end{tabular} &
          \multicolumn{1}{c}{---} &
          \multicolumn{1}{c}{\checkmark} &
          \multicolumn{1}{c}{---} &
          \multicolumn{1}{c}{\checkmark} &
          \multicolumn{1}{c}{\checkmark} &
          \multicolumn{1}{c}{\checkmark} &
          \multicolumn{1}{c}{---} &
          \multicolumn{1}{c}{---} &
          \multicolumn{1}{c}{---} &
          --- 
          
          \\ 
        8 &
          ~\href{https://xueliancheng.github.io/SLT-Net-project/}{\textit{MoCA-Mask}}\cite{cheng2022implicit} &
          $CVPR_{22}$ &
          V &
          \multicolumn{1}{c}{22939} &
          \multicolumn{1}{c}{19313} &
          3626 &
          \multicolumn{1}{c}{---} &
          animals &
          \multicolumn{1}{c}{\checkmark} &
          \multicolumn{1}{c}{---} &
          \multicolumn{1}{c}{\checkmark} &
          \multicolumn{1}{c}{\checkmark} &
          \multicolumn{1}{c}{---} &
          \multicolumn{1}{c}{---} &
          \multicolumn{1}{c}{---} &
          \multicolumn{1}{c}{---} &
          \multicolumn{1}{c}{---} &
          --- 
          
          \\ 
        9 &
          ~\href{https://sites.google.com/view/ltnghia/research/camo_plus_plus}{\textit{CAMO++}}\cite{ltnghia-TIP2022} &
          $TIP_{22}$ &
          I &
          \multicolumn{1}{c}{5500} &
          \multicolumn{1}{c}{3500} &
          2000 &
          \multicolumn{1}{c}{\checkmark} &
          \begin{tabular}[t]{@{}c@{}}animals\\ \&humans\end{tabular} &
          \multicolumn{1}{c}{\checkmark} &
          \multicolumn{1}{c}{---} &
          \multicolumn{1}{c}{\checkmark} &
          \multicolumn{1}{c}{\checkmark} &
          \multicolumn{1}{c}{\checkmark} &
          \multicolumn{1}{c}{---} &
          \multicolumn{1}{c}{---} &
          \multicolumn{1}{c}{---} &
          \multicolumn{1}{c}{---} &
          --- 
          
          \\ 
        10 &
          ~\href{https://dengpingfan.github.io/pages/COD.html}{\textit{COD10K}}\cite{fan2022concealed} &
          $TPAMI_{22}$ &
          I &
          \multicolumn{1}{c}{10000} &
          \multicolumn{1}{c}{6000} &
          4000 &
          \multicolumn{1}{c}{\checkmark} &
          \begin{tabular}[t]{@{}c@{}}animals\\ \&humans\end{tabular} &
          \multicolumn{1}{c}{\checkmark} &
          \multicolumn{1}{c}{---} &
          \multicolumn{1}{c}{\checkmark} &
          \multicolumn{1}{c}{\checkmark} &
          \multicolumn{1}{c}{\checkmark} &
          \multicolumn{1}{c}{---} &
          \multicolumn{1}{c}{---} &
          \multicolumn{1}{c}{---} &
          \multicolumn{1}{c}{---} &
          --- 
          
          \\ 
        11 &
          ~\href{https://drive.google.com/file/d/1u7PRtZDu2vXCRe0o2SplVYa7ESoZQFR-/view?usp=sharing}{\textit{S-COD}}\cite{S-COD} &
          $AAAI_{23}$ &
          I &
          \multicolumn{1}{c}{4040} &
          \multicolumn{1}{c}{4040} &
          0 &
          \multicolumn{1}{c}{---} &
          \begin{tabular}[t]{@{}c@{}}animals\\ \&humans\end{tabular} &
          \multicolumn{1}{c}{---} &
          \multicolumn{1}{c}{---} &
          \multicolumn{1}{c}{---} &
          \multicolumn{1}{c}{---} &
          \multicolumn{1}{c}{---} &
          \multicolumn{1}{c}{---} &
          \multicolumn{1}{c}{\checkmark} &
          \multicolumn{1}{c}{---} &
          \multicolumn{1}{c}{---} &
          --- 
          
          \\ 
        12 &
          ~\href{https://github.com/JingZhang617/COD-Rank-Localize-and-Segment}{\textit{CAM-LDR}}\cite{Lv2022TowardDU} &
          $TCSVT_{23}$ &
          I &
          \multicolumn{1}{c}{6066} &
          \multicolumn{1}{c}{4040} &
          2026 &
          \multicolumn{1}{c}{---} &
          \begin{tabular}[t]{@{}c@{}}animals\\ \&humans\end{tabular} &
          \multicolumn{1}{c}{---} &
          \multicolumn{1}{c}{\checkmark} &
          \multicolumn{1}{c}{---} &
          \multicolumn{1}{c}{\checkmark} &
          \multicolumn{1}{c}{\checkmark} &
          \multicolumn{1}{c}{\checkmark} &
          \multicolumn{1}{c}{---} &
          \multicolumn{1}{c}{---} &
          \multicolumn{1}{c}{---} &
          --- 
          
          \\ 
        13 &
          ~\href{https://github.com/dengpingfan/csu}{\textit{CDS2K}}\cite{fan2023csu} &
          $VI_{23}$ &
          I &
          \multicolumn{1}{c}{2492} &
          \multicolumn{1}{c}{0} &
          2492 &
          \multicolumn{1}{c}{\checkmark} &
          \begin{tabular}[t]{@{}c@{}}industrial\\ defect\end{tabular} &
          \multicolumn{1}{c}{\checkmark} &
          \multicolumn{1}{c}{---} &
          \multicolumn{1}{c}{\checkmark} &
          \multicolumn{1}{c}{\checkmark} &
          \multicolumn{1}{c}{---} &
          \multicolumn{1}{c}{---} &
          \multicolumn{1}{c}{---} &
          \multicolumn{1}{c}{---} &
          \multicolumn{1}{c}{---} &
          --- 
          
          \\ 
        14 &
          ~\href{https://github.com/liumaozhen-lmz/Military-Camouflage-MHCD2022}{\textit{MHCD2022}}\cite{LIU2023126466} &
          $Neuro_{23}$ &
          I &
          \multicolumn{1}{c}{3000} &
          \multicolumn{1}{c}{2400} &
          600 &
          \multicolumn{1}{c}{---} &
          military &
          \multicolumn{1}{c}{\checkmark} &
          \multicolumn{1}{c}{---} &
          \multicolumn{1}{c}{\checkmark} &
          \multicolumn{1}{c}{---} &
          \multicolumn{1}{c}{---} &
          \multicolumn{1}{c}{---} &
          \multicolumn{1}{c}{---} &
          \multicolumn{1}{c}{---} &
          \multicolumn{1}{c}{---} &
          --- 
          
          \\ 
        15 &
          ~\href{https://github.com/zc199823/BBNet--CoCOD}{\textit{CoCOD8K}}\cite{zhang2023collaborative} &
          $TNNLS_{23}$ &
          I &
          \multicolumn{1}{c}{8528} &
          \multicolumn{1}{c}{5933} &
          2595 &
          \multicolumn{1}{c}{---} &
          \begin{tabular}[t]{@{}c@{}}animal\\\&human\end{tabular} &
          \multicolumn{1}{c}{\checkmark} &
          \multicolumn{1}{c}{---} &
          \multicolumn{1}{c}{---} &
          \multicolumn{1}{c}{\checkmark} &
          \multicolumn{1}{c}{---} &
          \multicolumn{1}{c}{---} &
          \multicolumn{1}{c}{---} &
          \multicolumn{1}{c}{\checkmark} &
          \multicolumn{1}{c}{---} &
          --- 
          
          \\ 
        
        16 &
        ~\href{https://github.com/zhangxuying1004/RefCOD}{\textit{R2C7K}}\cite{zhang2023referring} &
        $arXiv_{23}$ &
        I &
        \multicolumn{1}{c}{6615} &
        \multicolumn{1}{c}{---} &
        --- &
          
        \multicolumn{1}{c}{\checkmark} &
        \begin{tabular}[t]{@{}c@{}}animals\\ \&humans\end{tabular} &
        
        \multicolumn{1}{c}{\checkmark} &
        \multicolumn{1}{c}{---} &
        \multicolumn{1}{c}{---} &
        \multicolumn{1}{c}{\checkmark} &
        \multicolumn{1}{c}{---} &
        \multicolumn{1}{c}{---} &
        \multicolumn{1}{c}{---} &
        \multicolumn{1}{c}{---} &
        \multicolumn{1}{c}{---} &
        \checkmark 
        
        \\
        17 &
        ~\href{https://github.com/lartpang/OVCamo?tab=readme-ov-file}{\textit{OVCamo}}\cite{OVCOS_ECCV2024} &
        $ECCV_{24}$ &
        I &
        \multicolumn{1}{c}{11483} &
        \multicolumn{1}{c}{7713} &
        3770 &
        
        \multicolumn{1}{c}{---} & 
        \begin{tabular}[t]{@{}c@{}}animal\\\&human\end{tabular}&
        \multicolumn{1}{c}{\checkmark} &
        \multicolumn{1}{c}{---} &
        \multicolumn{1}{c}{---} &
        \multicolumn{1}{c}{\checkmark} &
        \multicolumn{1}{c}{---} &
        \multicolumn{1}{c}{---} &
        \multicolumn{1}{c}{---} &
        \multicolumn{1}{c}{---} &
        \multicolumn{1}{c}{\checkmark} &
        --- 
        
        \\ 
        \bottomrule[1.5pt]
        \end{tabular}%
        }
        \end{table*}

        \subsubsection{Characteristic COD datasets}
        \noindent \textbf{\textit{CHAMELEON}\cite{skurowski2018animal}} is a small-scale, unpeer-reviewed dataset consisting of 76 camouflaged images collected from the internet using the keyword ``camouflaged animals''. Each image is manually annotated with labels focusing on animals camouflaged within complex ecological backgrounds. This dataset is typically used as a test dataset for model performance evaluation.
        
        \noindent \textbf{\textit{CAMO-COCO}\cite{ltnghia-CVIU2019}} consists of 2,500 images across eight categories, created by merging the camouflaged dataset \textit{CAMO} with the non-camouflaged dataset \textit{MS-COCO}\cite{cocolin2014microsoft}, each contributing 1,250 images. In \textit{CAMO-COCO}, 80\% of the images are designated for training and the remaining for testing. \textit{CAMO} includes both natural camouflaged objects, such as animals, and artificial camouflaged objects, such as human beings, featuring seven challenging attributes that complicate detection and segmentation.

        \begin{figure*}[ht]
        {\includegraphics[width=\textwidth]{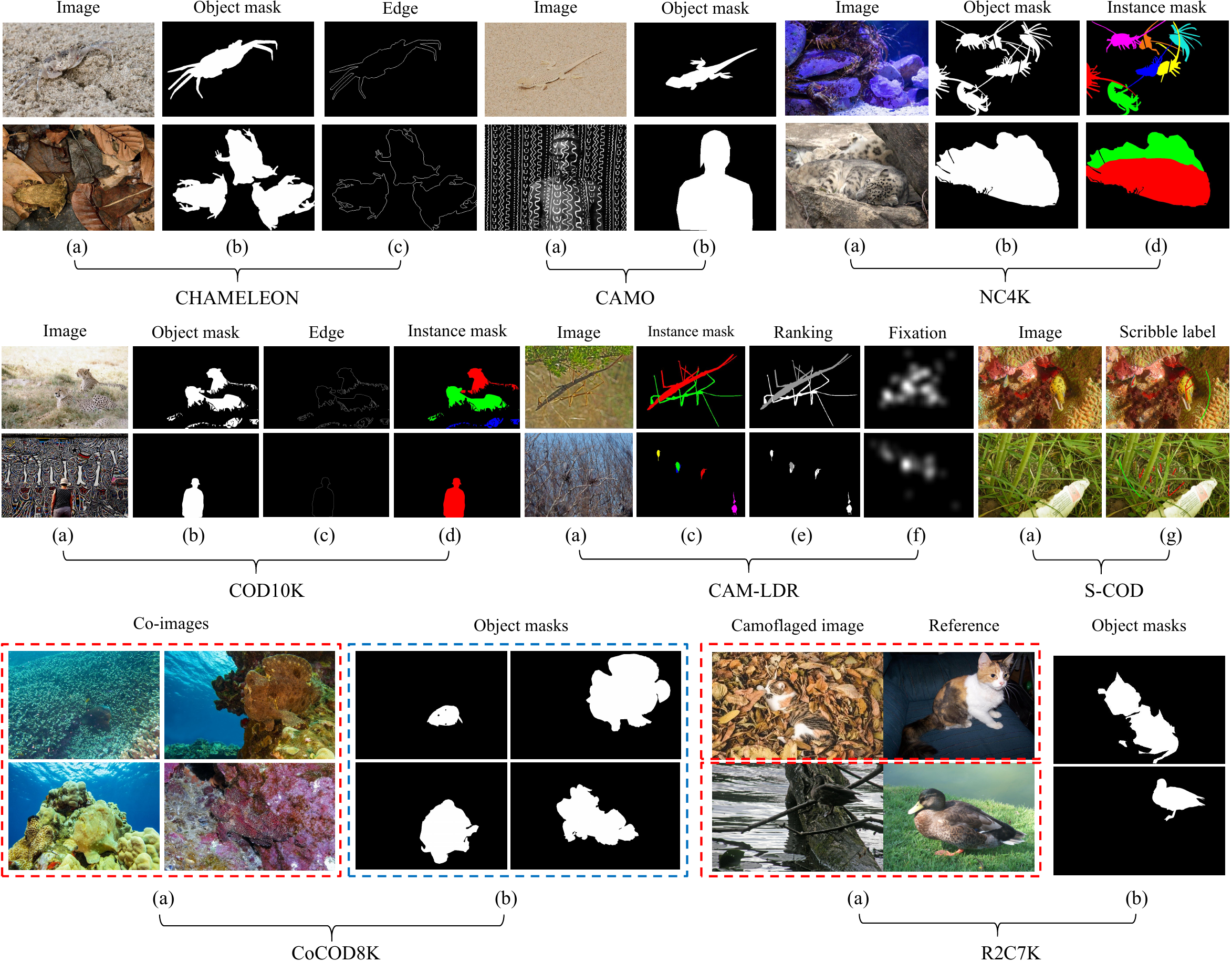}}
        \caption{Examples from eight characteristic COD datasets. Left to right: (a) RGB images, (b) Object-level ground-truths, (c) Edge maps (d) Instance-level ground-truths, (d) Edge maps, (e) Ranking maps, (f) fixation
        maps and (g) Scribble annotation. The red and blue boxes indicate a set of inputs and outputs, respectively. Section \ref{section:dataset} for more details}
        \label{fig:cod-dataset}
        \vspace{-2mm}
        \end{figure*}
        
        \noindent \textbf{\textit{NC4K}\cite{yunqiu_cod21}}, the largest image-level COD test dataset currently available, contains 4,121 camouflaged scene images sourced from the internet, each annotated at both object and instance levels. The dataset covers predominantly natural scenes along with some artificial camouflage, making it a preferred choice for evaluating the generalization capabilities of COD models.
        
        \noindent \textbf{\textit{COD10K}\cite{fan2022concealed}} includes 10 superclasses and 78 subclasses, totaling 10,000 images, with 60\% designated as training data. Within this largest image-level COD dataset to date, there are 5,066 camouflaged images (3,040 for training and 2,026 for testing), 1,934 non-camouflaged images, and 3,000 background images. All camouflaged images are densely annotated with categories, bounding boxes, and object and instance level labels, supporting a wide range of research tasks, \textit{e.g.}, COD, COS, and CIS. The high-quality annotations and diversity of \textit{COD10K} make it an essential dataset for COD research.
        
        \noindent \textbf{\textit{S-COD}\cite{S-COD}} is the first dataset created for weakly supervised learning based on scribble annotations. It uses a tagging process that relies on initial impressions to outline the rough structure of objects, including both foreground and background. The dataset comprises 3,040 images from the \textit{COD10K}\cite{fan2022concealed} training set and 1,000 images from the \textit{CAMO}\cite{ltnghia-CVIU2019} training set, totaling 4,040 samples. Compared to pixel-level annotations, the annotations in \textit{S-COD} are simpler and more efficient.
        
        \noindent \textbf{\textit{CAM-LDR}\cite{Lv2022TowardDU}} facilitates research on COL and CIR by recording the time taken to detect camouflaged instances with an eye tracker, which correlates with the difficulty of detection. This dataset includes 4,040 training images, derived from the \textit{CAMO}\cite{ltnghia-TIP2022} and \textit{COD10K}\cite{fan2022concealed} training sets, and 2,026 testing images from the \textit{COD10K} testing set. Detection times are used to rank camouflaged objects into six categories: background, easy, three medium levels, and hard. This approach offers a novel metric for understanding the detectability of camouflaged objects.
        
        \noindent \textbf{\textit{CoCOD8K}\cite{zhang2023collaborative}}
        is the first dataset for CoCOD, which comprises 8,528 images reorganized from four COD datasets—\textit{CHAMELEON}\cite{skurowski2018animal}, \textit{CAMO}\cite{ltnghia-CVIU2019}, \textit{COD10K}\cite{fan2022concealed}, and \textit{NC4K}\cite{yunqiu_cod21}. This dataset includes diverse natural and artificial scenes, classified into 5 superclasses and 70 subclasses, each annotated with object masks and category labels. Designed to foster research in detecting co-camouflaged objects across grouped images, \textit{CoCOD8K} also filters images to fit specific criteria, supporting robust model training with 5,933 training and 2,595 test images.

        \noindent \textbf{\textit{R2C7K}\cite{zhang2023referring}} encompasses 6,615 images across 64 categories from real-world scenarios to facilitate RefCOD. It consists of a Camo-subset with 5,015 camouflaged images, primarily derived from \textit{COD10K}\cite{fan2022concealed}, and a Ref-subset with 1,600 images of salient objects, uniformly sourced with 25 per category from Flickr and Unsplash webpages with no copyright disputes. For research, a referring split is provided where each category in the Ref-subset has 20 images for training and 5 for testing, while the distribution in the Camo-subset follows the original split from \textit{COD10K} with additional samples from \textit{NC4K} to ensure at least 6 samples in each category.
        
        \noindent \textbf{\textit{OVCamo}\cite{OVCOS_ECCV2024}} advances the task of OVCOS by offering a robust dataset featuring 11,483 images across 75 object classes, derived from merged public datasets such as \cite{bideauECCV16,cheng2022implicit,fan2022concealed,PlantCamo,CPD1K}. This dataset uniquely tackles semantic ambiguities by redefining annotation standards, which ensures clear and distinct class definitions to improve segmentation accuracy. For realistic performance evaluation, OVCamo divides its dataset by allocating 14 seen classes to the training set, while the remaining 61 unseen classes are reserved for testing. This distribution maintains a training-to-testing sample ratio of 7:3, simulating real-world application challenges and ensuring a rigorous assessment of model generalization.

        In current research practices, the most representative datasets typically selected for experiments of image-level COD models are \textit{CHAMELEON}\cite{skurowski2018animal}, \textit{CAMO-COCO}\cite{ltnghia-CVIU2019}, \textit{COD10K}\cite{fan2022concealed}, and \textit{NC4K}\cite{yunqiu_cod21}. The common setup\cite{fan2022concealed} for training includes 1,000 images from \textit{CAMO} and 3,040 images from \textit{COD10K}. The remaining parts of these datasets are used to test the generalization ability and viability of the models. To maintain consistency in our analysis, we will adopt this setup in our subsequent performance comparison of image-level COD models.

        \subsubsection{Characteristic VCOD Datasets}

        \noindent \textbf{\textit{CAD2016}\cite{bideauECCV16}} is composed of nine short video sequences sourced from YouTube, total having 836 frames. Each sequence is manually annotated every five frames. The camouflaged objects in these images are exclusively biological entities found in natural settings.
        
        \noindent \textbf{\textit{MoCA}\cite{Lamdouar20}}  is currently the largest dataset for camouflaged animal detection in video format. It consists of 141 video sequences, also sourced from YouTube, representing 67 different categories of animals found in natural scenarios. The dataset spans over 37250 frames and 26 minutes of video content. Annotations include PWC-Net optical flow data for each frame and bounding boxes with motion labels provided every five frames, with linear interpolation used for the intervening frames.

        \begin{figure*}[ht]
        {\includegraphics[width=\textwidth]{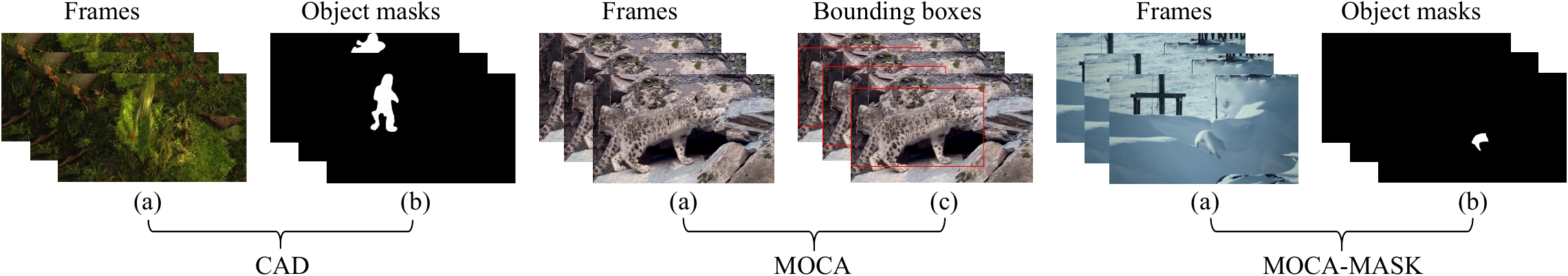}}
        \caption{Examples from three characteristic VCOD datasets. Left to right: (a) RGB Frames, (b) Object-level ground-truths, (c) Bounding boxes. Section \ref{section:dataset} for more details.}
        \label{fig:vcod-dataset}
        \end{figure*}
        
        \noindent \textbf{\textit{MoCA-Mask}\cite{cheng2022implicit}} is an extension of \textit{MoCA}\cite{Lamdouar20} and includes 87 video sequences with a total of 22,939 frames, following the removal of irrelevant scenes. This dataset enhances \textit{MoCA} by providing manually annotated masks every five frames, resulting in 4,691 bounding boxes and pixel-level masks. The dataset is divided into a training set that comprises 71 videos (19,313 frames), along with a testing set consisting of 16 videos (3,625 frames).
        
        
        \subsection{Evaluation metrics}
        We evaluate existing COD models using four common evaluation metrics as recommended in\cite{fan2022concealed}. These metrics, \textit{i.e.}, 
        S-measure (${S_\alpha }$)\cite{Cheng2021sMeasure}, F-measure (${F_\beta }$)\cite{6909433}, Mean Absolute Error (MAE)\cite{6247743}, and E-measure (${E_\phi }$)\cite{Fan2018Enhanced}, provide a comprehensive assessment of performance. Here, we detail these evaluation metrics:
        
        \noindent \textbf{Precision-Recall (PR)} curve is generated by transforming the input into a binary mask \(M\), which is segmented across a range of thresholds from 0 to 255. Precision (\(P\)) and Recall (\(R\)) are calculated by comparing the binary mask \(M\) with the ground truth mask (\(G\)) at each threshold, producing the PR curve. 
        \(P\) and \(R\) can be calculated as
        \begin{equation}
        \begin{aligned}
            P &= \frac{|M(T) \cap G|}{|M(T)|}, \quad 
            R &= \frac{|M(T) \cup G|}{|G|}
        \end{aligned}
        \end{equation}
        where \(M(T)\) is the mask obtained by thresholding the non-binary prediction map at threshold \(T\).
        
        \noindent \textbf{S-measure (${S_\alpha }$)} quantifies the spatial structural similarity between the predicted map (\(C\)) and the ground truth (\(G\)). It combines both object-aware (\(S_o\)) and region-aware (\(S_r\)) assessments with the following definition:
        \begin{equation}
        S_{\alpha} = \alpha S_o + (1 - \alpha)S_r,
        \end{equation}
        where $\alpha  \in \left[ {0,1} \right]$ is a weighting factor, typically set at 0.5, that balances the contribution of \(S_o\) and \(S_r\).
        
        \noindent \textbf{F-measure (${F_\beta }$)} is used to calculate the relationship between Precision (\(P\)) and Recall (\(R\)). Initially, the input is transformed into a binary mask, \(M\), segmented over a range of thresholds from 0 to 255. \(P\) and \(R\) are calculated by comparing \(M\) with \(G\) across these thresholds. The formulas for \(P\) and \(R\) are defined as follows:
        \begin{equation}
        \begin{aligned}
            P &= \frac{|M(T) \cap G|}{|M(T)|}, \quad 
            R &= \frac{|M(T) \cup G|}{|G|},
        \end{aligned}
        \end{equation}
        where \(M(T)\) represents the binary mask obtained by thresholding the non-binary prediction map at threshold \(T\), and \(|\cdot|\) denotes the total area of the mask.
        But ${F_\beta }$ further demonstrates the average harmonic mean value between them. The formula for \(F_{\beta}\) is defined as
        \begin{equation}
        F_{\beta} = \frac{(\beta^2 + 1)PR}{\beta^2P + R},
        \end{equation}
        with \(\beta^2\) typically set to 0.3. From the range of thresholds, three variants of \(F_{\beta}\) are computed: the maximum (\(F_{\beta}^{\max}\)), the mean (\(F_{\beta}^{\text{mean}}\)) and the adaptive (\(F_{\beta}^{\text{adj}}\)). Besides, (\(F_{\beta}^{\omega}\)) is also a  widely used metric where the \(P\) and \(R\) are both weighted averages. \(F_{\beta}^{\text{mean}}\) is adopted in this paper.
        

        \noindent \textbf{E-measure (${E_\phi }$)} evaluates both the local and global similarity between \(C\) and \(G\). It is defined as follows:
        \begin{equation}
        E_{\phi} = \frac{1}{W \times H} \sum_{x=1}^W \sum_{y=1}^H \phi[C(x,y), G(x,y)],
        \end{equation}
        where \(\phi\) represents an enhanced alignment matrix, and \(W\) and \(H\) are the width and height of the input image, respectively. \(E_{\phi}\) also provides three indicative values: maximum (\(E_{\phi}^{\max}\)), mean (\(E_{\phi}^{\text{mean}}\)), and adaptive (\(E_{\phi}^{\text{adj}}\)). \(E_{\phi}^{\text{mean}}\) is adopted for evaluation in this survey.

        \noindent \textbf{Mean Absolute Error (\(M\))} quantifies the average absolute difference per pixel between the normalized predicted map $C$ and the ground truth map $G$, where $C, G \in [0,1]$. The mean absolute error \(M\) is formulated as follows:
        \begin{equation}
        M = \frac{1}{H \times W} \sum_{x=1}^H \sum_{y=1}^W |C(x,y) - G(x,y)|,
        \end{equation}
        where $W$ and $H$ denote the width and height of the input image, and $(x, y)$ represents the pixel coordinates. Unlike ${F_\beta }$, ${E_\phi }$ and ${S_\alpha }$, a lower \(M\) suggests a more accurate model. 

        For VCOD, 
        mean Dice (mDice)~\cite{he2023strategic} for similarity evaluation and mean IoU (mIoU) for overlap measurement are also used for evaluation. Notice that larger mDice and mIoU scores indicate better performance.


         \begin{table*}[!t]
        \centering
        \caption{\textbf{Quantitative comparison among COD methods on the \textit{CHAMELEON}, \textit{CAMO}, \textit{COD10K} and \textit{NC4K} testing datasets, where the competing approaches are classified based on their use of convolution-based or transformer-based backbone. {\color[HTML]{FF0000}\textbf{Red}}  and {\color[HTML]{00B0F0}\textbf{blue}} indicate the best and the second best within each category, respectively.}}
        \label{tab:cos_benchmark}
        \scriptsize
        \renewcommand{\arraystretch}{1}
        \resizebox{\textwidth}{!}{
        \begin{tabular}{l|c|c|cccc|cccc|cccc|cccc}
        \toprule[1.5pt]
        \multirow{2}{*}{Model} & \multirow{2}{*}{Pub./Year} & \multirow{2}{*}{Backbone} & \multicolumn{4}{c}{\textit{CHAMELEON}} & \multicolumn{4}{c}{\textit{CAMO}-test}& \multicolumn{4}{c}{\textit{COD10K}-test} & \multicolumn{4}{c}{\textit{NC4K}}\\ 
         &  &  & \cellcolor{gray!40}$S_{\alpha}\uparrow$ & \cellcolor{gray!40}$F_\beta\uparrow$ & \cellcolor{gray!40}$E_\phi\uparrow$  & \cellcolor{gray!40}$M\downarrow$ & \cellcolor{gray!40}$S_{\alpha}\uparrow$ & \cellcolor{gray!40}$F_\beta\uparrow$ & \cellcolor{gray!40}$E_\phi\uparrow$  & \cellcolor{gray!40}$M\downarrow$ & \cellcolor{gray!40}$S_{\alpha}\uparrow$ & \cellcolor{gray!40}$F_\beta\uparrow$ & \cellcolor{gray!40}$E_\phi\uparrow$  & \cellcolor{gray!40}$M\downarrow$ & \cellcolor{gray!40}$S_{\alpha}\uparrow$ & \cellcolor{gray!40}$F_\beta\uparrow$ & \cellcolor{gray!40}$E_\phi\uparrow$  & \cellcolor{gray!40}$M\downarrow$ \\ 
         \midrule[1pt]
        \multicolumn{19}{c}{Convolution-based backbone}\\ 
        \midrule[1pt]
        UR-SINet\cite{kajiura2021improving} & ${ACM MM}_{21}$ & ResNet50 & 0.876  & 0.824  & 0.942  & 0.031  & 0.741  & 0.649  & 0.804  & 0.091  & 0.775  & 0.643  & 0.869  & 0.041  & 0.806  & 0.731  & 0.873  & 0.057  \\ 
        LSR\cite{yunqiu_cod21} & ${CVPR}_{21}$ & ResNet50 & 0.842  & 0.794  & 0.896  & 0.046  & 0.708  & 0.645  & 0.755  & 0.105  & 0.760  & 0.658  & 0.831  & 0.045  & 0.797  & 0.758  & 0.854  & 0.061  \\ 
        PFNet\cite{Mei_2021_CVPR} & ${CVPR}_{21}$ & ResNet50 & 0.882  & 0.828  & 0.931  & 0.033  & 0.782  & 0.746  & 0.841  & 0.085  & 0.800  & 0.701  & 0.877  & 0.040  & 0.829  & 0.784  & 0.887  & 0.053  \\ 
        MGL\cite{9577564} & ${CVPR}_{21}$ & ResNet50 & 0.893  & 0.833  & 0.917  & 0.031  & 0.775  & 0.726  & 0.812  & 0.088  & 0.814  & 0.711  & 0.852  & 0.035  & 0.833  & 0.782  & 0.867  & 0.052  \\ 
        C$^{2}$F-Net\cite{sun2021c2fnet} & ${IJCAI}_{21}$ & Res2Net50 & 0.888  & 0.828  & 0.935  & 0.032  & 0.796  & 0.762  & 0.854  & 0.080  & 0.813  & 0.723  & 0.890  & 0.036  & 0.838  & 0.795  & 0.897  & 0.049  \\ 
        TINet\cite{Zhu_Zhang_Zhang_Liu_2021} & ${AAAI}_{21}$ & ResNet50 & 0.874  & 0.783  & 0.916  & 0.038  & 0.781  & 0.728  & 0.836  & 0.087  & 0.793  & 0.679  & 0.861  & 0.042  & 0.829  & 0.773  & 0.879  & 0.055  \\ 
        BASNet\cite{DBLP:journals/corr/abs-2101-04704} & ${arXiv}_{21}$ & ResNet34 & 0.847  & 0.795  & 0.883  & 0.044  & 0.615  & 0.503  & 0.671  & 0.124  & 0.661  & 0.486  & 0.729  & 0.071  & 0.696  & 0.610  & 0.762  & 0.095  \\ 
        UGTR\cite{yang2021uncertainty} & ${ICCV}_{21}$ & ResNet50 & 0.888  & 0.819  & 0.910  & 0.031  & 0.785  & 0.738  & 0.823  & 0.086  & 0.818  & 0.712  & 0.853  & 0.035  & 0.839  & 0.787  & 0.874  & 0.052  \\ 
        FAP-Net\cite{zhou2022feature} & ${TIP}_{22}$ & Res2Net50 & 0.893  & 0.842  & 0.940  & 0.028  & 0.815  & 0.776  & 0.865  & 0.076  & 0.822  & 0.731  & 0.888  & 0.036  & 0.851  & 0.810  & 0.899  & 0.047  \\ 
        FindNet\cite{9923635} & ${TIP}_{22}$ & Res2Net & 0.895  & 0.845  & 0.944  & 0.027  & 0.803 & 0.763 & 0.862  & 0.077  & 0.811  & 0.706  & 0.883  & 0.036  & 0.841  & 0.802  & 0.895  & 0.048  \\ 
        PreyNet\cite{10.1145/3503161.3548178} & ${ACM MM}_{22}$ & ResNet50 & 0.895  & 0.859  & 0.951  & 0.028  & 0.790  & 0.757  & 0.842  & 0.077  & 0.813  & 0.736  & 0.881  & 0.034  & 0.834  & 0.803  & 0.887  & 0.050  \\ 
        BGNet\cite{sun2022bgnet} & ${IJCAI}_{22}$ & Res2Net50 & 0.901  & 0.860  & 0.943  & 0.027  & 0.812  & 0.789  & 0.870  & 0.073  & 0.831  & 0.753  & 0.901  & 0.033  & 0.851  & 0.820  & 0.907  & 0.044  \\ 
        C$^{2}$F-Net-V2\cite{chen2022camouflaged} & ${TCSVT}_{22}$ & Res2Net50 & 0.893  & 0.836  & 0.947  & 0.028  & 0.799  & 0.770  & 0.859  & 0.077  & 0.811  & 0.725  & 0.887  & 0.036  & 0.840  & 0.802  & 0.896  & 0.048  \\ 
        SegMaR\cite{Jia_2022_CVPR} & ${CVPR}_{22}$ & ResNet50 & \color[HTML]{00B0F0}0.906  & \color[HTML]{00B0F0}0.872  & 0.951  & 0.025  & 0.815  & 0.795  & 0.874  & 0.071  & 0.833  & 0.757  & 0.899  & 0.034  & 0.841  & 0.821  & 0.896  & 0.046  \\ 
        ZoomNet\cite{ZoomNet-CVPR2022} & ${CVPR}_{22}$ & ResNet50 & 0.902  & 0.864  & 0.943  & \color[HTML]{00B0F0}0.023  & 0.820  & 0.794  & 0.877  & 0.066  & 0.838  & \color[HTML]{00B0F0}0.766  & 0.888  & \color[HTML]{00B0F0}0.029  & 0.853  & 0.818  & 0.896  & 0.043  \\ 
        OCENet\cite{Liu_2022_WACV} & ${WACV}_{22}$ & ResNet50 & 0.901  & 0.843  & 0.940  & 0.028  & 0.802  & 0.766  & 0.852  & 0.080  & 0.827  & 0.741  & 0.894  & 0.033  & 0.853  & 0.818  & 0.902  & 0.045  \\ 
        ERRNet\cite{ji2022fast} & ${PR}_{22}$ & ResNet50 & 0.877  & 0.825  & 0.927  & 0.036  & 0.779  & 0.729  & 0.842  & 0.085  & 0.786  & 0.675  & 0.867  & 0.043  & 0.827  & 0.778  & 0.887  & 0.054  \\ 
        SINet-V2\cite{fan2022concealed} & ${TPAMI}_{22}$ & Res2Net50 & 0.888  & 0.835  & 0.942  & 0.030  & 0.820  & 0.782  & 0.882  & 0.070  & 0.815  & 0.718  & 0.887  & 0.037  & 0.847  & 0.805  & 0.903  & 0.048  \\ 
        MRR-Net\cite{10180211} & ${TNNLS}_{23}$ & ResNet50 & 0.882  & 0.821  & 0.931  & 0.033  & 0.811  & 0.772  & 0.869  & 0.076  & 0.822  & 0.730  & 0.889  & 0.036  & 0.848  & 0.801  & 0.898  & 0.049  \\ 
        UJSCOD-V2\cite{li2023jointsalientobjectdetection} & ${arXiv}_{23}$ & ResNet50 & 0.892  & 0.848  & 0.948  & 0.025  & 0.803  & 0.768  & 0.858  & 0.071  & 0.817  & 0.733  & 0.895  & 0.033  & 0.856  & 0.824  & \color[HTML]{00B0F0}0.913  & \color[HTML]{FF0000}0.040  \\ 
        PUENet\cite{zhang:hal-04142929} & ${TIP}_{23}$ & ResNet50 & 0.888  & 0.844  & 0.943  & 0.030  & 0.794  & 0.762  & 0.857  & 0.080  & 0.813  & 0.727  & 0.887  & 0.035  & 0.836  & 0.798  & 0.892  & 0.050  \\ 
        WS-SAM\cite{he2023weaklysupervised} & ${NeurIPS}_{23}$ & ResNet50 & 0.824  & 0.777  & 0.897  & 0.046  & 0.759  & 0.742  & 0.818  & 0.092  & 0.803  & 0.719  & 0.878  & 0.038  & 0.829  & 0.802  & 0.886  & 0.052  \\ 
        FEDER\cite{He2023Camouflaged} & ${CVPR}_{23}$ & ResNet50 & 0.894  & 0.855  & 0.947  & 0.028  & 0.807  & 0.785  & 0.873  & 0.069  & 0.823  & 0.740  & 0.900  & 0.032  & 0.846  & 0.817  & 0.905  & 0.045  \\ 
        PopNet\cite{10377562} & ${ICCV}_{23}$ & Res2net50 & \color[HTML]{FF0000}0.917  & \color[HTML]{FF0000}0.885  & \color[HTML]{FF0000}0.965  & \color[HTML]{FF0000}0.020  & 0.808  & 0.784  & 0.859  & 0.077  & \color[HTML]{FF0000}0.851  & \color[HTML]{FF0000}0.786  & \color[HTML]{00B0F0}0.910  & \color[HTML]{FF0000}0.028  & \color[HTML]{00B0F0}0.861  & \color[HTML]{00B0F0}0.833  & 0.909  & 0.042  \\ 
        DGNet\cite{ji2023gradient} & ${MIR}_{23}$ & EfficientNet & 0.890  & 0.834  & 0.938  & 0.029  & \color[HTML]{FF0000}0.839  & \color[HTML]{FF0000}0.806  & \color[HTML]{FF0000}0.901  & \color[HTML]{FF0000}0.057  & 0.822  & 0.728  & 0.896  & 0.033  & 0.857  & 0.814  & 0.911  & 0.042  \\ 
        LSR-V2\cite{Lv2022TowardDU} & ${TCSVT}_{23}$ & ResNet50 & 0.878  & 0.828  & 0.929  & 0.034  & 0.789  & 0.751  & 0.840  & 0.079  & 0.805  & 0.711  & 0.880  & 0.037  & 0.840  & 0.801  & 0.896  & 0.048  \\ 
        Camouflageator\cite{he2023strategic} & ${ICLR}_{24}$ & ResNet50 & 0.903  & 0.863  & \color[HTML]{00B0F0}0.952  & 0.026  & \color[HTML]{00B0F0}0.829  & \color[HTML]{00B0F0}0.805  & \color[HTML]{00B0F0}0.891  & \color[HTML]{00B0F0}0.066  & \color[HTML]{00B0F0}0.843  & 0.763  & \color[HTML]{FF0000}0.920  & \color[HTML]{FF0000}0.028  & \color[HTML]{FF0000}0.869  & \color[HTML]{FF0000}0.835  & \color[HTML]{FF0000}0.922  & \color[HTML]{00B0F0}0.041  \\ 
        \midrule[1pt]
        \multicolumn{19}{c}{Transformer-based backbone} \\ 
        \midrule[1pt]
        CamoFormer\cite{yin2022camoformermaskedseparableattention} & ${TPAMI}_{24}$ & Swin-B & 0.898  & 0.867  & 0.945  & 0.025  & 0.876  & 0.856  & \color[HTML]{00B0F0}0.930  & \color[HTML]{00B0F0}0.043  & 0.838  & 0.753  & 0.916  & 0.029  & 0.888  & 0.863  & 0.937  & 0.031  \\ 
        FRINet\cite{10.1145/3581783.3611773} & ${ACM MM}_{23}$ & ViT & 0.905 & 0.871 & 0.949 & 0.023 & 0.865  & 0.848  & 0.924  & 0.046  & 0.864  & 0.810  & 0.930  & 0.023  & 0.889  & 0.866  & 0.937  & \color[HTML]{00B0F0}0.030  \\ 
        UCOS-DA\cite{zhang2023unsupervisedcamouflagedobjectsegmentation} & ${ICCVW}_{23}$ & DINO & 0.715  & 0.629  & 0.802  & 0.095  & 0.701  & 0.646  & 0.784  & 0.127  & 0.689  & 0.546  & 0.740  & 0.086  & 0.755  & 0.689  & 0.819  & 0.085  \\ 
        PUENet\cite{zhang:hal-04142929} & ${TIP}_{23}$ & ViT & 0.910  & 0.869  & 0.957  & 0.022  & \color[HTML]{00B0F0}0.878  & 0.860  & \color[HTML]{00B0F0}0.930  & 0.045  & \color[HTML]{00B0F0}0.873  & 0.812  & \color[HTML]{FF0000}0.938  & 0.022  & \color[HTML]{00B0F0}0.898  & \color[HTML]{00B0F0}0.874  & \color[HTML]{FF0000}0.945  & \color[HTML]{FF0000}0.028  \\ 
        OAFormer\cite{10219627} & ${ICME}_{23}$ & PVTv2 & 0.904  & 0.868  & 0.961  & 0.023  & 0.867  & 0.849  & 0.924  & 0.048  & 0.860  & 0.795  & 0.927  & 0.025  & 0.883  & 0.857  & 0.934  & 0.032  \\ 
        XMSNet\cite{wu2023object} & ${ACM MM}_{23}$ & PVT & 0.904  & 0.895  & 0.950  & 0.025  & 0.864  & \color[HTML]{FF0000}0.871  & 0.923  & 0.048  & 0.861  & \color[HTML]{00B0F0}0.828  & 0.927  & 0.024  & 0.879  & \color[HTML]{FF0000}0.877  & 0.933  & 0.034  \\ 
        FSNet\cite{10103836} & ${TIP}_{23}$ & SwinT & 0.905  & 0.868  & \color[HTML]{00B0F0}0.963  & 0.022  & \color[HTML]{FF0000}0.880  & \color[HTML]{00B0F0}0.861  & \color[HTML]{FF0000}0.933  & \color[HTML]{FF0000}0.041  & 0.870  & 0.810  & \color[HTML]{FF0000}0.938  & 0.023  & 0.891  & 0.866  & \color[HTML]{00B0F0}0.940  & 0.031  \\ 
        FSPNet\cite{Huang2023Feature} & ${CVPR}_{23}$ & ViT & 0.908  & 0.867  & 0.943  & 0.023  & 0.857  & 0.830  & 0.899  & 0.050  & 0.851  & 0.769  & 0.895  & 0.026  & 0.879  & 0.843  & 0.915  & 0.035  \\ 
        HitNet\cite{Hu_Wang_Qin_Dai_Ren_Luo_Tai_Shao_2023} & ${AAAI}_{23}$ & PVTv2 & \color[HTML]{00B0F0}0.921  & \color[HTML]{00B0F0}0.900  & \color[HTML]{FF0000}0.967  & \color[HTML]{FF0000}0.019  & 0.849  & 0.831  & 0.906  & 0.055  & 0.871  & 0.823  & \color[HTML]{00B0F0}0.935  & 0.023  & 0.875  & 0.853  & 0.926  & 0.037  \\ 
        MLKG\cite{cheng2023largemodelbasedreferring} & ${arXiv}_{23}$ & SAM, CLIP & \color[HTML]{FF0000}0.935 & \color[HTML]{FF0000}0.923 & 0.941 & \color[HTML]{00B0F0}0.020 & 0.828 & 0.806 & 0.877 & 0.075 & \color[HTML]{FF0000}0.910 & \color[HTML]{FF0000}0.838 & 0.916 & \color[HTML]{FF0000}0.019 & \color[HTML]{FF0000}0.900 & 0.872 & 0.918 & 0.036 \\
        VSCode\cite{luo2024vscode} & ${CVPR}_{24}$ &Swin-S & --- & --- & --- & ---  & 0.873 & 0.844 &0.925 &0.046 &0.869 &0.806 &0.931 &0.023 &0.891 &0.863 &0.935 &0.032\\
        CamoFocus\cite{10483928} & ${WACV}_{24}$ & PVTv2 & 0.912  & 0.884  & 0.957  & 0.023  & 0.873  & 0.861  & 0.926  & \color[HTML]{00B0F0}0.043  & \color[HTML]{00B0F0}0.873  & 0.818  & \color[HTML]{00B0F0}0.935  & \color[HTML]{00B0F0}0.021  & 0.889  & 0.870  & 0.936  & \color[HTML]{00B0F0}0.030  \\ 
        GenSAM\cite{hu2023relax} & ${AAAI}_{24}$ & CLIP, BLIP2 & 0.764  & 0.680  & 0.807  & 0.090  & 0.719  & 0.659  & 0.775  & 0.113  & 0.775  & 0.681  & 0.838  & 0.067 & --- & --- & --- & --- \\ 
        \toprule[1.5pt]
    \end{tabular}
    }
\end{table*}

        \subsection{Quantitative analysis}
        \begin{table*}[t!]
        \centering
        \caption{\textbf{Quantitative comparison among VCOD methods on \textit{CAD2016} and \textit{MoCA-Mask} testing dataset, where the competing approaches are classified based on the type of input data they utilize, into image-based and video-based methods. {\color[HTML]{FF0000}\textbf{Red}} and {\color[HTML]{00B0F0}\textbf{blue}} indicate the best and the second best within each category, respectively.}}
        \label{tab:vcos_benchmark_cad}
        \scriptsize
        \renewcommand{\arraystretch}{1}
        \resizebox{\textwidth}{!}{
        \begin{tabular}{l|c|c|cccccc|cccccc}
        \toprule[1.5pt]
            \multirow{2}{*}{Method} & \multirow{2}{*}{Pub./Year} & \multirow{2}{*}{Backbone} &\multicolumn{6}{c}{\textit{CAD2016}} &\multicolumn{6}{c}{\textit{MoCA-Mask}}\\
            
            & & & \cellcolor{gray!40}${S_\alpha }\uparrow$ & \cellcolor{gray!40}${F_{\beta}^{\omega}}\uparrow$ & \cellcolor{gray!40}${E_{\phi}}\uparrow$ & \cellcolor{gray!40}$M\downarrow$ & \cellcolor{gray!40}$mDice\uparrow$ & \cellcolor{gray!40}$mIoU\uparrow$ & \cellcolor{gray!40}${S_\alpha }\uparrow$ & \cellcolor{gray!40}${F_{\beta}^{\omega}}\uparrow$ & \cellcolor{gray!40}${E_{\phi}}\uparrow$ & \cellcolor{gray!40}$M\downarrow$ & \cellcolor{gray!40}$mDice\uparrow$ & \cellcolor{gray!40}$mIoU\uparrow$ \\ 
            \midrule[1pt]
            \multicolumn{15}{c}{Image-based methods}\\
            \midrule[1pt]
            SINet\cite{fan2020camouflaged} & $CVPR_{20}$ & ResNet50 & 0.601 & 0.204 & 0.589 & 0.089 & 0.289 & 0.209 & 0.574 & 0.185 & \color[HTML]{FF0000}0.655 & \color[HTML]{00B0F0}0.030 & 0.221 & 0.156\\ 
            SINet-V2\cite{fan2022concealed} & $TPAMI_{22}$ & Res2Net50 &0.544 &0.181 &0.546 &\color[HTML]{00B0F0}0.049 &0.170 &0.110 &0.571 &0.175 &0.608 &0.035 &0.211 &0.153\\
            ZoomNet\cite{ZoomNet-CVPR2022} & $CVPR_{22}$ & ResNet50 &0.587 &0.225 &0.594 &0.063 &0.246 &0.166 &0.582 &\color[HTML]{FF0000}0.211 &0.536 &0.033 &0.224 &\color[HTML]{00B0F0}0.167\\ 
            BGNet\cite{sun2022bgnet} & $IJCAI_{22}$ & Res2Net50 &\color[HTML]{00B0F0}0.607 &0.203 &0.666 &0.089 &0.345 &0.256 &\color[HTML]{00B0F0}0.590 &\color[HTML]{FF0000}0.203 &\color[HTML]{00B0F0}0.647 &0.023 &0.225 &\color[HTML]{00B0F0}0.167\\ 
            FEDER\cite{He2023Camouflaged} & $CVPR_{23}$ & ResNet50 &\color[HTML]{00B0F0}0.607 &\color[HTML]{00B0F0}0.246 &\color[HTML]{00B0F0}0.725 &0.061 &\color[HTML]{00B0F0}0.361 &\color[HTML]{00B0F0}0.257 &0.555 &0.158 &0.542 &0.049 &0.192 &0.132\\
            FSPNet\cite{Huang2023Feature} & $CVPR_{23}$ & ViT &0.539 &0.220 &0.553 &0.145 &0.309 &0.212 &\color[HTML]{FF0000}0.594 &0.182 &0.608 &0.044 &\color[HTML]{00B0F0}0.238 &\color[HTML]{00B0F0}0.167\\ 
            PUENet\cite{zhang:hal-04142929} & $TIP_{23}$ & ViT &\color[HTML]{FF0000}0.673 &\color[HTML]{FF0000}0.427 &\color[HTML]{FF0000}0.803 &\color[HTML]{FF0000}0.034 &\color[HTML]{FF0000}0.499 &\color[HTML]{FF0000}0.389 &\color[HTML]{FF0000}0.594 &\color[HTML]{00B0F0}0.204 &0.619 &0.037 &\color[HTML]{FF0000}0.302 &\color[HTML]{FF0000}0.212\\ 
            \midrule[1pt]
            \multicolumn{15}{c}{Video-based methods}\\
            \midrule[1pt]
            RCRNet\cite{yan2019semi} & $ICCV_{19}$ & ResNet50 &0.627 &0.287 &0.666 &0.048 &0.309 &0.229 &0.597 &0.174 &0.583 &0.025 &0.194 &0.137\\ 
            PNS-Net\cite{ji2021progressively} & $MICCAI_{21}$ & ResNet50 &0.678 &0.369 &0.720 &0.043 &0.409 &0.308 &0.576 &0.134 &0.562 &0.038 &0.189 &0.133\\ 
            MG\cite{yang2021self} & $ICCV_{21}$ & VGG-style &0.484 &0.314 &0.558 &0.370 &0.351 &0.260 &0.547 &0.165 &0.537 &0.095 &0.197 &0.141\\ 
            SLT-Net\cite{cheng2022implicit} & $CVPR_{22}$ & PVT &0.704 &0.524 &\color[HTML]{FF0000}0.912 &0.028 &0.543 &0.438 &0.656 &0.357 &\color[HTML]{00B0F0}0.785 &0.021 &0.387 &0.310\\ 
            ZoomNeXt\cite{pang2024zoomnextunifiedcollaborativepyramid} & $TPAMI_{24}$ & PVTv2 & \color[HTML]{FF0000}0.757 & \color[HTML]{00B0F0}0.593 & \color[HTML]{00B0F0}0.865 & 0.020 & \color[HTML]{FF0000}0.599 & \color[HTML]{FF0000}0.510 & \color[HTML]{00B0F0}0.734 & \color[HTML]{00B0F0}0.476 & 0.497 & \color[HTML]{00B0F0}0.010 & \color[HTML]{00B0F0}0.497 & \color[HTML]{FF0000}0.422\\ 
            TMNet\cite{yu2024tokenmotion} & $ICASSP_{24}$ & SegFormer & \color[HTML]{00B0F0}0.740 & 0.485 & 0.735 & \color[HTML]{FF0000}0.008 & 0.503 & 0.417 &\color[HTML]{FF0000}0.740 &\color[HTML]{FF0000}0.485 &0.735 &\color[HTML]{FF0000}0.008 &\color[HTML]{FF0000}0.503 &\color[HTML]{00B0F0}0.417\\ 
            IMEX\cite{hui2024implicit} & $TMM_{24}$ & ResNet50 &0.684 &0.452 &0.813 &0.033 &0.469 &0.370 &0.661 &0.371 &0.778 &0.020 &0.409 &0.319\\ 
            TSP-SAM\cite{hui2024endow} & $CVPR_{24}$ & SAM &0.704 &0.524 &\color[HTML]{FF0000}0.912 &0.028 &0.543 &0.438 &0.689 &0.444 &\color[HTML]{FF0000}0.808 &\color[HTML]{FF0000}0.008 &0.458 &0.388\\ 
            SAM-PM\cite{meeran2024sam} & $CVPRW_{24}$ & SAM &0.729 &\color[HTML]{FF0000}0.602 &0.746 &\color[HTML]{00B0F0}0.018 &\color[HTML]{00B0F0}0.594 &\color[HTML]{00B0F0}0.493 &0.695 &0.464 &0.732 &0.011 &\color[HTML]{00B0F0}0.497 &0.416\\ 
            EMIP\cite{zhang2024explicit} & $arXiv_{24}$ & PVTv2 &0.719 &0.514 &---  &0.028 &0.536 &0.425 &0.675 &0.381 &--- &0.015 &0.426 &0.333\\ 
            \toprule[1.5pt]
        \end{tabular}
        }
    \end{table*}

        \noindent\textbf{Results of deep COD models.} In this section, we conduct experiments on \textbf{39} cutting-edge techniques, covering six strategies mentioned earlier, and categorize them based on their backbones into two types: convolution-based and transformer-based. As shown in Tab.\ref{tab:cos_benchmark}.

        PopNet~\cite{10377562} stands out as a top performer, achieving the highest ${S_\alpha}$ and ${F_\beta}$ scores across multiple datasets. Its success is attributed to its innovative integration of source-free depth estimation and object-popping techniques, followed by the precise separation of objects from their contact surfaces. Overall, methods utilizing transformer-based backbones, such as FSNet~\cite{10103836} and HitNet~\cite{Hu_Wang_Qin_Dai_Ren_Luo_Tai_Shao_2023}, exhibit superior performance compared to those using convolution-based backbones. The self-attention mechanism inherent in transformers is highly effective in modeling long-range dependencies, which are crucial for accurate camouflaged object detection (COD). Furthermore, the results reveal that certain strategies are particularly effective for specific datasets. For example, methods incorporating multi-scale context information, such as C$^{2}$F-Net-V2~\cite{chen2022camouflaged}, tend to perform better on datasets with varying object sizes, such as \textit{COD10K}-test. Similarly, mechanism simulation strategies, such as LSR-V2~\cite{Lv2022TowardDU}, are beneficial for datasets with complex backgrounds, like \textit{CAMO}-test. Interestingly, GenSAM~\cite{hu2023relax}, despite incorporating advanced models like CLIP and BLIP2, does not consistently outperform other methods. This underscores that while powerful pre-trained models can be beneficial, their effectiveness also depends on their integration into the overall COD framework. Additionally, models designed for more challenging settings, such as UCOS-DA~\cite{zhang2023unsupervisedcamouflagedobjectsegmentation} for UCOS and MLKG~\cite{cheng2023largemodelbasedreferring} for RefCOD, do not always achieve outstanding performance. These models require more sophisticated approaches and larger datasets to generalize effectively. Consequently, some researchers are developing new datasets to train proposed models, such as R2CNet~\cite{zhang2023referring}, OVCoser~\cite{OVCOS_ECCV2024}, and BBNet~\cite{zhang2023collaborative}.
        
        
        \noindent\textbf{Results of deep VCOD models.} We compare \textbf{17} cutting-edge methods across two camouflaged video datasets, as detailed in Tab. \ref{tab:vcos_benchmark_cad}. This includes 7 image-based methods, 2 related video-oriented object segmentation methods, and 8 deep VCOD methods.
        
        ZoomNeXt~\cite{pang2024zoomnextunifiedcollaborativepyramid} stands out as the most effective VCOD approach, achieving state-of-the-art results in ${S_\alpha }$, $mDice$ and $mIoU$. This is mainly due to the zoom-in-and-out operation and scale integration for robust feature fusion and efficient temporal modeling in a unified framework. 
        Video-based methods overall outperform their image-based counterparts across several key metrics, which is attributed to their ability to exploit temporal consistency across frames, which helps distinguish camouflaged objects from their backgrounds even when they are moving at a high speed. 
        The end-to-end frameworks, \textit{e.g.}, TMNet~\cite{yu2024tokenmotion}, IMEX~\cite{hui2024implicit} and EMIP~\cite{zhang2024explicit}, demonstrate the efficacy of integrating temporal modeling directly into the network architecture. In contrast, two-stage frameworks, like SLT-Net~\cite{cheng2022implicit} and MG~\cite{yang2021self}, perform slightly worse, which follow an explicit or implicit motion-based approach, first estimating optical flow or pseudo masks before performing COD. While effective, this decoupled strategy may introduce errors that propagate through the pipeline. What's more, based on the powerful SAM, TSP-SAM~\cite{hui2024endow} and SAM-PM~\cite{meeran2024sam} also perform well, which showcases the remarkable spatial segmentation versatility and potential of SAM when applied to the challenging task of VCOD.

        
        \subsection{Qualitative analysis}
        \label{section:qualitative-image}
        \begin{figure*}[htb]
        \centering
        {\includegraphics[width=1.0\textwidth]{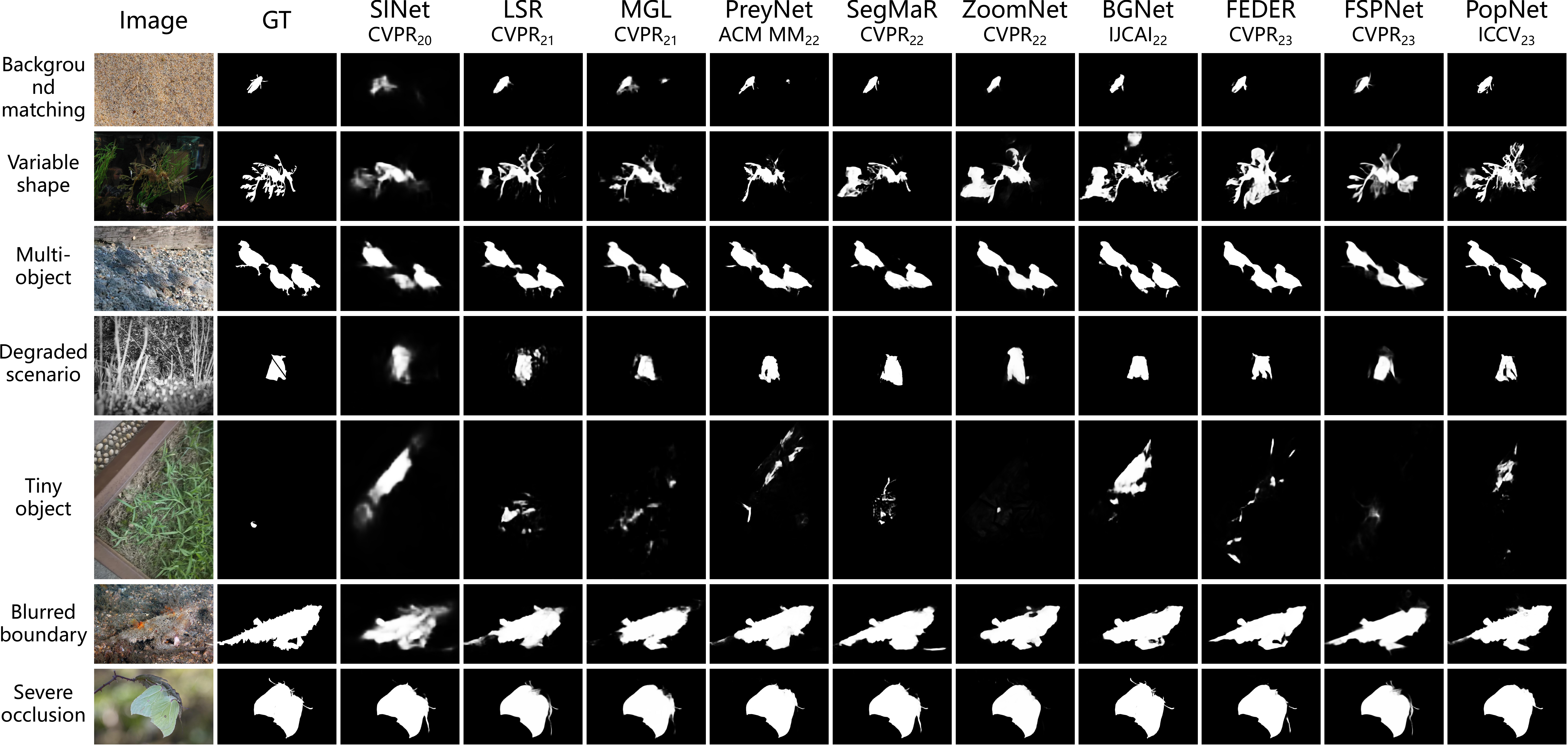}}
        \caption{Qualitative results of 10 image-level COD approaches. For more descriptions, please refer to section \ref{section:qualitative-image}.}
        \label{fig:exe-image-qualitative}
        \end{figure*}

        As depicted in Fig. \ref{fig:exe-image-qualitative}, we present a comprehensive visual comparison of \textbf{10} cutting-edge image-level COD methods.  We selected various challenging camouflaged images across \textbf{7} typical complex scenarios, including background matching, variable shape, multi-object environments, degraded scenarios, tiny objects, blurred boundaries, and severe occlusion. Additional, more challenging scenarios for COD are discussed in Section \ref{section:future-work}, and typical examples can also be seen in Fig. \ref{fig:difficult-cases}.

        In the background-matching scenario, where insects blend seamlessly with their surroundings, we observe that SINet\cite{fan2020camouflaged} struggles to identify insects, often resulting in missed detections. Conversely, SegMaR\cite{Jia_2022_CVPR} demonstrates robustness by effectively outlining insect contours, even under low-contrast conditions. In the variable shape scenario, such as with the leafy sea dragon, whose appendages resemble leaves, the adaptability of models is tested. While ZoomNet\cite{ZoomNet-CVPR2022} misidentifies parts of these large appendages as separate objects, FPNet\cite{cong2023frequencyperceptionnetworkcamouflaged} exhibits a superior ability to segment more complete instances, illustrating its robustness to shape variations.

        Under multi-object configurations, where the scene is crowded with numerous camouflaged entities, the models generally succeed in locating all targets but struggle with accurate segmentation. However, PopNet\cite{10377562} performs commendably, achieving better coverage without excessive false positives. Degraded scenarios, including poor lighting or blurring, significantly impact the localization and identification processes of the models. The detection results across all models fall short of expectations, indicating considerable room for improvement in these challenging conditions.

        For tiny objects, prediction precision degrades, with most methods failing to detect them accurately. However, ZoomNet\cite{ZoomNet-CVPR2022}, with its zoom-in-and-out operation, shows potential in highlighting even the smallest targets. In blurred boundary scenarios, defining precise contours becomes challenging. Here, FEDER\cite{He2023Camouflaged} stands out with its ODE-inspired edge reconstruction for complete edge prediction, whereas other methods often produce incomplete boundary predictions.
        Lastly, in severe occlusion scenarios, where two insects partially obscure each other, the results from all methods are generally acceptable. Nonetheless, BGNet\cite{sun2022bgnet} displays an improved capacity through its edge-guidance feature module, though further enhancement in detail detection is needed.

        In conclusion, while all cutting-edge methods demonstrate acceptable performance across various challenging COD scenarios, there are still notable limitations, particularly in degraded conditions, tiny objects, and blurred boundaries. Further research focusing on enhancing robustness and precision under these extreme conditions is crucial for advancing the field.

        \section{Future directions}
        \label{section:future-work}

        \begin{figure*}[ht]
        {\includegraphics[width=\textwidth]{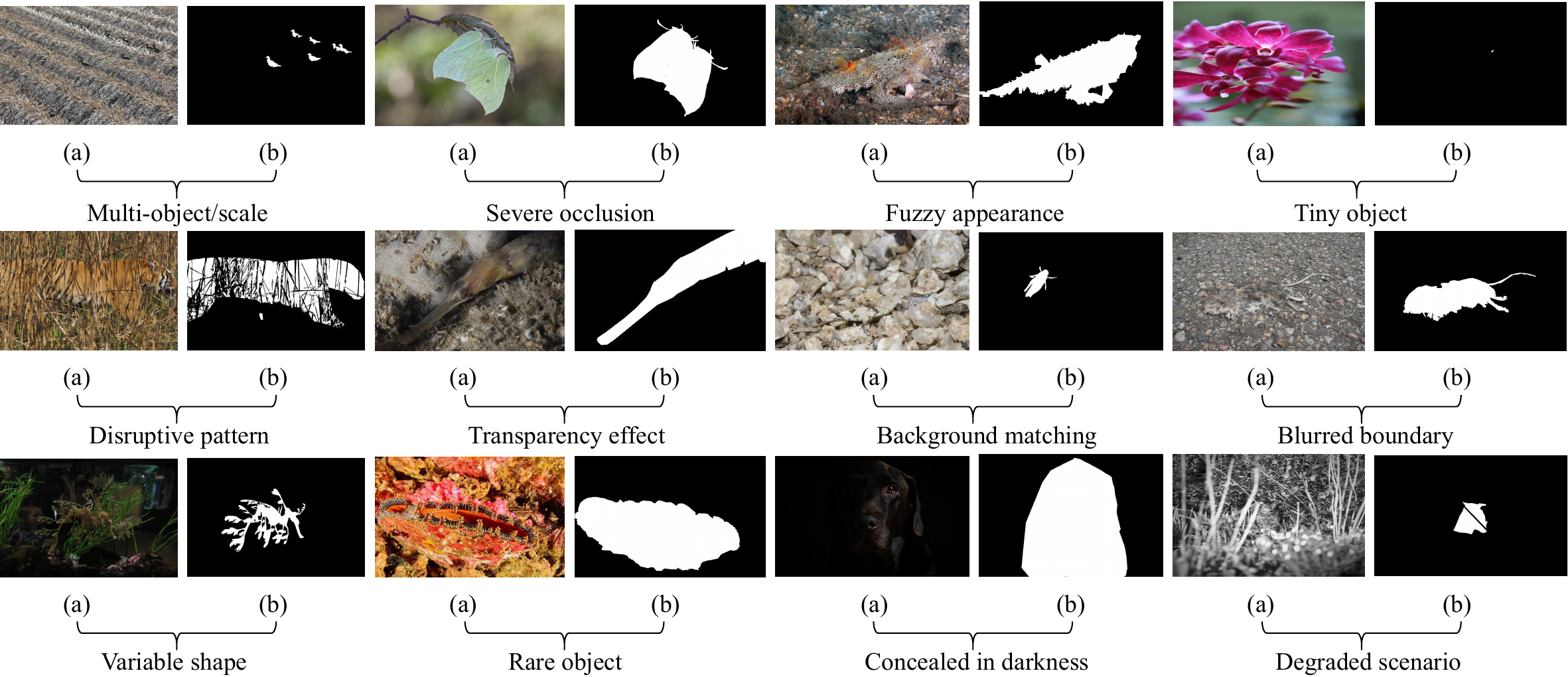}}
        \caption{Illustration of 12 exemplary challenging samples in extremely complex scenarios. Left to right: (a) RGB images, (b) Object-level ground-truths.}
        \label{fig:difficult-cases}
        \vspace{-2mm}
        \end{figure*}

        \subsection{Mitigating current issues}

        \noindent\textbf{Deep generative models for data scarcity.}
        To mitigate the scarcity of data, leveraging deep generative models to synthesize diverse, realistic camouflaged images will bolster training effectiveness by dataset augmentation~\cite{ma2023follow}, enhancing model robustness in dealing with camouflaged scenarios.
        With the rise of image generation models represented by GANs\cite{goodfellow2014generative} and diffusion models\cite{rombach2022high}, diverse and high-quality images can now be controlled and generated using other multimodal inputs such as text. This advancement can effectively address the longstanding issue of dataset scarcity in this field. By training with both generated camouflaged object data and the original datasets, performance can be significantly improved. 
        Adopting an adversarial manner, where the camouflaged sample generator and the segmentation network are pitted against each other, may help unlock further potential~\cite{he2023strategic}. However, there are no quantitative metrics to evaluate the camouflage degree and sample quality in deep datasets, making the widespread use of generated samples for training a topic of debate. For VCOD datasets, the generation of such data still has a long way to go due to the current immaturity and lack of a general framework in video generation technology.
        Diffusion models have become a research hotspot in the field of computer vision due to their stability in algorithm training and high-quality sample generation. 
        CamDiff\cite{luo2023camdiff} innovates by synthesizing salient objects within camouflage scenes, mitigating the scarcity of multi-pattern training data and enhancing robustness to salient misclassifications. The authors also use CamDiff to propose \textit{Diff-COD} dataset from the original COD datasets\cite{skurowski2018animal,ltnghia-CVIU2019,fan2022concealed,yunqiu_cod21} to enhance the robustness to saliency. Meanwhile, LAKE-RED\cite{zhao2024lake} tackles the limitations of dataset diversity and expensive data collection by automatically generating camouflage images without manual background specification, fostering scalability and interpretability. Both approaches emphasize the potential of diffusion models to address data-scarce vision tasks.

        \noindent\textbf{Tackling complex scenarios \& challenging samples.}
        Addressing the intricacies of COD requires grappling with various challenges posed by complex scenes and difficult samples. Key issues include extremely complex backgrounds and advanced camouflage techniques that render objects nearly invisible. Fig.~\ref{fig:difficult-cases} depicts some extremely concealed scenarios with challenging samples. Multi-object, multi-scale scenarios present difficulties in capturing objects of varying sizes and scales, while severely occluded objects demand robust algorithms capable of disentangling overlapping structures. Objects with fuzzy appearances, particularly small ones, challenge the limits of detection frameworks due to their minimal visual cues. Disruptive patterns, transparency, and background matching further complicate detection by seamlessly blending objects into their surroundings. Blurred boundaries and structural ambiguity add to the difficulty of delineating object edges, while variable shapes require flexible recognition capabilities.
        Moreover, detecting extremely rare camouflaged objects necessitates specialized methods to distinguish them from the vast majority of non-camouflaged instances. Objects concealed in darkness, affected by drastic illumination variations, pose unique challenges that require innovative approaches to handle varying light conditions. 
        Overcoming these obstacles demands the development of advanced detection models capable of adapting to diverse and extreme scenarios, ensuring continued progress and real-world applicability of COD research. Additionally, performance degradation in challenging environments, such as low-light conditions and foggy settings, exacerbates the difficulty of detecting camouflaged objects~\cite{he2023hqg,hu2021toward,hu2021blind}. Low light intensifies contrast issues, making camouflaged objects nearly invisible against their surroundings, while fog introduces noise and occlusion, further complicating feature extraction and localization. Enhancing robustness in such extreme conditions requires innovations in image enhancement techniques~\cite{fang2024real}, coupled with adaptive feature learning strategies specifically tailored for degraded images.
        In actual application scenarios, the above problems are very common, for example, in the agricultural domain, detecting numerous small and concealed crops amidst severe occlusions highlights the need for precision and robustness. To address this, Wang \textit{et al.}\cite{Wang_2024_CVPR} propose a depth-aware concealed crop detection method for dense agricultural scenes, along with the corresponding dataset ACOD-12K, specifically designed to tackle COD-related agricultural tasks.

        \noindent\textbf{Annotation-efficient learning in limited conditions.}
        In the context of COD under constrained conditions, annotation-efficient learning emerges as a crucial approach to addressing the scarcity of labeled data. Strategies such as few/zero-shot learning, weakly supervised learning, and open-world learning aim to alleviate the challenge of exhaustive annotation requirements. Specifically, few/zero-shot learning\cite{10234216,nguyen2024art} addresses the detection of unseen object classes by leveraging knowledge transfer from related categories. Weakly supervised learning\cite{he2023weaklysupervised, S-COD}, which relies on image-level labels to identify object locations, provides a less precise but still valuable alternative. Unsupervised learning\cite{zhang2023unsupervisedcamouflagedobjectsegmentation}, through techniques like clustering and representation learning, uncovers patterns in unlabeled data, thereby enhancing model generalization. Self-supervised learning further contributes by exploiting inherent data properties to generate pseudo-labels, fostering the development of robust representations. The limitation of training data also necessitates innovative data augmentation and domain adaptation techniques to combat overfitting. Open-world applications\cite{OVCOS_ECCV2024} introduce unique challenges, requiring models to detect novel objects while maintaining performance on known classes. Collectively, these approaches strive to overcome the difficulties posed by scarce annotations and unseen object classes, thereby pushing the boundaries of concealed object detection in practical, annotation-constrained scenarios.

        \noindent\textbf{Camouflage loss functions.}
        The introduction of camouflage-specific loss functions enhances the training process for COD by directing model optimization toward accurately distinguishing camouflaged regions from salient ones. For instance, SCLoss\cite{yang2024spatialcoherencelosssalient} shows promise by incorporating spatial coherence into the learning process for ambiguous regions. This approach significantly improves the network's ability to discern boundaries and subtle nuances in camouflaged objects. By focusing on both individual pixel responses and their mutual interactions, SCLoss addresses the limitations of single-response loss functions, providing a more comprehensive solution to the challenges posed by camouflaged objects.

        \noindent\textbf{Real-time performance constraints.}
        A significant challenge in COD is the lack of real-time performance, primarily due to the high computational and memory requirements of current models. This limitation impedes deployment on resource-constrained devices, restricting real-world applications such as surveillance and autonomous navigation. Train-free learning\cite{tang2023generalizationhallucinationlargevisionlanguage,hu2023relax} offers a solution by enabling rapid adaptation without retraining, thereby facilitating quicker deployment. Additionally, green learning-based gradient boosting\cite{chen2024greencodgreencamouflagedobject} emphasizes the development of energy-efficient models that reduce computational costs by selectively focusing on challenging samples, thereby optimizing real-time performance. The development of efficient networks\cite{10095226,ji2023gradient} with lightweight architectures aims to minimize latency and memory usage, making real-time COD feasible in practical scenarios. Overcoming these constraints is essential for the broader adoption and practical impact of COD technologies.
        
        \subsection{Exploring expansive potentials}
        \noindent\textbf{Embracing Novel Tasks.}
        CoCOD\cite{zhang2023collaborative}, a collaborative approach to COD, holds the potential to significantly enhance performance by leveraging multi-source data and cross-modal interactions. However, challenges persist in efficiently fusing heterogeneous information and ensuring robust performance across diverse scenarios. Another promising avenue, RefCOD\cite{cheng2023largemodelbasedreferring,zhang2023referring}, demands advancements in techniques for multi-modal alignment and the comprehension of complex textual or visual references, particularly for rare or ambiguous species. The primary difficulties include bridging cross-modal gaps and extracting fine-grained visual cues that are relevant to the provided textual or visual descriptions. Beyond CoCOD and RefCOD, there are numerous unexplored opportunities for novel tasks that could further empower COD. For instance, Interactive COD (I-COD) could incorporate user feedback during detection, enabling iterative refinement and personalization. This would require the development of robust interaction mechanisms and ensuring the system's responsiveness to user inputs~\cite{ma2022visual}. By continually exploring and innovating in these novel tasks, we can substantially enhance the capabilities and applicability of COD, thereby pushing the boundaries of what is possible in this exciting field.
        
        \noindent\textbf{Differentialting SOD and COD.}
        At the feature level, delving into the nuances between SOD and COD is crucial, as both, though under the umbrella of abnormal segmentation, target distinct abnormalities~\cite{yang2023directional,yang2022biconnet}. Current COD techniques struggle with salient objects~\cite{he2023strategic,He2023Camouflaged}, as they misinterpret prominence as camouflage, thereby necessitating robustness enhancement. Transferring salient objects to concealed scenarios, or vice versa, could alleviate data scarcity and boost model generalization. The partial positive correlation between SOD and COD highlights opportunities for conversion strategies, increasing sample diversity. Integrating generative adversarial mechanisms between the two tasks fosters innovation. Ultimately, a multi-task unified learning framework that encapsulates domain self-generalization for saliency and camouflage detection within the broader AI landscape represents a compelling frontier.
        
        \noindent\textbf{Multimodal information fusion.}
        Integrating multiple modalities, such as text\cite{luo2024vscode}, audio, video\cite{pang2024zoomnextunifiedcollaborativepyramid}, optical\cite{yang2021self}, depth\cite{Wang_2024_CVPR}, infrared, and 3D, presents a promising yet challenging direction in COD. Each modality offers unique insights but also introduces additional complexities. For instance, RGB-T and thermal infrared modalities can enhance detection in low-light or obscured environments, yet they require robust fusion strategies to effectively combine visual cues with temperature variations. Audio cues, while informative for certain concealed objects, present challenges in accurately localizing and interpreting sounds amidst background noise. Video data\cite{hui2024endow}, though rich in temporal information, necessitates efficient processing techniques to handle high dimensionality and real-time requirements. The key challenges lie in designing robust fusion mechanisms that can leverage the complementary strengths of these modalities while mitigating their individual limitations. Furthermore, annotating multi-modal datasets for concealed objects is both time-consuming and expensive, exacerbating the scarcity of labeled data. Addressing these challenges will require innovative approaches in multi-modal representation learning, fusion algorithms, and data augmentation techniques that are specifically tailored for COD.

        \noindent\textbf{Efficiency-oriented large-scale vision-language models.}
        Large vision-language models (VLMs) have demonstrated superior performance across various tasks, attracting significant attention from researchers. However, training VLMs specifically tailored for camouflaged object detection (COD) is not ideal due to limited data availability and the heavy computational demands involved. In this context, training-free prompt engineering has emerged as a promising direction to address the resource-intensive challenges associated with training large models. The primary challenge lies in designing effective prompts that can elicit nuanced understanding from these models without requiring extensive training. Recent advancements, such as SAM\cite{kirillov2023segment} and LVLMs\cite{hu2023relax}, offer compelling avenues for integrating visual and linguistic modalities, yet efficiency remains a major concern. Developing relatively lightweight models, combined with advanced prompt engineering techniques, could potentially mitigate efficiency issues while still handling diverse and challenging samples. SAM 2\cite{ravi2024sam2segmentimages}, as a pioneering effort, significantly advances the frontier of real-time video processing by leveraging a streaming memory transformer architecture. Its success in reducing interactions by 3x in video segmentation underscores the potential for enhancing efficiency in COD tasks, particularly within the video domain. To balance performance with efficiency, we could leverage the strengths of both large and lightweight models, potentially through techniques such as knowledge distillation\cite{hinton2015distilling}, transfer learning\cite{tan2018survey}, model compression\cite{buciluǎ2006model}, domain adaptation\cite{zhang2023unsupervisedcamouflagedobjectsegmentation}, meta-learning\cite{hospedales2021meta}, or modularized architectures tailored for specific tasks. Additionally, when there is a significant domain gap between training data and test data, designing a plug-and-play component is a promising approach. For instance, adapters\cite{xing2023pretrainadaptdetectmultitask} enable efficient fine-tuning of large pre-trained models, allowing them to adapt to the nuances of camouflaged objects without the full cost of retraining and thereby significantly reducing computational overheads.

        \section{Conclusion}
        We present a comprehensive and exhaustive overview of the rapidly evolving field of COD, encompassing both traditional and deep learning methods in image and video domains. By reviewing approximately 150 relevant COD studies, we offer the most extensive and detailed survey to date, providing a concise yet thorough perspective for both newcomers and established scholars. Additionally, we provide both quantitative and qualitative benchmarks for representative image and video models across 6 characteristic datasets and 6 evaluation metrics. Through rigorous benchmarking, we have identified key limitations and challenges in current COD methods, thereby paving the way for future research directions. We hope this survey will inspire innovative solutions that push the boundaries of COD technology. Additionally, we have established a dedicated \href{https://github.com/ChunmingHe/awesome-concealed-object-segmentation}{GitHub repository} to house COD techniques, datasets, and resources, ensuring that the latest developments and insights are readily accessible to the research community for further exploration. 

\ifCLASSOPTIONcompsoc
  \section*{Acknowledgments}
\else
  \section*{Acknowledgment}
\fi

The authors express their sincere appreciation to Dr. Dengping Fan for his insightful comments, which greatly improved the quality of this paper.
	
\bibliographystyle{ieeetr}
\bibliography{reference}
\end{document}